\documentclass[preprint]{imsart}
\arxiv{arXiv:xxxx.xxxxx}
\usepackage{amsmath,amsthm,verbatim,amssymb,amsfonts,amscd,graphicx}
\usepackage[colorlinks,citecolor=blue,urlcolor=blue]{hyperref}
\usepackage[ruled,vlined,commentsnumbered,linesnumbered]{algorithm2e}
\usepackage[numbers]{natbib}
\usepackage{bm}
\usepackage{mathrsfs}
\usepackage{subfig}
\usepackage{constants}
\usepackage{enumerate}
\usepackage{mathabx} 
\usepackage{algorithmic} 

\newcommand{\rom}[1]{\uppercase\expandafter{\romannumeral #1\relax}}
\newcommand{\bms}{\begin{multline*}}
\newcommand{\ems}{\end{multline*}}
\newcommand{\bels}{\begin{align*}}
\newcommand{\enls}{\end{align*}}
\newcommand{\bel}{\begin{align}}

\newcommand{\ignore}[1]{}

\newcommand{\bml}{\begin{multline}}
\newcommand{\eml}{\end{multline}}

\newcommand{\bL}{\widebar{\mathcal L}}
\newcommand{\hL}{\widehat{\mathcal L}^{(k)}}
\newcommand{\hEk}{{\widehat e}^{(k)}}
\newcommand{\Ek}{{e}^{(k)}}
\newcommand{\hGk}{{\widehat G}_k}
\newcommand{\hGj}{{\widehat G}_{|J|}}
\newcommand{\Gk}{G_k}

\startlocaldefs

\numberwithin{equation}{section}

\newtheorem{theorem}{Theorem}[section]

\newtheorem{lemma}{Lemma}[section]
\newtheorem{assumption}{Assumption}
\newtheorem{remark}{Remark}[section]

\endlocaldefs

\def\blfootnote{\xdef\@thefnmark{}\@footnotetext}

\newcommand{\dotp}[2]{\left\langle#1,#2\right\rangle}
\newcommand{\m}{\mathcal}
\newcommand{\mb}{\mathbb}
\newcommand\argmin{\mathop{\mbox{argmin}}}

\newcommand{\card}{\mathrm{Card}}
\newcommand{\sign}{\mathrm{sign}}
\def\r{\right}
\def\l{\left}
\newcommand{\eps}{\varepsilon}
\newcommand{\var}{\mbox{Var}}

\newcommand{\wh}{\widehat}

\newcommand{\pr}[1]{\mathrm{Pr}{\left(#1 \right)}}

\begin{document}

\begin{frontmatter}
\title{Excess risk bounds in robust empirical risk minimization}
\runauthor{}
\runtitle{Robust ERM}

\begin{aug}
\author{\fnms{Timoth\'{e}e} \snm{Mathieu }\thanksref{a,e1}\ead[label=e1,mark]{timothee.mathieu@u-psud.fr}}
\and
\author{\fnms{Stanislav} \snm{Minsker}\thanksref{b,e2}\ead[label=e2,mark]{minsker@usc.edu}}

\address[a]{1. Laboratoire de Mathematiques d'Orsay, Univ. Paris-Sud, CNRS, Université Paris-Saclay, 91405 Orsay, France and Inria Saclay - Ile-de-France, Bt. Turing, Campus de l'Ecole Polytechnique, 91120 Palaiseau, France.\newline
\printead{e1}}
\address[b]{Department of Mathematics, University of Southern California, Los Angeles, CA 90089.
\printead{e2}}

\end{aug}

\begin{abstract}
This paper investigates robust versions of the general empirical risk minimization algorithm, one of the core techniques underlying modern statistical methods. Success of the empirical risk minimization is based on the fact that for a ``well-behaved'' stochastic process $\l\{ f(X), \ f\in \m F\r\}$ indexed by a class of functions $f\in \m F$, averages $\frac{1}{N}\sum_{j=1}^N f(X_j)$ evaluated over a sample $X_1,\ldots,X_N$ of i.i.d. copies of $X$ provide good approximation to the expectations $\mb E f(X)$ uniformly over large classes $f\in \m F$. However, this might no longer be true if the marginal distributions of the process are heavy-tailed or if the sample contains outliers. We propose a version of empirical risk minimization based on the idea of replacing sample averages by robust proxies of the expectation, and obtain high-confidence bounds for the excess risk of resulting estimators. In particular, we show that the excess risk of robust estimators can converge to $0$ at fast rates with respect to the sample size.  
We discuss implications of the main results to the linear and logistic regression problems, and evaluate the numerical performance of proposed methods on simulated and real data. 
\end{abstract}
\end{frontmatter}


\section{Introduction}
\label{sec:intro}

This work is devoted to robust algorithms in the framework of statistical learning. A recent Forbes article \cite{forbes1} states that  ``Machine learning algorithms are very dependent on accurate, clean, and well-labeled training data to learn from so that they can produce accurate results'' and ``According to a recent report from AI research and advisory firm Cognilytica, over 80\% of the time spent in AI projects are spent dealing with and wrangling data.''  
While some abnormal samples, or outliers, can be detected and filtered during the preprocessing steps, others are more difficult to detect: for instance, a sophisticated adversary might try to ``poison'' data to force a desired outcome \cite{mayzlin2014promotional}. Other seemingly abnormal observations could be inherent to the underlying data-generating process. An ``ideal'' learning method should not discard informative samples, while limiting the effect of individual observation on the output of the learning algorithm at the same time. 
We are interested in robust methods that are model-free, and require minimal assumptions on the underlying distribution. 
We study two types of robustness: robustness to heavy tails expressed in terms of the moment requirements, as well as robustness to adversarial contamination. Heavy tails can be used to model variation and randomness naturally occurring in the sample, while adversarial contamination is a convenient way to model outliers of unknown nature. 

The statistical framework used throughout the paper is defined as follows. Let $(S,\m S)$ be a measurable space, and let $X\in S$ be a random variable with distribution $P$. 
Suppose that $X_1,\ldots,X_N$ are i.i.d. copies of $X$. Moreover, assume that $\m F$ is a class of measurable functions from $S$ to $\mb R$ and $\ell: \mb R\to \mb R_+$ is a loss function. 
Many problems in statistical learning theory can be formulated as risk minimization of the form 
\[
\mb E\, \ell(f(X)) \to \min_{f\in \m F}.
\]
We will frequently write $P \ell(f)$ or simply $\m L(f)$ in place of the expected loss $\mb E \ell \l(f(X)\r)$. 
Throughout the paper, we will also assume that the minimum is attained for some (unique) $f_\ast \in \m F$.
For example, in the context of regression, $X=(Z,Y)\in \mb R^d\times \mb R$, $f(Z,Y) = Y - g(Z)$ for some $g$ in a class $\m G$ (such as the class of linear functions), $\ell(x) = x^2$, and $f_\ast(z,y)= y - g_\ast(z)$, where $g_\ast(z) = \mb E\l[ Y| Z=z \r]$ is the conditional expectation. As the true distribution $P$ is usually unknown, a proxy of $f_\ast$ is obtained via \emph{empirical risk minimization} (ERM), namely
\begin{equation}
\label{eq:erm-standard}
\tilde f_N := \argmin_{f\in \m F} \m L_N(f), 
\end{equation}
where $P_N$ is the empirical distribution based on the sample $X_1,\ldots,X_N$ and 
\[
\m L_N(f) := P_N\, \ell_f = \frac{1}{N}\sum_{j=1}^N \ell\l(f(X_j)\r).
\] 
Performance of any $f\in \m F$ (in particular, $\tilde f_N$) is measured via the excess risk $\m E(f):= P \ell(f) - P \ell\l(f_\ast\r).$
The excess risk of $\tilde f_N$ is a random variable 
\[
\m E(\tilde f_N) := P \,\ell \big(\tilde f_N\big) - P \ell \l(f_\ast\r) = \mb E\l[\ell \big(\tilde f(X)\big)|X_1,\ldots,X_N \r] - \mb E \ell\l(f_\ast (X)\r).
\] 
General bounds for the excess risk have been extensively studied; a small subsample of the relevant works includes the papers \cite{van2000empirical,wellner2013weak,Koltchinskii2011Oracle-inequali00,anthony2009neural,bartlett2005local,tsybakov2004optimal} and references therein. 
However, until recently sharp estimates were known only in the situation when the functions in the class 
$\ell(\m F):=\l\{ \ell(f), \ f\in \m F\r\}$ are uniformly bounded, or when the envelope 
$F_\ell(x) := \sup_{f\in \m F}|\ell(f(x))|$ of the class $\ell(\m F)$ possesses finite exponential moments. 
Our focus is on the situation when marginal distributions of the process $\{\ell(f(X)), \ f\in \m F\}$ indexed by $\m F$ are allowed to be heavy-tailed, meaning that they possess finite moments of low order only (in this paper, ``low order'' usually means between 2 to 4). 
In such cases, the tail probabilities of the random variables 
$\l\{ \frac{1}{\sqrt N}\sum_{j=1}^N \ell(f(X_j)) - \mb E\ell(f(X)), \ f\in \m F\r\}$ decay polynomially, thus rendering many existing techniques ineffective. 
Moreover, we consider a challenging framework of \emph{adversarial contamination} where the initial dataset of cardinality $N$ is merged with a set of $\m O<N$ outliers which are generated by an adversary who has an opportunity to inspect the data, and the combined dataset of cardinality $N^\circ = N+\m O$ is presented to an algorithm; in this paper, we assume that the proportion of contamination $\frac{\m O}{N}$ (or its upper bound) is known.  

The approach that we propose is based on replacing the sample mean that is at the core of ERM by a more ``robust'' estimator of $\mb E\, \ell(f(X))$ that exhibits tight concentration under minimal moment assumptions. 
Well known examples of such estimators include the median-of-means estimator \cite{Nemirovski1983Problem-complex00,alon1996space,lerasle2011robust} and Catoni's estimator \cite{catoni2012challenging}. 
Both the median-of-means and Catoni's estimators gain robustness at the cost of being biased. 
The ways that the bias of these estimators is controlled is based on different principles however. Informally speaking, Catoni's estimator relies on delicate ``truncation'' of the data, while the median-of-means (MOM) estimator exploits the fact that the median and the mean of a symmetric distribution both coincide with its center of symmetry. 
In this paper, we will use ``hybrid'' estimators that take advantage of both symmetry and truncation. This family of estimators has been introduced and studied in \cite{minsker2017distributed,minsker2018uniform}, and we review the construction below. 

\subsection{Organization of the paper.}

The main ideas behind the proposed estimators are explained in Section \ref{sec:construction}, followed by the high-level overview of the main theoretical results and comparison to existing literature in Section \ref{sec:overview}. In Section \ref{sec:numerical}, we discuss practical implementation and numerical performance of our methods for two problems, linear regression and binary classification. The complete statements of the key results are given in Section \ref{section:main}, and in Section \ref{sec:examples} we deduce the corollaries of these results for specific examples. 
Finally, the architecture of the proofs is explained in Section \ref{section:proofs}, while the remaining technical arguments and additional numerical results are contained in the appendix.

\subsection{Notation.}
\label{sec:definitions}

For two sequences $\l\{ a_j\r\}_{j\geq 1}\subset \mb R$ and $\l\{ b_j \r\}_{j\geq 1}\subset \mb R$ for $j\in\mb N$, the expression 
$a_j\lesssim b_j$ means that there exists a constant $c>0$ such that $a_j\leq c b_j$ for all $j\in \mb N$; $a_j\asymp b_j$ means that $a_j \lesssim b_j$ and $b_j \lesssim a_j$. 
Absolute constants will be denoted $c,c_1,C,C'$, etc, and may take different values in different parts of the paper. 
For a function $h:\mb R^d\mapsto\mb R$, we define 
\[
\argmin_{y\in\mb R^d} h(y) = \{y\in\mb R^d: h(y)\leq h(x)\text{ for all }x\in \mb R^d\},
\]
and $\|h\|_\infty:=\mathrm{ess \,sup}\{ |h(y)|: \, y\in \mb R^d\}$. 
Moreover, $L(h)$ will stand for a Lipschitz constant of $h$. 
For $f\in \m F$, let $\sigma^2(\ell,f) = \var\l(\ell(f(X))\r)$ and for any subset $\m F'\subseteq \m F$, denote 
$\sigma^2(\ell,\m F') = \sup_{f\in \m F'}\sigma^2(\ell,f)$. 
Additional notation and auxiliary results are introduced on demand.

\subsection{Robust mean estimators.}
\label{sec:construction}

Let $k\leq N$ be an integer, and assume that $G_1,\ldots,G_k$ are disjoint subsets of the index set $\{1,\ldots,N\}$ of cardinality $|G_j| = n\geq \lfloor N/k\rfloor$ each. Given $f\in \m F$, let 
\[
\bL_j (f) := \frac{1}{n}\sum_{i\in G_j} \ell(f(X_i))
\]
be the empirical mean evaluated over the subsample indexed by $G_j$. Given a convex, even function $\rho:\mb R\mapsto \mb R_+$ and $\Delta>0$, set
\begin{align}
\label{eq:M-est}
\hL(f):= \argmin_{y\in \mb R}\sum_{j=1}^k \rho\l(\sqrt{n}\,\frac{\bL_j (f) - y}{\Delta}\r).
\end{align}
Clearly, if $\rho(x) = x^2$, $\hL(f)$ is equal to the sample mean. 
If $\rho(x)=|x|$, then $\hL(f)$ is the median-of-means estimator \cite{Nemirovski1983Problem-complex00,alon1996space,devroye2016sub}. 
We will be interested in the situation when $\rho$ is similar to Huber's loss, whence $\rho'$ is bounded and Lipchitz continuous (exact conditions imposed on $\rho$ are specified in Assumption \ref{ass:1} below). 
It is instructive to consider two cases: first, when $k=N$ (so that $n=1$) and $\Delta \asymp \sqrt{\var(\ell(f(X)))} \sqrt{N}$, 
$\hL(f)$ is akin to Catoni's estimator \cite{catoni2012challenging}, and when $n$ is large and $\Delta \asymp \sqrt{\var(\ell(f(X)))}$, we recover the ``median-of-means type'' estimator.
\footnote{The ``standard'' median-of-means estimator corresponds to $\rho(x)=x$ and can be seen as a limit of $\hL(f)$ when $\Delta\to 0$; this case is not covered by results of the paper, as we will require that $\rho'$ is smooth and $\Delta$ is bounded from below.} 

We also construct a permutation-invariant version of the estimator $\hL(f)$ that does not depend on the specific choice of the subgroups $G_1,\ldots,G_k$. Define
\[
\m A_N^{(n)}:=\l\{  J: \ J\subseteq \{1,\ldots,N\}, \card(J)=n \r\}.
\] 
Let $h$ be a measurable, permutation-invariant function of $n$ variables. Recall that a U-statistic of order $n$ with kernel $h$ based on an i.i.d. sample $X_1,\ldots,X_N$ is defined as \citep{hoeffding1948class}
\begin{equation}
\label{u-stat}
U_{N,n} = \frac{1}{{N\choose n}}\sum_{J\in \m A_N^{(n)}} h \l( \{X_j\}_{j \in J} \r).
\end{equation}
Given $J\in A_N^{(n)}$, let $\bL(f; J):=\frac{1}{n}\sum_{i\in J} f(X_i)$. Consider U-statistics of the form 
\[
U_{N,n}(z;f) = \sum_{J\in \m A_N^{(n)}} \rho\l(\sqrt{n}\,\frac{\bL(f;J) - z}{\Delta}\r).
\]
Then the permutation-invariant version of $\hL(f)$ is naturally defined as
\begin{equation}
\label{eq:M-est-U}
\hL_U(f):= \argmin_{z\in \mb R} U_{N,n}(z;f).
\end{equation}
Finally, assuming that $\wh{\m L}^{(k)}(f)$ provides good approximation of the expected loss $\m L(f)$ of each individual $f\in \m F$, it is natural to consider
\begin{equation}
\label{eq:fhat}
\wh f_N := \argmin_{f\in \m F} \wh{\m L}^{(k)}(f),
\end{equation}
as well as its permutation-invariant analogue 
\begin{equation}
\label{eq:fhat-U}
\wh f^U_{N} := \argmin_{f\in \m F} \wh{\m L}_U^{(k)}(f)
\end{equation}
as an alternative to standard empirical risk minimization \eqref{eq:erm-standard}. 
The main goal of this paper is to obtain general bounds for the excess risk of the estimators $\wh f_N$ and $\wh f^U_N$ under minimal assumptions on the stochastic process $\l\{ \ell(f(X)), \ f\in \m F\r\}$. 
More specifically, we are interested in scenarios when the excess risk converges to $0$ at fast, or ``optimistic'' rates, referring to the rates faster than $N^{-1/2}$. 
Rate of order $N^{-1/2}$ (``slow rates'') are easier to establish: in particular, results of this type follow from bounds on the uniform deviations $\sup_{f\in \m F}\l| \hL(f) - \m L(f)\r|$ that have been investigated in \cite{minsker2018uniform}. 
Proving fast rates is a more technically challenging task: to achieve the goal, we study remainder terms in Bahadur-type representations of the estimators $\hL(f)$ and $\hL_U(f)$ that provide linear (in $\ell(f)$) approximations of these nonlinear statistics and are easier to study. 

Let us remark that exact evaluation of the U-statistics based estimators $\hL_U(f)$ and $\wh f_N^U$ is not feasible due to the number of summands $N\choose n$ being very large even for small values of $n$. 
However, exact computation is typically not required, and throughout our detailed simulation studies, gradient descent methods proved to be very efficient for the problem \eqref{eq:fhat-U} in scenarios like least-squares and logistic regression. Moreover, numerical performance of the permutation-invariant estimator $\wh f_N^U$ is never worse than $\wh f_N$, and often is significantly better; these points are further discussed in Section \ref{sec:numerical}.  

\subsection{Overview of the main results and comparison to existing bounds.}
\label{sec:overview}

Our main contribution is the proof of high-confidence bounds for the excess risk of the estimators $\wh f_N$ and $\wh f^U_N$. 
First, we show that rates of order $N^{-1/2}$ are achieved with exponentially high probability if 
$\sigma(\ell,\m F) = \sup_{f\in \m F} \sigma^2(\ell,f) < \infty$ and 
$\mb E\sup_{f\in \m F}\frac{1}{\sqrt N}\sum_{j=1}^N \l( \ell(f(X_j)) - \mb E\ell(f(X))\r)<\infty$. The latter is true if the class $\l\{ \ell(f), \ f\in \m F\r\}$ is P-Donsker \cite{dudley2014uniform}, in other words, if the empirical process $f\mapsto \frac{1}{\sqrt N}\sum_{j=1}^N \l( \ell(f(X_j)) - \mb E\ell(f(X))\r)$ converges weakly to a Gaussian limit. 
Next, we demonstrate that under additional assumption requiring that any $f\in \m F$ with small excess risk must be close to $f_\ast$ that minimizes the expected loss, $\wh f_N$ and $\wh f^U_N$ attain fast rates; we state the bounds only for 
$\wh f_N$ while the results for $\wh f^U_N$ are similar, up to the change in absolute constants. 
\begin{theorem}[Informal]
\label{th:1.1}
Assume that $\sigma(\ell, \m F)<\infty$. Then, for appropriately set $k$ and $\Delta$, 
\[
\m E(\wh f_N) \leq \widebar\delta + C(\m F,P)\l(\frac{s}{N^{2/3}} + \l(\frac{\m O}{N}\r)^{2/3}\r)
\]
with probability at least $1-e^{-s}$ for all $s\lesssim k$. 
Moreover, if $\sup_{f\in \m F}\mb E^{1/4}\l( \ell(f(X)) - \mb E\ell(f(X)) \r)^4<\infty$, then
\[
\m E(\wh f_N) \leq \widebar\delta + C(\m F,P)\l(\frac{s}{N^{3/4} } +  \l(\frac{\m O}{N}\r)^{3/4}\r),
\]
again with probability at least $1-e^{-s}$ for all $s\lesssim k$ simultaneously. 
\end{theorem}
Here, $\widebar\delta$ is the quantity (formally defined in \eqref{eq:delta-bar} below) that often coincides with the optimal rate for the excess risk \cite{alquier2017estimation,lugosi2016risk}. 
Moreover, we design a two-step estimator based on $\wh f_N$ that is capable of achieving faster rates whenever $\widebar \delta \ll N^{-3/4}$. 
\begin{theorem}[Informal]
\label{th:1.2}
Assume that $\sup_{f\in \m F}\mb E^{1/4}\l( \ell(f(X)) - \mb E\ell(f(X)) \r)^4<\infty$. 
There exists an estimator $\wh f_N''$ such that 
\[
\m E\l( \widehat f''_N\r)\leq \widebar \delta + C(\m F,P,\rho)\l(\frac{\m O}{N} + \frac{s}{N}\r)
\]
with probability at least $1-e^{-s}$ for all $1\leq s\leq s_{\max}$ where $s_{\max}\to\infty$ as $N\to \infty$. 
\end{theorem}
\noindent Estimator $\wh f_N''$ is based on a two-step procedure, where $\wh f_N$ serves as an initial approximation that is refined on the second step via the risk minimization restricted to a ``small neighborhood'' of $\wh f_N$. 

Robustness of statistical learning algorithms has been studied extensively in recent years. 
Existing research has mainly focused on addressing robustness to heavy tails as well as adversarial contamination. 
One line of work investigated robust versions of the gradient descent for the optimization problem \eqref{eq:erm-standard} based on variants of the multivariate median-of-means technique \cite{prasad2018robust,chen2017distributed,yin2018byzantine,alistarh2018byzantine}, as well as Catoni's estimator \cite{holland2017efficient}. While these algorithms admits strong theoretical guarantees, they require robustly estimating the gradient vector at every step hence are computationally demanding; moreover, results are weaker for losses that are not strongly convex (for instance, the hinge loss). 

The line of work that is closest in spirit to the approach of this paper has includes the works that employ robust risk estimators based on Catoni's approach \cite{audibert2011robust,brownlees2015empirical,holland2017robust} and the median-of-means technique, such as ``tournaments'' and the ``min-max median-of-means'' \cite{lugosi2016risk,lugosi2017regularization,lecue2017robust,lecue2018robust,chinot2018statistical}. 
As it was mentioned in the introduction, the core of our methods can be viewed as a ``hybrid'' between Catoni's and the median-of-means estimators. We provide a more detailed comparison to the results of the aforementioned papers:
\begin{enumerate}
\item We show that risk minimization based on Catoni's estimator is capable of achieving fast rates, thus improving the results and weakening the assumptions stated in \cite{brownlees2015empirical};
\item Existing approaches based on the median-of-means estimators are either computationally intractable \cite{lugosi2016risk}, or outputs of practically efficient algorithms do not admit strong theoretical guarantees \cite{lecue2017robust,lecue2018robust,chinot2018statistical}. Our algorithms are designed specifically for the estimators $\wh f_N$ and $\wh f^U_N$, and enjoy good performance in numerical experiments along with strong theoretical guarantees simultaneously.
\item We develop new tools and techniques to analyze proposed estimators. 
In particular, we do not rely on the ``small ball'' method \cite{koltchinskii2015bounding,mendelson2014learning} and the standard ``majority vote-based'' analysis of the median-of-means estimators. Instead, we provide accurate bounds for the bias and investigate the remainder terms for the Bahadur-type linear approximations of the estimators \eqref{eq:M-est}. 
In particular, we demonstrate that the typical deviations of the estimator $\hL(f)$ around $\m L(f)$ are significantly smaller than the deviations of the subsample averages $\bL_j(f)$; consequently, this fact allows us to ``decouple'' the parameter $k$ responsible for the cardinality of subsamples from the confidence parameter $s$ that controls the deviation probabilities, and establish bounds that are uniform over a certain range of $s$ instead of a fixed level $s \asymp k$. 
Moreover, in cases when adversarial contamination is insignificant (e.g. $\m O = O(1)$), our algorithms, unlike existing results, admit a ``universal'' choice of $k$ that is independent of the parameter $\widebar \delta$ controlling the optimal rate.

We are able to treat the case of Lipschitz as well as non-Lipschitz (e.g., quadratic) loss functions $\ell$. At the same time, in some situations (e.g. linear regression with quadratic loss), our required assumptions are slightly stronger compared to the best results in the literature tailored specifically to the task \cite[e.g.][]{lugosi2016risk,lecue2017robust}.
\end{enumerate}

\section{Numerical algorithms and examples.}
\label{sec:numerical}

The main goal of this section is to discuss numerical algorithms used to approximate estimators $\wh f_N$ and $\wh f^U_N$, as well as assess the quality of resulting solutions. We will also compare our methods with the ones known previously, specifically, the median-of-means based approach proposed in \cite{lecue2018robust}.  
Finally, we perform the numerical study of dependence of the solutions on the parameters $\Delta$ and $k$.
All evaluations are performed for logistic regression in the framework of binary classification as well as linear regression with quadratic loss using simulated data, while applications to real data are shown in the appendix. 
Let us mention that the numerical methods for closely related approach in the special case of linear regression have been investigated in a recent work \cite{holland2017robust}. 
Here, we focus on general algorithms that can easily be adapted to other predictions tasks and loss functions. 
Let us first briefly recall the formulations of both the binary classification and the linear regression problems. 

 \textbf{Binary classification and logistic regression.} Assume that $(Z,Y)\in S\times \{\pm1 \}$ is a random couple where $Z$ is an instance and $Y$ is a binary label, and let $g_\ast(z):=\mb E[Y|Z=z]$ be the regression function. It is well-known that the binary classifier $b_\ast(z):=\sign (g_\ast(z))$ achieves smallest possible misclassification error defined as $P\l( Y\ne g(Z)\r)$. Let $\m F$ be a given convex class of functions mapping $S$ to $\mb R$, $\ell:\mb R\mapsto \mb R_+$ -- a convex, nondecreasing, Lipschitz loss function, and let 
\[
\rho_\ast = \argmin_{\text{all measurable } f} \mb E \ell(Yf(Z)).
\]
The loss $\ell$ is classification-calibrated if $\sign(\rho_\ast(z)) = b_\ast(z)$ P-almost surely; we refer the reader to \cite{bartlett2006convexity} for a detailed exposition. 
In the case of logistic regression considered below, $S=\mb R^d$,
\[
\ell(y,f(z))=\ell(yf(z)) := \log\l(1 + e^{-yf(z)}\r)
\]
is a classification-calibrated loss and $\m F = \l\{ f_\beta(\cdot) = \dotp{\cdot}{\beta}, \ \beta\in \mb R^d\r\}$ (as usual, the intercept term can be included if the vector $Z$ is replaced by $\tilde Z=(Z,1)$). 

 \textbf{Regression with quadratic loss.} 
Let $(Z,Y)\in S\times \mb R$ be a random couple satisfying $Y = f_\ast(Z) + \eta$ where the noise variable $\eta$ is independent of $Z$ and $f_\ast(z)=\mb E[Y|Z=z]$ is the regression function. Linear regression with quadratic loss corresponds to $S=\mb R^d$, 
\[
\ell(y,f(z))=\ell(y-f(z)) := (y-f(z))^2
\] 
and $\m F = \l\{f_\beta(\cdot) = \dotp{\cdot}{\beta}, \  \ \beta\in \mb R^d\r\}$. 

In both examples, we will assume that we are given an i.i.d. sample $(Z_1,Y_1),\ldots,(Z_N,Y_N)$ having the same distribution as $(Z,Y)$. 

\subsection{Gradient descent algorithms.}
\label{sec:descent}

Optimization problems \eqref{eq:fhat} and \eqref{eq:fhat-U} are not convex, so we will focus our attention of the variants of the gradient descent method employed to find local minima. 
We will first derive the expression for $\nabla_{\beta}\hL(\beta)$, the gradient of $\hL(\beta):=\hL(f_\beta)$, for the problems corresponding to logistic regression and regression with quadratic loss. 
It follows from \eqref{eq:M-est} that $\hL(\beta)$ satisfies the equation 
\begin{equation}
\label{eq:M-est_beta}
\sum_{j=1}^k \rho'\l(\sqrt{n}\frac{\bL_j(\beta)- \wh{\m L}^{(k)}(\beta)}{\Delta} \r) = 0. 
\end{equation}
Taking the derivative in \eqref{eq:M-est_beta} with respect to $\beta$, we retrieve $\nabla_{\beta}\hL(\beta)$:
\begin{equation}
\label{eq:gradient_algo}
\nabla_{\beta}\wh{\m L}^{(k)}(\beta) = 
\frac{\sum_{j=1}^k \l(\frac{1}{n}\sum_{i\in G_j}Z_i \, \ell'(Y_i, f_\beta(Z_i))\r)\rho''\l(\sqrt{n}\frac{\bL_j(\beta) - \hL(\beta)}{\Delta}\r)}{\sum_{j=1}^k\rho''\l(\sqrt{n}\frac{\bL_j(\beta) - \hL(\beta)}{\Delta}\r)},
\end{equation}
where $\ell'(Y_i, f_\beta(Z_i))$ stands for the partial derivative $\frac{\partial \ell(y,t)}{\partial t}$ with respect to the second argument $t$, so that 
$\ell'(Y_i, f_\beta(Z_i)) = -Y_i \frac{e^{-Y_i \dotp{\beta}{Z_i}}}{1 + e^{-Y_i \dotp{\beta}{Z_i}}}$ in the case of logistic regression and $\ell'(Y_i, f_\beta(Z_i)) = 2\l( \dotp{\beta}{Z_i} - Y_i\r)$ for regression with quadratic loss. 
\noindent In most of our numerical experiments, we choose $\rho$ to be Huber's loss,
\[
\rho(y) = \frac{y^2}{2} I \l\{|y|\leq 1\r\} + \l( |y|-\frac{1}{2} \r) I\l\{ |y|>1\r\}.
\]
In this case, $\rho''(y)=I\{|y|\le 1\}$ for all $y\in \mb R$, hence the expression for the gradient can be simplified to 
\begin{equation}
\label{eq:gradient_algo_2}
\nabla_{\beta}\hL(\beta) = 
\frac{\sum_{j=1}^k \l(\frac{1}{n}\sum_{i\in G_j}Z_i\, \ell'(Y_i,f_\beta(Z_i))\r) I\l\{\l| \bL_j(\beta) - \hL(\beta)\r|\le \frac{\Delta}{\sqrt{n}} \r\}} {\#\l\{j: \, \l| \bL_j(\beta) - \hL(\beta)\r| \le \frac{\Delta}{\sqrt{n}} \r\}},
\end{equation} 
where we implicitly assume that $\Delta$ is chosen large enough so that the denominator is not equal to $0$. 
To evaluate $\hL(\beta)$, we use the  ``modified weights'' algorithm due to Huber and Ronchetti \cite[][section 6.7]{robuststat}.
Complete version of the gradient descent algorithm used to approximate $\wh\beta_N$ (identified with the solution $\wh f_N$ of the problem \eqref{eq:fhat}) is presented in Figure \ref{fig:algo1}.
\begin{figure}[h]
   \caption{Algorithm 1 -- evaluation of $\wh\beta_N$.}
   \label{fig:algo1}
    \begin{algorithmic}
      \STATE \textbf{Input:} the dataset $(Z_i,Y_i)_{1\le i\le N}$, number of blocks $k\in \mb N$, step size parameter $\eta>0$, maximum number of iterations $M$, initial guess $\beta_0\in\mb R^d$, tuning parameter $\Delta \in \mb R$.
      \STATE Construct blocks $G_1,\ldots,G_k$;
      \FORALL{$t=0,\ldots,M$}
      \STATE Compute $\hL(\beta_t)$ using the Modified Weights algorithm;
      \STATE Compute $\nabla_{\beta}\hL(\beta_t)$ from equation~\ref{eq:gradient_algo_2};
      \STATE Update 
      \[
      \beta_{t+1} = \beta_t - \eta\nabla_{\beta} \hL(\beta_t).
      \]
	   \ENDFOR
      \STATE \textbf{Output:} $\beta_{M+1}$. 
    \end{algorithmic}
\end{figure}

\noindent Next, we discuss a variant of a stochastic gradient descent for approximating the ``permutation-invariant'' estimator $\wh f_N^U$ used when the subgroup size $n>1$; in our numerical experiments (see Section~\ref{sec:num_algo} for the numerical comparison of two approaches), this method demonstrated consistently superior performance. Below, we will identify $\wh f_N^U$ with the vector of corresponding coefficients $\wh \beta_N^U$. Recall that $\m A_N^{(n)}:=\l\{  J: \ J\subseteq \{1,\ldots,N\}, \, \card(J)=n \r\}$, and that 
\begin{equation}
\label{eq:M-est-U-2}
\hL_U(\beta) = \argmin_{z\in \mb R} \sum_{J\in \m A_N^{(n)}} \rho\l(\sqrt{n}\,\frac{\bL(f_\beta;J) - z}{\Delta}\r).
\end{equation}
Similarly to the way that we derived the expression for $\nabla_{\beta}\wh{\m L}^{(k)}(\beta)$ from \eqref{eq:M-est}, it follows from \eqref{eq:M-est-U-2}, with $\rho$ again being the Huber's loss, that 
\begin{align}
\label{eq:gradient-U}
& \nonumber
\sum_{J\in \m A_N^{(n)}} \rho'\l(\sqrt{n}\,\frac{\bL(f_\beta;J) - \hL_U(\beta)}{\Delta}\r) = 0 \text{ \quad   and }
\\ 
&\nabla_{\beta}\hL_U(\beta) =
\frac{\sum_{J \in \m A_N^{(n)}} \l(\frac{1}{n}\sum_{i\in J} Z_i\, \ell'(Y_i,f_\beta(Z_i))\r) I \l\{ \l| \bL(\beta; J) - \hL(\beta)\r| \le \frac{\Delta}{\sqrt{n}} \r\} } {\#\l\{ J \in \m A_N^{(n)}:  \, \l| \bL(\beta;J) - \hL(\beta)\r| \le \frac{\Delta}{\sqrt{n}} \r\}}.
\end{align}
Expressions in \eqref{eq:gradient-U} are closely related to U-statistics, and it will be convenient to write them in a slightly different form. 
To this end, let $\pi_N$ be the collection of all permutations $i:\{1,\ldots,N\}\mapsto \{1,\ldots,N\}$. 
Given $\tau=(i_1,\ldots,i_N)\in \pi_N$ and an arbitrary U-statistic $U_{N,n}$ defined in \eqref{u-stat}, let 
\begin{equation*}
T_{i_1,\ldots,i_N}:=\frac{1}{k}\l( h\l(X_{i_1},\ldots,X_{i_n} \r) + h\l(X_{i_{n+1}},\ldots,X_{i_{2n}}\r) + \ldots + 
h\l(X_{i_{(k-1)n+1}},\ldots,X_{i_{kn}} \r) \r).
\end{equation*}
Equivalently, for $\tau = (i_1,\ldots,i_N)\in \pi_N$, let 
\begin{equation}
\label{eq:G_j}
G_j(\tau) = \l( i_{(j-1)n+1},\ldots, i_{jn}\r), \  j=1,\ldots,k=\lfloor N/n \rfloor,
\end{equation}
which gives a compact form
\[
T_{\tau} = \frac{1}{k}\sum_{j=1}^k h\l( X_i, i \in G_j(\tau)\r).
\] 
It is well known (section 5 in \cite{hoeffding1963probability}) that the following representation of the U-statistic holds:
\begin{equation}
\label{eq:U-decomp-0}
U_{N,n} = \frac{1}{N!}\sum_{\tau\in \pi_N}  T_{\tau}.
\end{equation}
Applying representation \eqref{eq:U-decomp-0} to \eqref{eq:M-est-U-2}, we deduce that 
\begin{equation}
\label{eq:grad-simple-1}
\hL_U(\beta) = \argmin_{z\in \mb R} \sum_{\tau \in \pi_N} \m R_\tau(\beta,z),
\end{equation}
with $\m R_\tau(\beta,z) = \sum_{j=1}^k \rho\l(\sqrt{n}\,\frac{\bL(f_\beta;G_j(\tau)) - z}{\Delta}\r)$. 
Similarly, applying representation \eqref{eq:U-decomp-0} to the numerator and the denominator in \eqref{eq:gradient-U}, we see that $\nabla_{\beta}\hL_U(\beta)$ can be written as a weighted sum
\begin{equation*}
\nabla_{\beta}\hL_U(\beta) = 
\sum_{\tau \in \pi_N} \underbrace{  \frac{ \sum_{j=1}^k I\l\{ \l| \bL(\beta;G_j(\tau)) - \hL(\beta)\r| \le \frac{\Delta}{\sqrt{n}} \r\} }{\sum_{\pi \in \pi_N}\sum_{j=1}^k  I\l\{ \l| \bL(\beta;G_j(\pi)) - \hL(\beta)\r| \le \frac{\Delta}{\sqrt{n}} \r\}} }_{\text{=$\omega_\tau$, weight  corresponding to permutation }\tau} \cdot \widetilde \Gamma_\tau(\beta),
\end{equation*}
where 
\begin{equation}
\label{eq:grad-simple}
\widetilde \Gamma_\tau(\beta) := 
\frac{\sum_{j=1}^k \l(\frac{1}{n}\sum_{i\in G_j(\tau)}Z_i\, \ell'(Y_i,f_\beta(Z_i))\r) I\l\{\l| \bL(\beta; G_j(\tau)) - \hL(\beta)\r|\le \frac{\Delta}{\sqrt{n}} \r\}} {\sum_{j=1}^k I\l\{ \l| \bL(\beta;G_j(\tau)) - \hL(\beta)\r| \le \frac{\Delta}{\sqrt{n}} \r\}}
\end{equation}
is similar to the expression for the gradient of $\hL(\beta)$ defined for a fixed partition $G_1(\tau),\ldots, G_k(\tau)$, see equation \eqref{eq:gradient_algo_2}. 
Representations in \eqref{eq:grad-simple-1} and \eqref{eq:grad-simple} can be simplified even further noting that permutations that do not alter the subgroups $G_1,\ldots,G_k$ also do not change the values of $\m R_\tau(\beta,z)$, $\omega_\tau$ and $\widetilde \Gamma_\tau(\beta)$. 
To this end, let us say that $\tau_1, \tau_2\in \pi_N$ are equivalent if $G_j(\tau_1)=G_j(\tau_2)$ for all $j=1,\ldots,k$. It is easy to see that there are $\frac{N!}{(n!)^k \cdot (N - nk)!}$ equivalence classes, and let $\pi_{N,n,k}$ be the set of permutations containing exactly one permutation from each equivalence class. We can thus write 
\begin{align}
\label{eq:final-represent}
\nonumber
\hL_U(\beta) &= \argmin_{z\in \mb R} Q(\beta,z):=\argmin_{z\in \mb R} \sum_{\tau \in \pi_{N,n,k}} \m R_\tau(\beta,z), 
\\
\nabla_{\beta}\hL_U(\beta) &= 
\sum_{\tau \in \pi_{N,n,k}} \widetilde \omega_\tau \cdot \widetilde \Gamma_\tau(\beta),
\end{align}
where $\widetilde\omega_\tau = (n!)^k \, (N-nk)! \cdot \omega_\tau$.  
Representation \eqref{eq:final-represent} suggests that in order to obtain an unbiased estimator of $\nabla_z Q(\beta,z)$, one can sample a permutation $\tau\in \pi_{N,n,k}$ uniformly at random, compute 
$\nabla_z \m R_\tau(\beta,z)$ and use it as a descent direction. This yields a version of the stochastic gradient descent for evaluating $\hL_U(\beta)$ presented in Figure \ref{fig:algo2}. 
 \begin{figure}[h]
   \caption{Algorithm 2 -- evaluation of $\hL_U(\beta)$.}
   \label{fig:algo2}
    \begin{algorithmic}
      \STATE \textbf{Input:} the dataset $(Z_i,Y_i)_{1\le i\le N}$, number of blocks $k\in \mb N$, step size parameter $\eta>0$, maximum number of iterations $M$, initial guess $z_0\in\mb R$, tuning parameter $\Delta \in \mb R$.
      \FORALL{$t=0,\ldots,M$}
      \STATE Sample permutation $\tau$ uniformly at random from $\pi_{N,n,k}$, construct blocks $G_1(\tau),\ldots,G_k(\tau)$ according to \eqref{eq:G_j};
      \STATE Compute $\nabla_z \m R_\tau(\beta,z_t) = - \frac{\sqrt n}{\Delta} \sum_{j=1}^k \rho'\l(\sqrt{n}\frac{\bL(f_\beta;G_j(\tau)) - z_t}{\Delta} \r)$;
      \STATE Update 
      \[
      z_{t+1} = z_t - \eta \nabla_z \m R_\tau(\beta,z_t).
      \]
	   \ENDFOR
      \STATE \textbf{Output:} $z_{M+1}$. 
    \end{algorithmic}
\end{figure}
Once a method for computing $\hL_U(\beta)$ is established, similar reasoning leads to an algorithm for finding $\wh f^U_N$. Indeed, using representation \eqref{eq:final-represent}, it is easy to see that an unbiased estimator of $\nabla_{\beta}\hL_U(\beta)$ can be obtained by first sampling a permutation $\tau\in \pi_{N,n,k}$ according to the probability distribution given by the weights $\l\{ \widetilde \omega_\tau, \ \tau\in \pi_{N,n,k} \r\}$, then evaluating 
$\widetilde \Gamma_\tau(\beta)$ using formula \eqref{eq:grad-simple}, and using $\widetilde \Gamma_\tau(\beta)$ as a direction of descent. 
In most typical cases, the number $M$ of the gradient descent iterations is much smaller than $\frac{N!}{(n!)^k \cdot (N - nk)!}$, whence it is unlikely that the same permutation will be repeated twice in the sampling process. 
This reasoning suggests the idea of replacing the weights $\widetilde \omega_\tau$ by the uniform distribution over $\pi_{N,n,k}$ that leads to a much faster practical implementation which is detailed in Figure \ref{fig:algo3}. 
\begin{figure}[h]
   \caption{Algorithm 3 -- evaluation of $\wh \beta_N^U$.}
   \label{fig:algo3}
    \begin{algorithmic}
      \STATE \textbf{Input:} the dataset $(Z_i,Y_i)_{1\le i\le N}$, number of blocks $k\in \mb N$, step size parameter $\eta>0$, maximum number of iterations $M$, initial guess $\beta_0\in\mb R^d$, tuning parameter $\Delta \in \mb R$.
      \FORALL{$t=0,\ldots,M$}
      \STATE Sample permutation $\tau$ uniformly at random from $\pi_{N,n,k}$, construct blocks $G_1(\tau),\ldots,G_k(\tau)$ according to \eqref{eq:G_j};
      \STATE Compute $\hL_U(\beta_t)$ using Algorithm 2 in Figure \ref{fig:algo2};
      \STATE Compute $\widetilde \Gamma_\tau(\beta_t)$ via equation~\ref{eq:grad-simple};
      \STATE Update 
      \[
      \beta_{t+1} = \beta_t - \eta\widetilde \Gamma_\tau(\beta_t).
      \]
	   \ENDFOR
      \STATE \textbf{Output:} $\beta_{M+1}$. 
    \end{algorithmic}
\end{figure}
It is easy to see that presented gradient descent algorithms for evaluating $\wh f_N$ and $\wh f_N^U$ have the same numerical complexity. 
The following subsections provide several ``proof-of-concept'' examples illustrating the performance of proposed methods, as well as comparison to the existing techniques.

\subsection{Logistic regression.}

\begin{figure}[h]
\begin{center}
\subfloat[Training Dataset]{
      \includegraphics[width=0.3\textwidth]{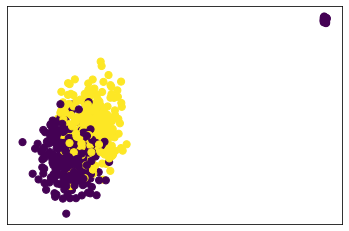}
      \label{sub:renonc}
                         }
    \subfloat[Decision function -- standard Logistic Regression]{
      \includegraphics[width=0.3\textwidth]{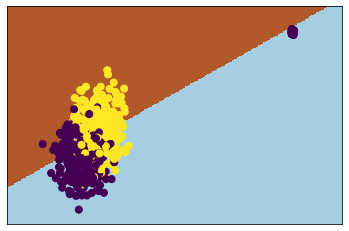}
      \label{sub:popul}
                         }
    \subfloat[Decision function -- Algorithm 3]{
      \includegraphics[width=0.3\textwidth]{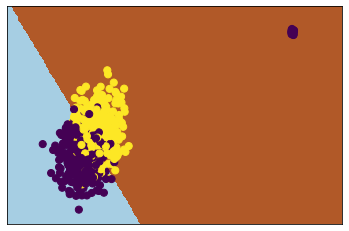}
      \label{sub:popul}
                         }
\caption{Scatter plot of 630 samples from the training dataset (600 informative observations, 30 outliers), the color of the points correspond to their labels and the background color -- to the predicted labels (brown region corresponds to ``yellow'' labels and blue -- to ``purple'').
\label{fig:scatter_plots}}
\end{center}
\end{figure}

The dataset consists of pairs $(Z_j,Y_j)\in \mb R^2\times\{\pm1\}$, where the marginal distribution of the labels is uniform and conditional distributions of $Z$ are normal, namely, $\mathrm{Law}\l(Z\,|\,Y=1\r) = \mathcal{N}\l((-1,-1)^T,1.4I_2\r)$, $\mathrm{Law}\l(Z\,|\,Y=-1\r) \sim \mathcal{N}\l((1,1),1.4I_2\r)$, and $\pr{Y=1}=\pr{Y=-1}=1/2$. 
The dataset includes outliers for which $Y\equiv 1$ and $Z\sim \mathcal{N}\l((24,8),0.1I_2\r)$, where $I_2$ stands for the $2\times2$ identity matrix. We generated 600 ``informative'' observations along with 30 outliers, and compared performance or our robust method (based on evaluating $\wh \beta_N^U$) with the standard logistic regression that is known to be sensitive to outliers in the sample (we used implementation available in the Scikit-learn package \cite{pedregosa2011scikit}). Results of the experiment are presented in Figure \ref{fig:scatter_plots}. 
Parameters $k$ and $\Delta$ in our implementation were tuned via cross-validation. 

\subsection{Linear regression.}
\label{sec:lin-reg}

In this section, we compare performance of our method (again based on evaluating $\wh \beta_N^U$) with standard linear regression as well as with robust Huber's regression estimator \cite[][section 7]{robuststat}; linear regression and Huber's regression were implemented using `LinearRregression' and `HuberRegressor' functions in the Scikit-learn package \cite{pedregosa2011scikit}. 
As in the previous example, the dataset consists of informative observations and outliers. 
Informative data $(Z_j,Y_j), \ j=1,\ldots,570$ are i.i.d. and satisfy the linear model $Y_j=10Z_j + \eps_j$ where $Z_j\sim \mathrm{Unif}[-3,3]$ and $\eps_j\sim \mathcal{N}(0,1)$. 
We consider two types of outliers: (a) outliers in the response variable $Y$ only, and (b) outliers in the predictor $Z$. It is well-known that standard linear regression is not robust in any of these scenarios, Huber's regression estimator is robust to outliers in response $Y$ only, while our approach is shown to be robust to corruption of both types. 
In both test scenarios, we generated $30$ outliers. 
Given $Z_j$, the outliers $Y_j$ of type (a) are sampled from a $\mathcal{N}\l(100,0.01\r)$ distribution, while the outliers of type (b) are $Z_j\sim \mathcal{N}\l((24,24)^T,0.01\,I_2\r)$. 
\begin{figure}[h]
\begin{center}
    \subfloat[Outliers in response variable]{
      \includegraphics[width=0.45\textwidth]{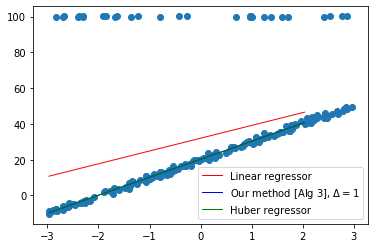}
                         }
    \subfloat[Outliers in predictors\label{fig:scatter_plots_reg:x}]{
      \includegraphics[width=0.45\textwidth]{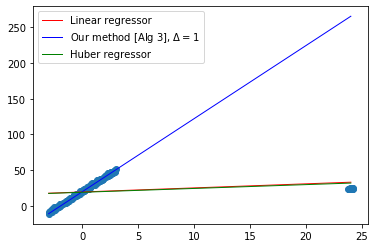}
                         }
\caption{Scatter plot of 600 training samples (570 informative data and 30 outliers) and the corresponding regression lines for our method, Huber's regression and regression with quadratic loss.
\label{fig:scatter_plots_reg}}
\end{center}
\end{figure}
Results are presented in Figure~\ref{fig:scatter_plots_reg}, and confirm the expected outcomes. 

\subsection{Choice of $k$ and $\Delta$.}

\begin{figure}[h]
\begin{center}
\subfloat[$MSE$ vs $k$\label{fig:choice_K}]{
\includegraphics[scale=0.45]{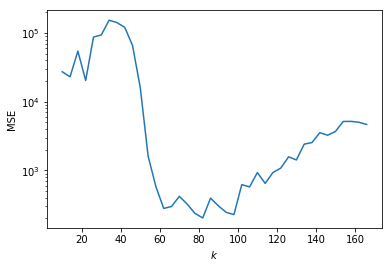}
}
    \subfloat[$MSE$ vs $\Delta$ (log-log scale)\label{fig:select_delta}]{
\includegraphics[scale=0.45]{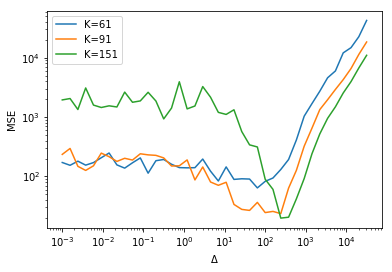}
      \label{sub:popul}
                         }
\caption{Plot of the tuning parameter ($x$-axis) against the MSE ($y$-axis) obtained with Algorithm 3. The MSE was evaluated via the Monte-Carlo approximation over $500$ samples of the data.}
\end{center}
\end{figure}

In this subsection, we evaluate the effect of different choices of $k$ and $\Delta$ in the linear regression setting of Section \ref{sec:lin-reg}, again with 570 informative observations and 30 outliers of type (b) as described in section \ref{sec:lin-reg} above. 
Figure~\ref{fig:choice_K} shows the plot of the resulting mean square error (MSE) against the number of subgroups $k$. As expected, the error decreases significantly when $k$ exceeds $60$, twice the number of outliers. At the same time, the MSE remains stable as $k$ grows up to $k\simeq 100$, which is a desirable property for practical applications. 
In this experiment, $\Delta$ was set using the ``median absolute deviation'' (MAD) estimator defined as follows. 
We start with $\Delta_0$ being a small number (e.g., $\Delta_0=0.1).$ 
Given a current approximate solution $\beta_t$, a permutation $\tau$ and the corresponding subgroups $G_1(\tau),\ldots,G_k(\tau)$, set $\widehat M(\beta_t):=\mathrm{median}\l(\hL(\beta_t; G_1(\tau),\ldots,\hL(\beta_t; G_k(\tau) \r)$, and
\[
\mathrm{MAD}(\beta_t) = \mathrm{median}\l( \l| \hL(\beta_t; G_1(\tau) - \widehat M(\beta_t)\r|, \ldots,\l| \hL(\beta_t; G_k(\tau) - \widehat M(\beta_t)\r| \r).
\] 
Finally, define $\wh \Delta_{t+1} := \frac{\mathrm{MAD}(\beta_t)}{\Phi^{-1}(3/4)}$, where $\Phi$ is the distribution function of the standard normal law.  After a small number $m$ (e.g. $m=10$) of ``burn-in'' iterations of Algorithm 3, $\Delta$ is fixed at the level $\wh \Delta_m$ for all the remaining iterations. 

Next, we study the effect of varying $\Delta$ for different but fixed values of $k$. 
To this end, we set $k\in \{ 61,91,151\}$, and evaluated the MSE as a function of $\Delta$. Resulting plot is presented in Figure~\ref{fig:select_delta}. 
The MSE achieves its minimum for $\Delta \asymp 10^{2}$; for larger values of $\Delta$, the effect of outliers becomes significant as the algorithm starts to resemble regression with quadratic loss (indeed, outliers in this specific example are at a distance $\approx 100$ from the bulk of the data). 

\subsubsection{Comparison with existing methods.}

\begin{figure}[h]
\begin{center}
    \subfloat[``Two moons'' dataset \cite{pedregosa2011scikit} with outliers.\label{fig:MOM_HOM:dataset}]{
\includegraphics[width=0.33\textwidth]{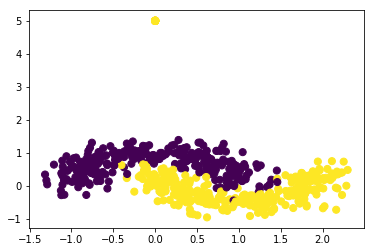}
}
\subfloat[$N=100$, no outliers \label{fig:MOM_HOM:no_outliers}]{
    \includegraphics[width=0.33\textwidth]{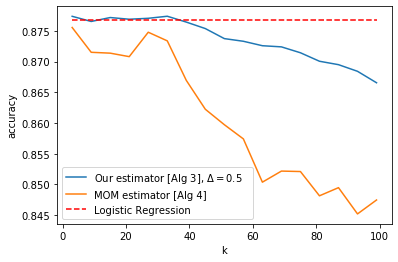}
}\\
\subfloat[$N=100$, with $10$ outliers\label{fig:MOM_HOM:with_outliers}]{
    \includegraphics[width=0.33\textwidth]{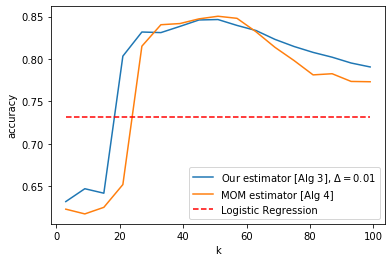}
}
\subfloat[$N=1000$, no outliers\label{fig:MOM_HOM:no_outliers}]{
    \includegraphics[width=0.33\textwidth]{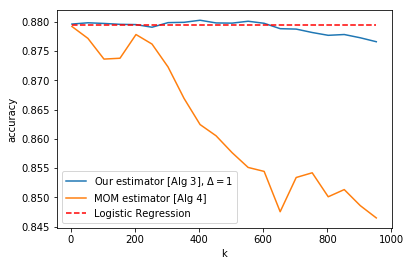}
}
\subfloat[$N=1000$, with $100$ outliers\label{fig:MOM_HOM:with_outliers}]{
    \includegraphics[width=0.33\textwidth]{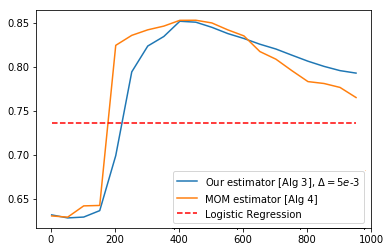}
}

\caption{Comparison of Algorithm 3, Algorithm 4 and standard logistic regression. The accuracy was evaluated using Monte-Carlo simulation over $300$ runs. 
\label{fig:MOM_HOM}}
\end{center}
\end{figure}

In this section, we compare performance of Algorithm 3 with a median-of-means-based robust gradient descent algorithm studied in~\cite{lecue2018robust}. The main difference of this method is in the way the descent direction is computed at every step. 
Specifically, $\widetilde \Gamma_\tau(\beta)$ employed in Algorithm 3 is replaced by $\nabla_\beta \m L^\diamond(\beta)$ where $\m L^\diamond(\beta):=\mathrm{median}\l( \bL(\beta; G_1(\tau),\ldots,\bL(\beta; G_k(\tau) \r)$, see Figure \ref{fig:algo4} and \cite{lecue2018robust} for the detailed description.
\begin{figure}[h]
   \caption{Algorithm 4.}
   \label{fig:algo4}
    \begin{algorithmic}
      \STATE \textbf{Input:} the dataset $(Z_i,Y_i)_{1\le i\le N}$, number of blocks $k\in \mb N$, step size parameter $\eta>0$, maximum number of iterations $M$, initial guess $\beta_0\in\mb R^d$.
      \FORALL{$t=0,\ldots,M$}
      \STATE Sample permutation $\tau$ uniformly at random from $\pi_{N,n,k}$, construct blocks $G_1(\tau),\ldots,G_k(\tau)$ according to \eqref{eq:G_j};
      \STATE Compute $\nabla_\beta \m L^\diamond(\beta)$;
      \STATE Update 
      \[
      \beta_{t+1} = \beta_t - \eta\nabla_\beta \m L^\diamond(\beta).
      \]
	   \ENDFOR
      \STATE \textbf{Output:} $\beta_{M+1}$. 
    \end{algorithmic}
\end{figure}
Experiments were performed for the logistic regression problem based on the ``two moons'' pattern, one of the standard datasets in the Scikit-learn package \cite{pedregosa2011scikit} presented in Figure~\ref{fig:MOM_HOM:dataset}. We performed two sets of experiments, one on the outlier-free dataset and one on the dataset consisting of 90\% of informative observations and 10\% of outliers, depicted as a yellow dot with coordinates $(0,5)$ on the plot. In both scenarios, we tested the ``small'' ($N=100$) and ``moderate" ($N=1000$) sample size regimes. We used standard logistic regression trained on an outlier-free sample as a benchmark; its accuracy is shown as a dotted red line on the plots. 
In all the cases, parameter $\Delta$ was tuned via cross-validation. In the outlier-free setting, our method (based on Algorithm 3) performed nearly as good as logistic regression; notably, performance of the method was strong even for large values of $k$, while classification accuracy decreased noticeably for Algorithm 4 for large $k$. 
In the presence of outliers, our method performed similar to Algorithm 4, while both methods outperformed standard logistic regression; for large values of $k$, our method was again slightly better. 
At the same time, Algorithm 4 was consistently faster than Algorithm 3 across the experiments.

\section{Theoretical guarantees for the excess risk.}
\label{section:main}

\subsection{Preliminaries.}
\label{sec:prelim}

In this section, we introduce the main quantities that appear in our results, and state the key assumptions. 
$\sigma^2(\ell,\m F') = \sup_{f\in \m F'}\sigma^2(\ell,f)$. 
The loss functions $\rho$ that will be of interest to us satisfy the following assumption.
\begin{assumption}
\label{ass:1}
Suppose that the function $\rho: \mb R\mapsto \mb R$ is convex, even, continuously differentiable 5 times and such that
\begin{enumerate}
\item[(i)] $\rho'(z)=z$ for $|z|\leq 1$ and $\rho'(z)=\mathrm{const}$ for $z\geq 2$.
\item[(ii)] $z - \rho'(z)$ is nondecreasing; 
\end{enumerate}
\end{assumption}
An example of a function $\rho$ satisfying required assumptions is given by ``smoothed'' Huber's loss defined as follows. Let
\[
H(y)=\frac{y^2}{2} I\{|y|\leq 3/2\} + \frac{3}{2}\l(|y| - \frac{3}{4}\r) I\{|y|>3/2\}
\] 
be the usual Huber's loss. Moreover, let $\phi$ be the ``bump function'' $\phi(x) = C \exp\l( -\frac{4}{1 - 4x^2}\r) \, \l\{ |x|\leq \frac{1}{2} \r\}$ where $C$ is chosen so that $\int_\mb R \phi(x)dx = 1$. Then $\rho$ given by the convolution $\rho(x) = (h\ast \phi)(x)$ satisfies assumption \ref{ass:1}. 
\begin{remark}
\label{rem:sup}
The derivative $\rho'$ has a natural interpretation of being a smooth version of the truncation function. Moreover, observe that $\rho'(2) - 2\leq \rho'(1)-1 = 0$ by (ii), hence $\|\rho'\|_\infty\leq 2$. It is also easy to see that for any $x>y$, 
$\rho'(x) - \rho'(y) = y - \rho'(y) - (x - \rho'(x)) + x - y \leq x-y$, hence $\rho'$ is Lipschitz continuous with Lipschitz constant $L(\rho')=1$. 
\end{remark}

Everywhere below, $\Phi(\cdot)$ stands for the cumulative distribution function of the standard normal random variable and $W(f)$ denotes a random variable with distribution $N\l( 0,\sigma^2(f) \r)$. 
For $f\in \m F$ such that $\sigma(f)>0$, $n\in \mb N$ and $t>0$, define
\begin{equation*}
\m R_f(t,n):= \l| \pr{ \frac{\sum_{j=1}^n \l(f(X_j) - Pf\r)}{\sigma(f)\sqrt{n}} \leq t} - \Phi( t ) \r|,
\end{equation*} 
where $Pf:=\mb E f(X)$. In other words, $g_f(t,n)$ controls the rate of convergence in the central limit theorem. 
It follows from the results of L. Chen and Q.-M. Shao \cite[Theorem 2.2 in][]{chen2001non} that 
\begin{multline*}
\m R_f(t,n)\leq g_f(t,n):=
C \Bigg( \frac{ \mb E (f(X) - \mb Ef(X))^2 \, I\l\{ \frac{ |f(X)-\mb Ef(X)|}{\sigma(f)\sqrt{n}} >  1+\l|\frac{t}{\sigma(f)}\r| \r\} }{\sigma^2(f) \l(1+\l|\frac{t}{\sigma(f)}\r| \r)^2} 
\\
+ \frac{1}{\sqrt{n}}
\frac{\mb E |f(X)-\mb Ef(X)|^3 \, I\l\{ \frac{|f(X) - \mb Ef(X)|}{\sigma(f)\sqrt{n}} \leq 1+ \l|\frac{t}{\sigma(f)}\r|\r\}}{\sigma^3(f) \l(1+\l|\frac{t}{\sigma(f)}\r| \r)^3}\Bigg)
\end{multline*}
given that the absolute constant $C$ is large enough. Moreover, let 
\[
G_f(n,\Delta) := \int_{0}^{\infty} g_f\l(\Delta\l( \frac{1}{2} + t \r),n \r)dt.
\] 
This quantity (more specifically, its scaled version $\frac{G_f(n,\Delta)}{\sqrt n}$ plays the key role in controlling the bias of the estimator $\hL(f)$. The following statement provides simple upper bounds for $g_f(t,n)$ and $G_f(n,\Delta)$. 
\begin{lemma}
\label{lemma:BE-nonunif}
Let $X_1,\ldots,X_n$ be i.i.d. copies of $X$, and assume that $\var(f(X))<\infty.$ Then $g_f(t,n)\to 0$ as $|t|\to\infty$ and $g_f(t,n)\to 0$ as $n\to\infty$, with convergence being monotone. Moreover, if $\mb E| f(X) - \mb Ef(X)|^{2+\delta}<\infty$ for some $\delta\in [0,1]$, then for all $t>0$
\begin{align}
\label{eq:g_f}
g_f(t,n)&\leq C'  \frac{\mb E\big|f(X) - \mb Ef(X) \big|^{2+\delta}}{n^{\delta/2}\l(\sigma(f)+ \l| t \r| \r)^{2+\delta}} \leq C' \frac{\mb E\big|f(X) - \mb Ef(X)\big|^{2+\delta}}{n^{\delta/2} |t|^{2+\delta}},
\\
\nonumber
G_f(n,\Delta) & \leq C'' \frac{\mb E\big|f(X) - \mb Ef(X)\big|^{2+\delta}}{\Delta^{2+\delta} n^{\delta/2}},
\end{align}
where $C',C''>0$ are absolute constants. 
\end{lemma}


\subsection{Slow rates for the excess risk.}
\label{section:M-est}

Let 
\begin{align*}
\wh\delta_N&:=\m E(\wh f_N) = \m L\big(\wh f_N\big) - \m L(f_\ast),
\\
\wh\delta^U_N&:=\m E(\wh f^U_N) = \m L\big(\wh f^U_N\big) - \m L(f_\ast)
\end{align*}
be the excess risk of $\wh f_N$ and its permutation-invariant analogue $\wh f_N^U$ which are the main objects of our interest. The following bound for the excess risk is well known: 
\begin{multline}
\label{eq:excess-slow}
\m E\big( \wh f_N \big) = \m L \big(\wh f_N \big) - \m L(f_\ast)
\\
= \m L \big(\wh f_N\big) + \hL(\wh f_N) - \hL(\wh f_N) + \hL(f_\ast) - \hL(f_\ast) - \m L(f_\ast) 
\\ 
= \l( \m L \big(\wh f_N\big) - \hL(\wh f_N) \r) - \l( \m L(f_\ast) - \hL(f_\ast)\r) + \underbrace{\hL(\wh f_N) - \hL(f_\ast)}_{\leq 0} 
\\
\leq 2\sup_{f\in \m F} \l| \hL(f) - \m L(f) \r|.
\end{multline}
The first result, Theorem \ref{th:unif} below, together with the inequality \eqref{eq:excess-slow} immediately implies the ``slow rate bound'' (meaning rate not faster than $N^{-1/2}$) for the excess risk. This result has been previously established in \cite{minsker2018uniform}.
Define 
\[
\widetilde \Delta:=\max\l( \Delta,\sigma(\ell,\m F)\r).
\]
\begin{theorem}
\label{th:unif}
There exist absolute constants $c,\, C>0$ such that for all $s>0,$ $n$ and $k$ satisfying
\begin{equation}
\label{eq:assump}
\frac{1}{\Delta} \l( \frac{1}{\sqrt{k}} \, \mb E\sup_{f\in \m F} \frac{1}{\sqrt{N}}\sum_{j=1}^N \l( \ell(f(X_j)) - P\,\ell(f) \r) + \sigma(\ell,\m F)\sqrt{\frac{s}{k}} \r) +
\sup_{f\in\m F} G_f(n,\Delta)
+ \frac{s}{k} + \frac{\m O}{k} \leq c,
\end{equation}
the following inequality holds with probability at least $1 - 2e^{-s}$:
\begin{multline*}
\sup_{f\in \m F}\l| \hL(f) - \m L(f) \r| \leq 
C 
\Bigg[ \frac{\widetilde \Delta}{\Delta} \l( \mb E\sup_{f\in \m F} \frac{1}{N} \sum_{j=1}^N \l( \ell(f(X_j)) - P\,\ell(f) \r)
+ \sigma(\ell,\m F) \sqrt{\frac{s}{N}} \r)
\\
+ \widetilde \Delta \l( \sqrt{n}\frac{s}{N}
+ \frac{\sup_{f\in\m F} G_f(n,\Delta)}{\sqrt{n}} + \frac{\m O}{k\sqrt{n}} \r) \Bigg].
\end{multline*}
Moreover, same bounds hold for the permutation-invariant estimators $\hL_U(f)$, up to the change in absolute constants.
\end{theorem}

\noindent An immediate corollary is the bound for the excess risk 
\begin{multline}
\label{eq:slow-rates1}
\m E(\wh f_N) \leq 
C
\Bigg[ \frac{\widetilde \Delta}{\Delta} \l( \mb E\sup_{f\in \m F} \frac{1}{N} \sum_{j=1}^N \l( \ell(f(X_j)) - P\,\ell(f) \r)
+ \sigma(\ell,\m F) \sqrt{\frac{s}{N}} \r)
\\
+ \widetilde \Delta \sqrt{n} \l( \frac{s}{N}
+ \frac{\sup_{f\in\m F} G_f(n,\Delta)}{n} + \frac{\m O}{N} \r) \Bigg]
\end{multline}
that holds under the assumptions of Theorem \ref{th:unif} with probability at least $1 - 2e^{-s}$. 
When the class $\l\{ \ell(f), \ f\in \m F\r\}$ is P-Donsker \cite{dudley2014uniform}, $\limsup\limits_{N\to\infty} \Big| \mb E\sup\limits_{f\in \m F} \frac{1}{\sqrt N} \sum_{j=1}^N \l( \ell(f(X_j)) - P\ell(f) \r) \Big|$ is bounded, hence condition \eqref{eq:assump} holds for $N$ large enough whenever $s$ is not too big and $\Delta$ and $k$ are not too small, namely, $s\leq c'k$ and $\Delta\sqrt{k} \geq c'' \sigma(\m F)$. 
The bound of Theorem \ref{th:unif} also suggests that the natural ``unit'' to measure the magnitude of parameter $\Delta$ is $\sigma(\ell,\m F)$. 
We will often use the ratio $M_\Delta:=\frac{\Delta}{\sigma(\ell,\m F)}$ that can be interpreted as a level of truncation expressed in the units of $\sigma(\ell,\m F)$, and is one of the two main quantities controlling the bias of the estimator $\hL(f)$, the second one being the subgroup size $n$.

To put these results in perspective, let us consider two examples. 
First, assume that $n=1$, $k=N$ and set $\Delta = \Delta(s):= \sigma(\m F)\sqrt{\frac{N}{s}}$ for $s\leq c' N$. Using Lemma \ref{lemma:BE-nonunif} with $\delta=0$ to estimate $G_f(n,\Delta)$, we deduce that 
\[
\m E(\wh f_N) \leq 
C
\Bigg[ \mb E\sup_{f\in \m F} \frac{1}{N} \sum_{j=1}^N \l( \ell(f(X_j)) - P \ell(f) \r)
+ \sigma(\ell,\m F) \l( \sqrt{\frac{s}{N}} + \frac{\m O}{\sqrt N}\r) \Bigg]
\] 
with probability at least $1-2e^{-s}$. 
This inequality improves upon excess risk bounds obtained for Catoni-type estimators in \cite{brownlees2015empirical}, as it does not require functions in $\m F$ to be uniformly bounded. 

The second case we consider is when $N\gg n\geq 2$. 
For the choice of $\Delta\asymp \sigma(\ell,\m F)$, the estimator $\hL(f)$ most closely resembles the median-of-means estimator. 
In this case, Theorem \ref{th:unif} yields the excess risk bound of the form
\[
\m E(\wh f_N) \leq 
C
\Bigg[ \mb E\sup_{f\in \m F} \frac{1}{N} \sum_{j=1}^N \l( \ell(f(X_j)) - P\ell(f) \r)
+ \sigma(\ell,\m F) \l(\sqrt{\frac{s}{N}} + \sqrt{\frac k N}\sup_{f\in \m F} G_f(n,\sigma(\m F))  + \frac{\m O}{k}\sqrt{\frac{k}{N}}\r) \Bigg]
\]
that holds with probability $\geq 1 - 2e^{-s}$ for all $s\leq c'k$.
As $\sup_{f\in \m F} G_f(n,\Delta)$ is small for large $n$ and $\frac{\m O}{k}\sqrt{\frac{k}{N}}\leq \sqrt{\frac{\m O}{N}}$ whenever $\m O\leq k$, this bound is improves upon Theorem 2 in \cite{lecue2018robust} that provides bounds for the excess risk for robust classifiers based on the the median-of-means estimators. 

\subsection{Towards fast rates for the excess risk.}
\label{section:fast-rates}

It is well known that in regression and binary classification problems, excess risk often converges to $0$ at a rate faster than $N^{-1/2}$, and could be as fast as $N^{-1}$. Such rates are often referred to as ``fast'' or ``optimistic'' rates. 
In particular, this is the case when there exists a ``link'' between the excess risk and the variance of the loss class, namely, if for some convex nondecreasing and nonnegative function $\phi$ such that $\phi(0)=0$, 
\[
\m E(f) = P\ell(f) - P\ell(f_\ast) \geq \phi\l( \sqrt{\var\l( \ell(f(X)) - \ell(f_\ast(X))\r)}\r).
\]  
It is thus natural to ask if fast rates can be attained by estimators produced by the ``robust'' algorithms proposed above. 
Results presented in this section give an affirmative answer to this question. 
Let us introduce the main quantities that appear in the excess risk bounds. For $\delta>0$, let 
\begin{align*}
\m F(\delta)&:=\l\{ \ell(f): \ f\in \m F, \ \m E(f) \leq \delta\r\}, 
\\
\nu(\delta)&:=\sup_{\ell(f)\in \m F(\delta)} \sqrt{\var\l( \ell(f(X)) - \ell(f_\ast(X)) \r)},
\\
\omega(\delta)&:=\mb E \sup_{\ell(f)\in \m F(\delta)} \l|  \frac{1}{\sqrt{N}}\sum_{j=1}^N \Big( (\ell(f) - \ell(f_\ast))(X_j) - P( \ell(f) - \ell(f_\ast)) \Big) \r|.
\end{align*}
Moreover, define 
\begin{align*}
\mathfrak{B}(\ell,\m F) &:= \frac{\sup_{f\in \m F}\mb E^{1/4}\l( \ell(f(X)) - \mb E\ell(f(X)) \r)^4}{\sigma(\ell, \m F)}.
\end{align*}
The following condition, known as \emph{Bernstein's condition} following \cite{bartlett2006empirical}, plays the crucial role in the analysis of excess risk bounds.  
\begin{assumption}
\label{ass:bern}
There exist constants $D>0$, $\delta_B>0$ such that   
\[
\var\l( \ell(f(X)) - \ell(f_\ast(X)) \r)\leq D^2 \, \m E(f)
\] 
whenever $\m E(f)\leq \delta_B$. 
\end{assumption}
\noindent Assumption \ref{ass:bern} is known to hold in many concrete cases of prediction and classification tasks, and we provide examples and references in Section \ref{sec:examples} below. Informally speaking, it postulates that any $f$ with small excess risk must be ``close'' to $f_\ast$. More general versions of the Bernstein's condition are often considered in the literature: for instance, it can be replaced by assumption \cite{bartlett2006empirical} requiring that $\var\l( \ell(f(X)) - \ell(f_\ast(X)) \r)\leq D^2 \, \l(\m E(f)\r)^{\tau}$ for some $\tau\in (0,1]$ (clearly, our assumption corresponds to $\tau=1$). Results of this paper admit straightforward extensions to the slightly less restrictive scenario when $\tau<1$; we omit the details to reduce the level of technical burden on the statements of our results.

Following \cite[][Chapter 4]{Koltchinskii2011Oracle-inequali00}, we will say the the function $\psi:\mb R_+\mapsto \mb R_+$ is of concave type if it is nondecreasing and $x\mapsto \frac{\psi(x)}{x}$ is decreasing. Moreover, if for some $\gamma\in(0,1)$ $x\mapsto \frac{\psi(x)}{x^\gamma}$ is decreasing, we will say that $\psi$ is of strictly concave type with exponent $\gamma$. 
We will assume that $\omega(\delta)$ admits an upper bound $\widetilde \omega(\delta)$ of strictly concave type (with some exponent $\gamma$), and that $\nu(\delta)$ admits an upper bound $\widetilde \nu(\delta)$ of concave type. 
For instance, when assumption \ref{ass:bern} holds, $\nu(\delta)\leq D \sqrt{\delta}$ for $\delta\leq \delta_B$, implying that $\widetilde \nu(\delta)=D\sqrt{\delta}$ is an upper bound for $\nu(\delta)$ of strictly concave type with $\gamma=\frac{1}{2}$. 
\footnote{this is only true in some neighborhood of $0$, but is sufficient for our purposes} 
Moreover, the function $\omega(\delta)$ often admits an upper bound of the form $\widetilde \omega(\delta) = R_1 + \sqrt{\delta}R_2$ where $R_1$ and $R_2$ do not depend on $\delta$; such an upper bound is also of concave type.  
Next, set 
\begin{align}
\label{eq:delta-bar}
\widebar \delta&:=\min \l\{ \delta>0: \ C_1(\rho)\frac{1}{\sqrt{N}} \frac{\widetilde \Delta}{\Delta}\frac{\widetilde \omega(\delta)}{\delta} \leq \frac{1}{7}\r\},
\end{align}
where $C_1(\rho)$ is a sufficiently large positive constant that depends only on $\rho$. This quantity plays an important role in controlling the excess risk, as shown by the following theorems.  
\begin{theorem}
\label{th:fast-rates-1}
Assume that conditions of Theorem \ref{th:unif} hold. Additionally, suppose that $M_\Delta:=\frac{\Delta}{\sigma(\ell,\m F)}\geq 1$. Then 
\[
\wh\delta_N \leq \widebar \delta + 
C(\rho)\l( D^2  \l(\frac{1}{M_\Delta^2 n} + \frac{s+\m O}{N}\r) + \sigma(\ell,\m F)\sqrt{n}M_\Delta \l( \frac{1}{M_\Delta^4 n} + \frac{s + \m O}{N}\r) \r).
\]
with probability at least $1-10e^{-s}$, where the constant $C(\rho)$ depends on $\rho$ only and $D$ is a constant appearing in Assumption \ref{ass:bern}. 
Moreover, same bound holds for $\wh\delta_N^U$, up to a change in absolute constants.
\end{theorem}
\noindent Under stronger moment assumptions, the excess risk bound can be strengthened and take the following form.                                                       
\begin{theorem}
\label{th:fast-rates-2}
Assume that conditions of Theorem \ref{th:unif} hold. Additionally, suppose that 
\[
\sup_{f\in \m F}\mb E^{1/4}\l( \ell(f(X)) - \mb E\ell(f(X)) \r)^4<\infty
\] 
and that $M_\Delta:=\frac{\Delta}{\sigma(\ell,\m F)}\geq 1$. Then 
\[
\wh\delta_N \leq \widebar \delta + 
C(\rho)\l( D^2 + \sigma(\ell,\m F)\sqrt{n}M_\Delta \r) \l(\frac{\mathfrak{B}^6(\ell,\m F)}{M_\Delta^4 n^2} +\frac{s+\m O}{N} \r).
\]
with probability at least $1-10e^{-s}$, where the constant $C(\rho)$ depends on $\rho$ only and $D$ is a constant appearing in Assumption \ref{ass:bern}. 
Moreover, same bound holds for $\wh\delta_N^U$, up to a change in absolute constants.
\end{theorem}
\begin{remark}
\mbox{ } 

\noindent \textbf{1.} It is evident that whenever $\m O = 0$, the best possible rates implied by Theorem \ref{th:fast-rates-1} are of order $N^{-2/3}$ (indeed, this is the case whenever $M_\Delta\sqrt{n}\asymp N^{1/3}$ and $\widebar \delta \lesssim N^{-2/3}$), while the best possible rates attained by Theorem \ref{th:fast-rates-2} are of order $N^{-3/4}$ (when $M_\Delta\sqrt{n}\asymp N^{1/4}$ and $\widebar \delta \lesssim N^{-3/4}$); in particular, in this case the choice of $M_\Delta$ and $n$ is independent of $\widebar\delta$. 
In general, if $\m O = \eps N$ for $\eps>0$, the best rates implied by Theorems \ref{th:fast-rates-1} and \ref{th:fast-rates-2} are $\widebar \delta + C(\m F,\rho,P)\eps^{-2/3}$ and $\widebar \delta + C(\m F,\rho,P)\eps^{-3/4}$ respectively.

\noindent \textbf{2. } Assumption requiring that $M_\Delta \geq 1$ is introduced for convenience: without it, extra powers of the ratio $\frac{\max\l(\Delta,\sigma(\ell,\m F)\r)}{\Delta}$ appear in the bounds.
\end{remark}

Our next goal is to describe an estimator that is capable of achieving excess risk rates up to $N^{-1}$.  
The approach that we follow is similar in spirit to the ``minmax'' estimators studied in \cite[][among others]{audibert2011robust,lerasle2011robust,lecue2017robust}, as well as the ``median-of-means tournaments'' introduced in \cite{lugosi2016risk}; all these methods focus on estimating the differences $\m L(f_1) - \m L(f_2)$ for all $f_1,f_2\in \m F$.  
Recall that $f_\ast = \argmin_{f\in \m F} P\ell(f)$, and observe that for any fixed $f'\in \m F$, $f_\ast$ can be equivalently defined via
\[
f_\ast = \argmin_{f\in \m F}\, P\l( \ell(f) - \ell(f') \r).
\]
A version of the robust empirical risk minimizer \eqref{eq:fhat} corresponding to this problem can be defined as 
\begin{equation}
\label{eq:diff}
\hL(f - f'):= \argmin_{y\in \mb R}\frac{1}{\sqrt{N}}\sum_{j=1}^k \rho\l(\sqrt{n}\,\frac{\l(\bL_j (f) - \bL_j(f')\r) - y}{\Delta}\r)
\end{equation}
for appropriately chose $\Delta>0$, and
\begin{equation*}
\wh f'_N := \argmin_{f\in \m F} \wh{\m L}^{(k)}(f - f').
\end{equation*}
Moreover, if $f'\in \m F$ is a priori known to be ``close'' to $f_\ast$, then it suffices to search for the minimizer in a neighborhood $\m F'$ of $f'$ that contains $f_\ast$ instead of all $f\in \m F$: 
\begin{equation*}
\wh f''_N := \argmin_{f\in \m F'} \wh{\m L}^{(k)}(f - f').
\end{equation*}
The advantage gained by this procedure is expressed by the fact that $\sup_{f\in \m F'} \var\l(\ell(f(X)) - \ell(f'(X)) \r)$ can be much smaller than $\sigma(\ell,\m F)$. 

We will now formalize this argument and provide performance guarantees; we use the framework of Theorem \ref{th:fast-rates-2} which leads to the bounds that are easier to state and interpret. However, similar reasoning applies to the setting of Theorem \ref{th:fast-rates-1} as well. Presented algorithms also admit straightforward permutation-invariant modifications that we omit. 
Let 
\[
\widehat{\m E}_N(f) := \hL(f) - \hL(\wh f_N)
\] 
be the ``empirical excess risk'' of $f$. Indeed, this is a meaningful notion as $\wh f_N$ is the minimizer of $\hL(f)$ over $f\in \m F$. Assume that the initial sample of size $N$ is split into two disjoint parts $S_1$ and $S_2$ of cardinalities that differ at most by 1:  $(X_1,Y_1),\ldots,(X_N,Y_N) = S_1 \cup S_2$. The algorithm proceeds in the following way:
\begin{enumerate}
\item Let $\wh f_{|S_1|}$ be the estimator \eqref{eq:fhat} evaluated over subsample $S_1$ of cardinality $|S_1|\geq \lfloor N/2\rfloor$, with the scale parameter $\Delta_1$ and the partition parameter $k_1$ corresponding the group size $n_1=\lfloor |S_1|/k_1 \rfloor$;
\item Let $\delta^{\prime} = \widebar \delta + 
C(\rho)\l( D^2 + \sigma(\ell,\m F)\sqrt{n}M_{\Delta_1} \r) \l(\frac{\mathfrak{B}^6(\ell,\m F)}{M_{\Delta_1}^4 n_1^2} +\frac{s+\m O}{N} \r)$ be a known upper bound on the excess risk in Theorem \ref{th:fast-rates-2} (while this condition is restrictive, it is similar to the requirements of existing approaches \cite{brownlees2015empirical,lugosi2016risk}; discussion of adaptation issues is beyond the scope of this paper and will be addressed elsewhere). Set 
\[
\widehat {\m F}(\delta^{\prime}) :=\l\{ f\in \m F: \ \widehat{\m E}_N(f) \leq \delta^{\prime}\r\}. 
\]
\item Define $\wh f''_N := \argmin_{f\in \widehat {\m F}(\delta^{\prime})} \wh{\m L}^{(k)}(f - \wh f_{|S_1|})$ where 
\begin{equation}
\label{eq:step2}
\hL\l(f - \wh f_{|S_1|}\r) = \argmin_{y\in \mb R} \sum_{j=1}^{k_2} \rho\l(\sqrt{n}\,\frac{\l(\bL_j (f) - \bL_j(\wh f_{|S_1|})\r) - y}{\Delta_2}\r)
\end{equation}
is based on the subsample $S_2$ of cardinality $|S_2|\geq \lfloor N/2\rfloor$, a scale parameter $\Delta_2$ and the partition parameter $k_2$ corresponding the group size $n_2=\lfloor |S_2|/k_2 \rfloor$.
\end{enumerate}
It will be demonstrated in the course of the proofs that on event of high probability, $\widehat{\m F}(\delta^{\prime})\subseteq \m F(c\delta^{\prime})$ for an absolute constant $c\leq 7$. Hence, on this event $\sup_{f\in \widehat{ \m F}(\delta^{\prime})} \var\l( \ell(f(X)) - \ell(f_\ast(X))\r) \leq \nu^2(c\delta^{\prime})\leq cD^2 \delta^{\prime}$ by the definition of $\nu(\delta)$ and Assumption \ref{ass:bern}, thus 
$\Delta_2 = D\,M_{\Delta_2}\sqrt{c\delta^{\prime}}$ with $M_{\Delta_2}\geq 1$ often leads to an estimator with improved performance. 
\begin{theorem}
\label{cor:fast}
Suppose that 
\[
\sup_{f\in \m F}\mb E^{1/4}\l( \ell(f(X)) - \mb E\ell(f(X)) \r)^4<\infty
\] 
and that $\Delta_1$, $\Delta_2$ satisfy $M_{\Delta_1}:=\frac{\Delta_1}{\sigma(\ell,\m F)}\geq 1$ and 
$M_{\Delta_2}:=\frac{\Delta_2}{D\sqrt{7\delta^{\prime}}}\geq 1$. 
Moreover, assume that for a sufficiently small absolute constant $c'>0$, 
$\sup_{f\in \m F} \max\l(G_f(n_1,\Delta_1), G_f(n_2,\Delta_2)\r)\leq c'$ and $\frac{s+O}{\min(k_1,k_2)}\leq c'$. Finally, we require that
\begin{align}
\label{eq:upp-b}
& \sqrt{k_1}M_{\Delta_1} \geq  \frac{c'}{\sigma(\ell,\m F)} \, \mb E\sup_{f\in \m F} \frac{1}{\sqrt{|S_1|}}\sum_{j=1}^{|S_1|} \l( \ell(f(X_j)) - P\,\ell(f) \r) \text{ and }
\\ \nonumber
& \sqrt{k_2} M_{\Delta_2} \geq c'\frac{\sqrt{N\delta^{\prime}}}{D}.
\end{align}
Then
\[
\m E\l( \widehat f''_N\r) \leq \widebar \delta + 
C(\rho)\l( D^2 + D\sqrt{\delta^{\prime}} \sqrt{n} M_{\Delta_2} \r) \l(\frac{\mathfrak{B}^6(\ell,\m F)}{M_{\Delta_2}^4 n^2} +\frac{s+\m O}{N} \r)
\]
with probability at least $1-20e^{-s}$, where $C(\rho)$ depends on $\rho$ only and $D$ is the constant appearing in Assumption \ref{ass:bern}.
\end{theorem}

\noindent The statement of Theorem \ref{cor:fast} is technical, so let us try to distill the main ideas. 
The key difference between Theorem \ref{th:fast-rates-2} and Theorem \ref{cor:fast} is that the ``remainder term'' 
\[
\sigma(\ell,\m F) \sqrt{n} M_\Delta \l(\frac{\mathfrak{B}^6(\ell,\m F)}{M_\Delta^4 n^2} +\frac{s+\m O}{N} \r)
\] 
is replaced by a potentially much smaller quantity $\sqrt{\delta^{\prime}} \sqrt{n} M_{\Delta} \l(\frac{\mathfrak{B}^6(\ell,\m F)}{M_\Delta^4 n^2} +\frac{s+\m O}{N} \r)$. In particular, if $\delta^{\prime} \ll \l(n M_{\Delta}^2\r)^{-1}$, this term often becomes negligible. 
To be more specific, assume that $\bar\delta = \frac{C(\m F)}{\sqrt{N}} \cdot h(N)$ where $h(N)\to 0$ as $N\to \infty$ (meaning that fast rates are achievable) and that $\m O = \eps \, N$ for $\eps\geq \frac{1}{N}$.
Moreover, suppose that $\mathfrak{B}(\ell,\m F)$ is bounded above by a constant. 
If $\Delta_1$ is chosen such that $\Delta_1\asymp \sigma(\ell,\m F)$, then 
$\delta^{\prime} = C\l( \widebar \delta + \sigma(\ell,\m F)\l( \l( \frac{k}{N} \r)^{3/2} + \frac{s+\m O}{\sqrt{kN}}\r)\r)$. 
Hence, if $\max\l( h(N)\sqrt{N}, N\eps^{2/3}\r)\ll k_j \leq C N\sqrt{\eps}$ for $j=1,2$ 
and $\Delta_2\asymp \sqrt{\delta^{\prime}}$, then
\[
\delta^{\prime} \cdot n M^2_{\Delta_2} = O(1),
\]
and the excess risk of $ \widehat f''_N$ admits the bound
\[
\m E\l( \widehat f''_N\r)\leq \widebar \delta + C(\rho,D)\l(\eps + \frac{s}{N}\r)
\]
that holds with probability at least $1-Ce^{-s}$. A possible choice satisfying all the required conditions is $k_j \asymp N\sqrt{\eps}, \ j=1,2$ (indeed, it this case it is straightforward to check that conditions \eqref{eq:upp-b} hold for sufficiently large $N$ as $k_j\gtrsim \sqrt{N}, \ j=1,2$). Analysis of the case when $\m O=0$ follows similar steps, with several simplifications. 
 
\section{Examples.}
\label{sec:examples}

We consider two common prediction problems, regression and binary classification, and discuss the implications of our main results for these problems. 

\subsection{Binary classification with convex surrogate loss.}

The key elements of the binary classification framework were outlined in Section \ref{sec:numerical}. 
Here, we recall few popular examples of classification-calibrated losses and present conditions that are sufficient for the Assumption \ref{ass:bern} to hold. 
\begin{description}
\item[Logistic loss] $\ell(yf(z)) = \log\l(1 + e^{-yf(z)}\r)$. Consider two scenarios: 
\begin{enumerate}
\item Uniformly bounded classes, meaning that for all $f\in \m F$, $\sup_{z\in S} |f(z)|\leq B$. In this case, Assumption \ref{ass:bern} holds with $D=2e^{B}$ for all $f\in \m F$. See \cite{bartlett2004large} and Proposition 6.1 in \cite{alquier2017estimation}. 
\item Linear separators and Gaussian design: in this case, we assume that $S=\mb R^d$, $Z\sim N(0,I)$ is Gaussian, and $\m F=\l\{ \dotp{\cdot}{v}: \ \|v\|_2\leq R\r\}$ is a class of linear functions. In this case, according to the Proposition 6.2 in \cite{alquier2017estimation}, Bernstein's assumption is satisfied with $D=cR^{3/2}$ for some absolute constant $c>0$.  
\end{enumerate}
\item[Hinge loss] $\ell(yf(z)) = \max\l( 0, 1 - yf(z)\r)$. In this case, sufficient condition for Assumption \ref{ass:bern} to hold is the following: there exists $\tau>0$ such that $|g_\ast(Z)|\geq \tau$ almost surely. It follows from Proposition 1 in \cite{lecue2007optimal} (see also \cite{tsybakov2004optimal}) that Assumption \ref{ass:bern} holds with $D = \frac{1}{\sqrt{2\tau}}$ in this case.
\end{description}

\noindent \textbf{Bound for $\widebar \delta$.} Let $\Pi$ stand for the marginal distribution of $Z$ and recall that 
\[
\omega(\delta):=\mb E \sup_{\ell(f)\in \m F(\delta)} \l|  \frac{1}{\sqrt{N}}\sum_{j=1}^N \Big( (\ell(Y_j f(Z_j)) - \ell(Y_j f_\ast(Z_j))) - \mb E( \ell(Yf(Z)) - \ell(Y f_\ast(Z))) \Big) \r|.
\]
Since $\ell$ is Lipchitz continuous by assumption (with Lipschitz constant denoted $L(\ell)$), consequent application of symmetrization and Talagrand's contraction inequalities \cite{LT-book-1991,van2016estimation} yields that 
\[
\omega(\delta) \leq 4L(\ell)\,\mb E\sup_{\|f - f_\ast\|_{L_2(\Pi)}\leq D\sqrt{\delta}}\l| \frac{1}{\sqrt N}\sum_{j=1}^N \eps_j (f - f_\ast)(Z_j)\r|
\]
where $\eps_1,\ldots,\eps_N$ are i.i.d. random signs independent from $Y_j$'s and $Z_j$'s. 
The latter quantity is the modulus of continuity of a Rademacher process, and various upper bounds for it are well known. 
For instance, if $\m F$ is a subset of a linear space of dimension $d$, then, according to Proposition 3.2 in \cite{Koltchinskii2011Oracle-inequali00}, 
$\mb E\sup_{\|f - f_\ast\|_{L_2(\Pi)}\leq D\sqrt{\delta}}\l| \frac{1}{\sqrt N}\sum_{j=1}^N \eps_j (f - f_\ast)(Z_j)\r| \leq D\sqrt{\delta}\sqrt{d}$, whence $\widetilde\omega(\delta):=4D\,L(\ell)\sqrt{\delta d}$ is an upper bound for $\omega(\delta)$ and is of concave type, implying that 
\[
\widebar \delta \leq C(\rho,\ell) D^2 \, \frac{d}{N}.
\] 
More generally, assume that the class $\m F$ has a measurable envelope $F(z):=\sup_{f\in \m F}|f(z)|$ that satisfies $\|F(Z)\|_{\psi_2}<\infty$, 
where $\|\xi\|_{\psi_2}:=\inf\l\{ C>0: \ \mb E\exp\l( |\xi/C|^2 \r)\leq 2 \r\}$ is the $\psi_2$ (Orlicz) norm. 
Moreover, suppose that the covering numbers $N\l( \m F, Q,\eps \r)$ of the class $\m F$ with respect to the norm $L_2(Q)$ satisfy the bound 
\begin{equation}
\label{eq:cov-number}
N\l( \m F, Q,\eps \r)\leq \l( \frac{A \|F\|_{L_2(Q)}}{\eps}\r)^V 
\end{equation}
for some constants $A\geq 1$, $V\geq 1$, all $0<\eps\leq 2\|F\|_{L_2(Q)}$ and all probability measures $Q$. For instance, VC-subgraph classes are known to satisfy this bound with $V$ being the VC dimension of $\m F$ \cite{wellner2013weak,Koltchinskii2011Oracle-inequali00}. 
In this case, it is not difficult to show (see for example the proof of Lemma \ref{lemma:quadratic} in the appendix) that
\begin{multline*}
\mb E\sup_{\|f - f_\ast\|_{L_2(\Pi)}\leq D\sqrt{\delta}}\l| \frac{1}{\sqrt N}\sum_{j=1}^N \eps_j (f - f_\ast)(Z_j)\r|
\\
\leq \widetilde \omega(\delta):=C\sqrt{V\log(e^2A^2 N)} \l( \sqrt{\delta} + \sqrt{\frac{V}{N}}\log(A^2 N)\| F \|_{\psi_2} \r),
\end{multline*}
hence it is easy to check that in this case 
\[
\widebar\delta \leq C(\rho) \frac{V \log^{3/2}(e^2 A^2 N) \| F \|_{\psi_2}}{N}.
\]
It immediately follows from the discussion following Theorem \ref{cor:fast} that the excess risk of the estimator $\wh f''_N$ satisfies
\[
\m E\l( \widehat f''_N\r)\leq C(\rho,D)\l(\frac{\m O}{N} +  \frac{V\log^{3/2}(e^2 A^2 N)\| F \|_{\psi_2}  + s}{N} \r)
\]
with probability at least $1-20e^{-s}$. Similar results hold for regression problems with Lipschitz losses, such as Huber's loss or quantile loss \cite{alquier2017estimation}. 

\subsection{Regression with quadratic loss.}

Let $X=(Z,Y)\in S\times \mb R$ be a random couple with distribution $P$ satisfying $Y = f_\ast(Z) + \eta$ where the noise variable $\eta$ is independent of $Z$ and $f_\ast(z)=\mb E[Y|Z=z]$ is the regression function. 
Let $\|\eta\|_{2,1}:=\int_0^\infty \sqrt{\pr{|\eta|>t}} dt$, and observe that $\|\eta\|_{2,1}<\infty$ as $\sup_{f\in \m F}\mb E(Y - f(Z))^4 < \infty$ by assumption. As before, $\Pi$ will stand for the marginal distribution of $Z$. 
Let $\m F$ be a given convex class of functions mapping $S$ to $\mb R$ and such that the regression function $f_\ast$ belongs to $\m F$, so that
\[
f_\ast =\argmin_{f\in \m F}\mb E \l( Y - f(Z)\r)^2.
\]
In this case, the natural choice for the loss function is the quadratic loss $\ell(x) = x^2$ which is not Lipschitz continuous on unbounded domains. Assume that the class $\m F$ has a measurable envelope $F(z):=\sup_{f\in \m F}|f(z)|$ that satisfies $\|F(Z)\|_{\psi_2}<\infty$. 
Moreover, suppose that the covering numbers \footnote{Definition..}$N\l( \m F, Q,\eps \r)$ of the class $\m F$ with respect to the norm $L_2(Q)$ satisfy the bound 
\begin{equation}
\label{eq:cov-number}
N\l( \m F, Q,\eps \r)\leq \l( \frac{A \|F\|_{L_2(Q)}}{\eps}\r)^V 
\end{equation}
for some constants $A\geq 1$, $V\geq 1$, all $0<\eps\leq 2\|F\|_{L_2(Q)}$, and all probability measures $Q$. For instance, VC-subgraph classes are known to satisfy this bound with $V$ being the VC dimension of $\m F$ \cite{wellner2013weak,Koltchinskii2011Oracle-inequali00}. 

\noindent \textbf{Bernstein's assumption.} 
It follows from Lemma 5.1 in \cite{Koltchinskii2011Oracle-inequali00} that 
\[
\m F(\delta)\subseteq \l\{ (y - f(z))^2: \ f \in \m F, \  \mb E(f(Z) - f_\ast(Z))^2\leq 2\delta \r\},
\]
hence $\nu(\delta)\leq \sqrt{2\delta}$ so $D$ can be taken to be $\sqrt{2}$ in Assumption \ref{ass:bern}. 

\noindent \textbf{Bound for $\widebar \delta$.} 
Required estimates follow from the following lemma:
\begin{lemma}
\label{lemma:quadratic}
Under the assumptions made in this section and for $\Delta\geq \sigma(\ell,\m F)$,
\[
\bar \delta\leq C(\rho) \frac{V\log^2(A^2 N)(\| F \|^2_{\psi_2} + \|\eta\|^2_{2,1})}{N}.
\]
\end{lemma}
The proof is given in the appendix. 
An immediate corollary of the lemma, according to the discussion following Theorem \ref{cor:fast}, is that the excess risk of the estimator $\wh f''_N$ satisfies the inequality
\[
\m E\l( \widehat f''_N\r)\leq C(\rho,D)\l(\frac{\m O}{N} +  \frac{V\log^2(A^2 N)(\| F \|^2_{\psi_2} + \|\eta\|^2_{2,1}) + s}{N} \r)
\]
with probability at least $1-20e^{-s}$, for $0< s\leq cN^{1/4}$.

\section{Proofs of the main results.}
\label{section:proofs}

In the proofs of the main results, we will rely on the following convenient change of variables. Denote 
\begin{align*}
\hGk(z;f) &= \frac{1}{\sqrt{k}}\sum_{j=1}^k \rho'\l( \sqrt{n}\,\frac{ (\bL_j(f) - \m L(f) ) - z}{\Delta} \r),
\\
\Gk(z;f) &= \sqrt{k} \, \mb E \rho'\l( \sqrt{n}\,\frac{ (\bL_j(f) - \m L(f) ) - z}{\Delta} \r).
\end{align*}
In particular, when $\m O = 0$, $\Gk(z;f) = \mb E \hGk(z;f)$. 
Let $\hEk(f)$ and $\Ek(f)$ be defined by the equations
\begin{align}
\label{eq:theta_0}
{\widehat G}_k\l(\hEk(f);f\r) &= 0,
\\
\nonumber
G_k\l(\Ek(f);f\r) &= 0.
\end{align} 
Comparing this to the definition of $\hL(f)$ \eqref{eq:M-est}, it is easy to see that $\hEk(f) =\hL(f) - \m L(f)$. 
Hence $\Ek(f)$, the ``population version'' of $\hEk(f)$, is a natural measure of bias of the estimator $\hL(f)$. 

\subsection{Technical tools.}

We summarize the key results that our proofs rely on.

\begin{lemma}
\label{lemma:variance}
Let $\rho$ satisfy Assumption \ref{ass:1}. Then for any random variable $Y$ with $\mb EY^2<\infty$, 
\[
\var\l(\rho'(Y)\r) \leq \var\l( Y \r).
\]
\end{lemma}
\begin{proof}
See Lemma 5.3 in \cite{minsker2018uniform}.
\end{proof}

\begin{lemma}
\label{lemma:lindeberg}
For any function $h$ of with bounded third derivative and a sequence of i.i.d. random variables $\xi_1,\ldots,\xi_n$ such that $\mb E\xi_1=0$ and $\mb E|\xi_1|^3<\infty$, 
\[
\l| \mb E h\l( \sum_{j=1}^n \xi_j \r) - \mb E h\l( \sum_{j=1}^n Z_j \r) \r|\leq C n\,\|h'''\|_\infty \,\mb E|\xi_1|^3,
\]
where $C>0$ is an absolute constant and $Z_1,\ldots,Z_n$ are i.i.d. centered normal random variables such that $\var(Z_1)=\var(\xi_1)$.
\end{lemma}
\begin{proof}
This bound follows from a standard application of Lindeberg's replacement method; see \cite[][chapter 11]{o2014analysis}. 
\end{proof}

\begin{lemma}
\label{lemma:step2}
Assume that $\mb E|f(X) - \mb Ef(X)|^{2}<\infty$ for all $f\in \m F$ and that $\rho$ satisfies Assumption \ref{ass:1}. Then for all $f\in \m F$ and $z\in \mb R$ satisfying $|z|\leq \frac{\Delta}{\sqrt{n}}\frac{1}{2}$,
\[
\l|\mb E\rho'\l(\sqrt{n} \frac{(\bar \theta_j(f) - Pf) - z}{\Delta} \r) - 
\mb E \rho'\l( \frac{W(f) - \sqrt{n} z }{\Delta} \r) \r|  \leq 
2 \,G_f(n,\Delta). 
\]
\end{lemma}
\begin{proof}
See Lemma 4.2 in \cite{minsker2018uniform}.
\end{proof}

\noindent Given $N$ i.i.d. random variables $X_1,\ldots,X_N\in \m S$, let $\| f - g\|_{L_\infty(\Pi_N)}:=\max_{1\leq j \leq N} |f(X_j) - g(X_j)|$. Moreover, define 
\[
\Gamma_{n,\infty}(\m F):=\mb E \gamma_2^2(\m F; L_\infty(\Pi_N)),
\] 
where $\gamma_2(\m F,L_\infty(\Pi_N))$ is Talagrand's generic chaining complexity \cite{talagrand2014upper}. 
\begin{lemma}
\label{lemma:chaining1}
Let $\sigma^2:=\sup_{f\in \m G}\mb E f^2(X)$. Then there exists a universal constant $C>0$ such that 
\[
\mb E\sup_{f\in \m F} \l| \frac{1}{N}\sum_{j=1}^N f^2(X_j) - \mb E f^2(X)\r| \leq C\l( \sigma\sqrt{\frac{\Gamma_{N,\infty}(\m F)}{N}} \bigvee \frac{\Gamma_{N,\infty}(\m F)}{N}\r).
\]
\end{lemma}
\begin{proof}
See Theorem 3.16 in \cite{Koltchinskii2011Oracle-inequali00}.
\end{proof}

\noindent The following form of Talagrand's concentration inequality is due to Klein and Rio (see section 12.5 in \cite{boucheron2013concentration}). 
\begin{lemma}
\label{lemma:klein}
Let $\{Z_j(f), \ f\in \m F\}, \ j=1,\ldots,N$ be independent (not necessarily identically distributed) separable stochastic processes indexed by class $\m F$ and such that $|Z_j(f)-\mb EZ_j(f)|\leq M$ a.s. for all $1\leq j\leq N$ and $f\in \m F$. Then the following inequality holds with probability at least $1-e^{-s}$:
\begin{align}
\label{eq:klein1}
\sup_{f\in \m F}\l( \sum_{j=1}^N (Z_j(f) - \mb EZ_j(f)) \r) &\leq 2\mb E\sup_{f\in \m F} \l( \sum_{j=1}^N (Z_j(f) - \mb EZ_j(f))\r) 
+V(\m F)\sqrt{2s}+\frac{4Ms}{3},
\end{align}
where $V^2(\m F)=\sup_{f\in \m F}\sum_{j=1}^N \var\l( Z_j(f)\r)$.
\end{lemma}
\noindent It is easy to see, applying \eqref{eq:klein1} to processes $\{-Z_j(f), \ f\in \m F\}$, that
\begin{equation}
\label{eq:klein2}
\inf_{f\in \m F}\l( \sum_{j=1}^N (Z_j(f) - \mb EZ_j(f))\r) \geq -2\mb E\sup_{f\in \m F} \l( \sum_{j=1}^N (\mb EZ_j(f) - Z_j(f))\r) 
 - V(\m F)\sqrt{2s} - \frac{4Ms}{3}
\end{equation}
with probability at least $1-e^{-s}$. 
Next, we describe the tools necessary to extend these concentration inequalities to nondegenerate U-statistics. 
Deviation inequality \eqref{eq:klein1} is a corollary of the following bound for the moment generating function (section 12.5 in \cite{boucheron2013concentration}): 
\begin{equation}
\label{eq:mgf}
\log \mb E e^{\lambda \l( \sum_{j=1}^N (Z_j(f) - \mb EZ_j(f))\r)}\leq \frac{e^{\lambda M} - \lambda M - 1}{M^2}\l( V^2(\m F) + 2M \,\mb E\sup_{f\in \m F} \l( \sum_{j=1}^N (Z_j(f) - \mb E Z_j(f))\r)\r)
\end{equation}
that holds for all $\lambda>0$. 
We use this fact to demonstrate a straightforward extension of Lemma \ref{lemma:klein} to the case of U-statistics. 
Let $\pi_N$ be the collection of all permutations $\pi:\{1,\ldots,N\}\mapsto \{1,\ldots,N\}$. 
Given $(i_1,\ldots,i_N)\in \pi_N$ and a U-statistic $U_{N,n}$ with kernel $h$ defined in \eqref{u-stat}, let 
\begin{equation*}
T_{i_1,\ldots,i_N}:=\frac{1}{k}\l( h\l(X_{i_1},\ldots,X_{i_n} \r) + h\l(X_{i_{n+1}},\ldots,X_{i_{2n}}\r) + \ldots + 
h\l(X_{i_{(k-1)n+1}},\ldots,X_{i_{kn}} \r) \r).
\end{equation*}
It is well known (e.g., see section 5 in \cite{hoeffding1963probability}) that the following representation holds:
\begin{equation}
\label{eq:U-decomp}
U_{N,n} = \frac{1}{N!}\sum_{(i_1,\ldots,i_N) \in \pi_N}  T_{i_1,\ldots,i_N}.
\end{equation}
Let $U_{N,n}'(z;f) = \frac{1}{{N\choose n}} \sum_{J\in \m A_N^{(n)}} \rho'\l(\sqrt{n}\,\frac{(\bL(f;J) - \mb E\ell(f(X))) - z}{\Delta}\r)$. 
Applied to $U'_{N,n}(z;f)$, relation \eqref{eq:U-decomp} yields that 
\[
U'_{N,n}(z;f) = \frac{1}{N!}\sum_{(i_1,\ldots,i_N)\in \pi_N} T_{i_1,\ldots,i_N}(z;f),
\]
where 
\begin{multline*}
T_{i_1,\ldots,i_N}(z;f) = \frac{1}{k}\Big(  \rho'\l(\sqrt{n}\,\frac{\bL(f;\{i_1,\ldots,i_n\}) - \mb E\ell(f(X)) - z}{\Delta}\r) +
\\ 
\ldots +  \rho'\l(\sqrt{n}\,\frac{\bL(f;\{i_{(k-1)n+1},\ldots,i_{kn}\}) - \mb E \ell(f(X)) - z}{\Delta}\r) \Big).
\end{multline*}
Jensen's inequality implies that for any $\lambda>0$,
\begin{multline*}
\mb E \exp\l( \frac{\lambda}{N!} \sum_{(i_1,\ldots,i_N)\in \pi_N} \l( T_{i_1,\ldots,i_N}(z;f) -\mb E T_{i_1,\ldots,i_N}(z;f)\r)\r) 
\\
\leq \frac{1}{N!} \sum_{(i_1,\ldots,i_N)\in \pi_N} \mb E \exp\Big( \lambda \l( T_{1,\ldots,N}(z;f) -\mb E T_{1,\ldots,N}(z;f)\r) \Big),
\end{multline*}
hence bound \eqref{eq:mgf} can be applied and yields that 
\begin{multline}
\label{eq:U-concentr}
\sup_{f\in \m F}\l( U'_{N,n}(z;f) - \mb E U'_{N,n}(z;f)\r) \leq 2\mb E\sup_{f\in \m F}\l( T_{1,\ldots,N}(z;f) - \mb E T_{1,\ldots,N}(z;f) \r) 
\\
+ \sup_{f\in \m F}\sqrt{\var\l( \rho'\l(\sqrt{n}\,\frac{\bar \theta(f;\{1,\ldots,n\}) - Pf - z}{\Delta}\r)\r)}\sqrt{\frac{2s}{k}} + \frac{8s\|\rho'\|_\infty}{3k}
\end{multline}
with probability at least $1-e^{-s}$. The expression can be further simplified by noticing that $\|\rho'\|_\infty\leq 2$ and that  
\[
\var\l( \rho'\l(\sqrt{n}\,\frac{\bar \theta(f;\{1,\ldots,n\}) - Pf - z}{\Delta}\r)\r) \leq \frac{\sigma^2(f)}{\Delta^2}.
\]
due to Lemma \ref{lemma:variance}.

\subsection{Proof of Theorems \ref{th:fast-rates-1} and \ref{th:fast-rates-2}.}
\label{proof:fast-rates}

We will provide detailed proofs for the estimator $\wh f_N$ that is based on disjoint groups $G_1,\ldots,G_k$. The bounds for its permutation-invariant version $\wh f^U_N$ follow exactly the same steps where all applications of the Talagrand's concentration inequality (Lemma \ref{lemma:klein}) are replaced by its version for nondegenerate U-statistics \eqref{eq:U-concentr}.

Let $J\subset \{1,\ldots,k\}$ of cardinality $|J|\geq k-\m O$ be the set containing all $j$ such that the subsample $\{ X_i, \ i\in G_j\}$ does not include outliers. Clearly, $\{ X_i: \ i\in G_j, \ j\in J\}$ are still i.i.d. as the partitioning scheme is independent of the data. 
Moreover, set $N_J:=\sum_{j\in J} |G_j|$, and note that, since $\m O<k/2$, 
\[
N_J\geq n |J|\geq \frac{N}{2}.
\] 
Consider stochastic process $R_N(f)$ defined as
\begin{equation}
\label{eq:R_N}
R_N(f) = \hGk\l(0;f \r) + \partial_z \Gk\l(0 ; f \r) \cdot \hEk(f),
\end{equation}
where $\partial_z \Gk\l(0; f \r) := \partial_z \Gk\l(z; f \r)_{|_{z=0}}$. 
Whenever $\partial_z \Gk\l(0; f \r)\ne 0$ (this assumption will be justified by Lemma \ref{lemma:derivative} below), we can solve \eqref{eq:R_N} for $\hEk(f)$ to obtain
\begin{equation}
\label{eq:risk-lin}
\hEk(f) =  -\frac{\hGk\l(0;f \r)}{\partial_z \Gk\l(0;f \r)} 
+ \frac{R_N(f)}{\partial_z \Gk\l(0; f \r)},
\end{equation}
which can be viewed as a Bahadur-type representation of $\hEk(f)$. 
Setting $f:=\wh f_N$ and recalling that $\hEk(f) =\hL(f) - \m L(f)$, we deduce that 
\begin{equation*}
\hL(\wh f_N) = \m L(\wh f_N) - \frac{\hGk\l( 0; \wh f_N \r)}{\partial_z \Gk\l(0; \wh f_N \r)} 
+ \frac{R_N(\wh f_N)}{\partial_z \Gk\l(0; \wh f_N \r)}.
\end{equation*}
By the definition \eqref{eq:fhat} of $\wh f_N$, $\hL(\wh f_N) \leq \hL(f_\ast)$, hence 
\begin{equation*}
\m L(\wh f_N) - \frac{\hGk\l( 0; \wh f_N \r)}{\partial_z \Gk\l(0; \wh f_N \r)} 
+ \frac{R_N(\wh f_N)}{\partial_z \Gk\l(0; \wh f_N \r)}
\leq \m L(f_\ast) - \frac{\hGk\l( 0;f_\ast \r)}{\partial_z \Gk\l( 0;f_\ast \r)} 
+ \frac{R_N(f_\ast)}{\partial_z \Gk\l( 0 ; f_\ast \r)}.
\end{equation*}
Rearranging the terms, it is easy to see that  
\begin{equation}
\label{eq:excess-risk}
\wh\delta_N = \m L(\wh f_N) -  \m L(f_\ast) \leq
\l| \frac{\hGk\l(0;\wh f_N \r)}{\partial_z G\l(0; \wh f_N \r)}  - \frac{\hGk\l( 0 ;f_\ast \r)}{\partial_z G\l(0; f_\ast \r)} \r|
+ 2\sup_{f\in \m F(\wh\delta_N)} \l|  \frac{R_N(f)}{\partial_z \Gk\l(0; f \r)} \r|.
\end{equation}
\begin{remark}
\label{remark:emp-risk}
Similar argument also implies, in view of the inequality $\m L(f_\ast)\leq \m L(\wh f_N)$, that
\begin{equation*}
\hL(f_\ast) + \frac{\hGk\l( 0;f_\ast \r)}{\partial_z \Gk\l( 0;f_\ast \r)} 
- \frac{R_N(f_\ast)}{\partial_z \Gk\l( 0 ; f_\ast \r)}
\leq \hL(\wh f_N) + \frac{\hGk\l( 0; \wh f_N \r)}{\partial_z \Gk\l(0; \wh f_N \r)} 
- \frac{R_N(\wh f_N)}{\partial_z \Gk\l(0; \wh f_N \r)},
\end{equation*}
hence 
\begin{equation*}
\hL(f_\ast) - \hL(\wh f_N) \leq \l| \frac{\hGk\l(0;\wh f_N \r)}{\partial_z G\l(0; \wh f_N \r)}  - \frac{\hGk\l( 0 ;f_\ast \r)}{\partial_z G\l(0; f_\ast \r)} \r|
+ 2\sup_{f\in \m F(\wh\delta_N)} \l|  \frac{R_N(f)}{\partial_z \Gk\l(0; f \r)} \r|.
\end{equation*}
\end{remark}

It follows from \eqref{eq:excess-risk} that in order to estimate the excess risk of $\wh f_N$, it suffices to obtain the upper bounds for 
\begin{equation}
\label{eq:A1}
A_1:=\l| \frac{\hGk\l(0;\wh f_N \r)}{\partial_z \Gk\l(0; \wh f_N \r)}  - \frac{\hGk\l( 0;f_\ast \r)}{\partial_z \Gk\l(0; f_\ast \r)} \r|
\end{equation} 
and 
\begin{equation}
\label{eq:A2}
A_2:=\sup_{f\in \m F(\wh \delta_N)} \l|  \frac{R_N(f)}{\partial_z \Gk\l(0; f \r)} \r|.
\end{equation}
Observe that
\begin{multline*}
\frac{\hGk\l(0;\wh f_N \r)}{\partial_z \Gk\l(0; \wh f_N \r)}  - \frac{\hGk\l( 0;f_\ast \r)}{\partial_z \Gk\l(0; f_\ast \r)}
\\
= \frac{\hGk\l(0;\wh f_N \r) - \hGk\l(0;f_\ast \r)}{\partial_z \Gk\l(0; \wh f_N \r)} 
+ \frac{\hGk\l( 0; f_\ast \r)}{\partial_z \Gk\l(0; f_\ast \r)\partial_z \Gk\l(0; \wh f_N \r)}
\l(\partial_z \Gk\l(0; f_\ast \r) - \partial_z \Gk\l(0; \wh f_N \r) \r).
\end{multline*}
Since $\rho''$ is Lipschitz continuous by assumption, 
\begin{multline}
\label{eq:A-1bound}
\l| \frac{\hGk\l( 0; f_\ast \r)}{\partial_z \Gk\l(0; f_\ast \r)\partial_z \Gk\l(0; \wh f_N \r)}
\l(\partial_z \Gk\l(0; f_\ast \r) - \partial_z \Gk\l(0; \wh f_N \r) \r) \r| 
\\
= \l| \frac{\hGk\l( 0; f_\ast \r) }{\partial_z \Gk\l(0; f_\ast \r)\partial_z \Gk\l(0; \wh f_N \r) } \frac{\sqrt{nk}}{\Delta} 
\mb E\l( \rho''\l( \sqrt{n}\frac{ \bL_1(f_\ast) - \m L(f_\ast) }{\Delta} \r) - \rho''\l(  \sqrt{n}\frac{ \bL_1(\wh f_N) - \m L(\wh f_N) }{\Delta}  \r) \r) \r| 
\\
\leq L(\rho'')\l| \frac{\hGk\l( 0; f_\ast \r) }{\partial_z \Gk\l(0; f_\ast \r)\partial_z \Gk\l(0; \wh f_N \r) } \frac{\sqrt{nk}}{\Delta^2} 
\var^{1/2}\l( \ell(\wh f_N(X)) - \ell(f_\ast(X)) \r) \r| 
\\
= C(\rho)\l| \frac{\hGk\l( 0; f_\ast \r) }{\partial_z \Gk\l(0; f_\ast \r)\partial_z \Gk\l(0; \wh f_N \r) }\r| \frac{\sqrt{nk}}{\Delta^2} 
\nu(\wh \delta_N).
\end{multline}
We following two lemmas are required to proceed.
\begin{lemma}
\label{lemma:derivative}
There exist $C(\rho)>0$ such that for any $f\in \m F$,
\[
\l| \partial_z \Gk\l(0;f \r) \r| \geq \frac{\sqrt{kn}}{\Delta} \l( \min\l(\frac{\Delta}{\sqrt{\var\l( \ell(f(X))\r)}}, 2\sqrt{\log 2}\r) - \frac{C(\rho)}{\sqrt{n}} \mb E \l|\frac{ \ell(f(X)) - P\ell(f)}{\Delta}\r|^3\r).
\]
\end{lemma}
\begin{proof}
See section \ref{proof:derivative}.
\end{proof}

\noindent In particular, the first bound of Lemma \ref{lemma:derivative} implies that for $n$ large enough,
\begin{equation}
\label{eq:derivative-lower}
\inf_{f\in \m F}\l| \partial_z \Gk\l(0;f \r) \r| \geq \frac{1}{2}\frac{\sqrt{kn}}{\max\l(\Delta,\sigma(\ell,\m F)\r)} = \frac{1}{2}\frac{\sqrt{kn}}{\widetilde \Delta }.
\end{equation}
It is also easy to deduce from the proof of Lemma \ref{lemma:derivative} that for small $n$ and $\Delta>\sigma(\ell,\m F)$, 
$\inf_{f\in \m F}\l| \partial_z \Gk\l(0;f \r) \r| \geq c(\rho)\frac{\sqrt{kn}}{\Delta}$ for some positive $c(\rho)$. 
\begin{lemma}
\label{lemma:hGk}
For any $f\in \m F$, 
\[
\hGk\l( 0; f \r) \leq 2\l( \sqrt{k} \,G_f(n,\Delta) + \frac{\sigma(\ell,f)}{\Delta}\sqrt{s} + \frac{2s}{\sqrt k} + \frac{\m O}{\sqrt{k}}\r)
\]
with probability at least $1 - 2e^{-s}$, where $C>0$ is an absolute constant.
\end{lemma}
\begin{proof}
See section \ref{proof:hGk}. 
\end{proof}
\noindent Lemma \ref{lemma:hGk} and \eqref{eq:derivative-lower} imply, together with \eqref{eq:A-1bound}, that 
\begin{multline}
\label{eq:final-1}
\l| \frac{\hGk\l( 0; f_\ast \r)}{\partial_z \Gk\l(0; f_\ast \r)\partial_z \Gk\l(0; \wh f_N \r)}
\l(\partial_z \Gk\l(0; f_\ast \r) - \partial_z \Gk\l(0; \wh f_N \r) \r) \r| 
\\
\leq C(\rho) \frac{\widetilde \Delta^2}{\Delta^2} \l( \frac{\sigma(\ell,f_\ast)}{\Delta}\sqrt{\frac{s}{N}} + \frac{G_{f_\ast}(n,\Delta)}{\sqrt{n}} + \sqrt{n}\frac{s}{N} + \sqrt{n}\frac{\m O}{N}\r) \, \nu(\wh\delta_N) 
\end{multline}
on event $\Theta_1$ of probability at least $1 - 2e^{-s}$. 
As $\widetilde \Delta \geq \sigma(\ell,\m F)$ by assumption, we deduce that 
\begin{multline*}
\l| \frac{\hGk\l( 0; f_\ast \r)}{\partial_z \Gk\l(0; f_\ast \r)\partial_z \Gk\l(0; \wh f_N \r)}
\l(\partial_z \Gk\l(0; f_\ast \r) - \partial_z \Gk\l(0; \wh f_N \r) \r) \r| 
\\
\leq C(\rho)\nu(\wh\delta_N) \l( \sqrt{\frac s N} + \frac{G_{f_\ast}(n,\Delta)}{\sqrt{n}} + \sqrt{n}\frac{s}{N} + \sqrt{n}\frac{\m O}{N} \r).
\end{multline*}
Define 
\begin{equation}
\label{eq:delta_1}
\widebar \delta_1:=\min \l\{ \delta>0: \ C_1(\rho) \l( \sqrt{\frac{s}{N}} + \frac{G_{f_\ast}(n,\Delta)}{\sqrt{n}} + \sqrt{n} \frac{s + \m O}{N}\r) \, \frac{\widetilde\nu(\delta)}{\delta} \leq \frac{1}{7}\r\}
\end{equation}
where $C_1(\rho)$ is sufficiently large. It is easy to see that on event $\Theta_1\cap \{ \wh\delta_N> \widebar\delta_1\}$, 
\begin{equation}
\label{eq:nail1}
\l| \frac{\hGk\l( 0; f_\ast \r)}{\partial_z \Gk\l(0; f_\ast \r)\partial_z \Gk\l(0; \wh f_N \r)}
\l(\partial_z \Gk\l(0; f_\ast \r) - \partial_z \Gk\l(0; \wh f_N \r) \r) \r| \leq \frac{\wh\delta_N}{7},
\end{equation}
for appropriately chosen $C_1(\rho)$. 

Our next goal is to obtain an upper bound for 
$\l|\frac{\hGk\l(0;\wh f_N \r) - \hGk\l(0;f_\ast \r)}{\partial_z \Gk\l(0; \wh f_N \r)} \r|$. 
To this end, we will need to control the local oscillations of the process $\hGk\l(0;f \r)$. Specifically, we are interested in the  bounds on the random variable $\sup_{f\in \m F(\delta)}\l|  \hGk(0 ;f) - \hGk(0 ;f_\ast) \r|.$
The following technical lemma is important for the analysis. 
\begin{lemma}
\label{lemma:bias}
Let $(\xi_1,\eta_1),\ldots,(\xi_n,\eta_n)$ be a sequence of independent identically distributed random couples such that $\mb E\xi_1 = 0, \ \mb E\eta_1 = 0$, and $\mb E|\xi_1|^2 + \mb E|\eta_1|^2<\infty$. 
Let $F$ be an odd, smooth function with bounded derivatives up to fourth order. Then
\[
\l| \mb E F\l( \sum_{j=1}^n \xi_j \r) - \mb E F\l( \sum_{j=1}^n \eta_j \r) \r| \leq \max_{\alpha\in [0,1]}
\sqrt{n} \, \var^{1/2}\l( \xi_1 - \eta_1\r) \l(\mb E \l| F'\l( S_n^\eta + \alpha\l( S_n^\xi - S_n^\eta \r) \r) \r|^{2} \r)^{1/2}.
\]
Moreover, if $\mb E|\xi_1|^4 + \mb E|\eta_1|^4 < \infty$, then
\begin{multline*}
\l| \mb E F\l( \sum_{j=1}^n \xi_j \r) - \mb E F\l( \sum_{j=1}^n \eta_j \r) \r| \leq 
C(F) \cdot n \bigg( \var^{1/2}(\xi_1 - \eta_1)\l( R_4^2 + \sqrt{n-1} R_4^3\r)
\\
+ \l(\mb E |\xi_1 - \eta_1|^4 \r)^{1/4} \, R_4^3 \bigg),
\end{multline*}
where $R_4 = \l(\max\l(\mb E|\xi_1|^4,\mb E|\eta_1|^4\r)\r)^{1/4}$and $C(F)>0$ is a constant that depends only on $F$.
\end{lemma}

\begin{proof}
See section \ref{proof:lemma-bias}.
\end{proof}
\noindent Now we are ready to state the bound for the local oscillations of the process $\hGk(0;f)$. 
Let 
\[
U(\delta,s):=\frac{2}{\Delta}\l( 8\sqrt{2}\omega(\delta) +  \nu(\delta)\sqrt{\frac{s}{2}}\r) + \frac{32 s}{3\sqrt{k}} + \frac{2\m O}{\sqrt{k}}.
\]
Moreover, if $\widetilde \omega(\delta)$ and $\widetilde \nu(\delta)$ are upper bounds for $\omega(\delta)$ and $\nu(\delta)$ and are of concave type, then  
\begin{equation}
\label{eq:U-tilde}
\widetilde U(\delta,s):=\frac{2}{\Delta}\l( c(\gamma)\,\widetilde\omega(\delta) +  \widetilde\nu(\delta)\sqrt{\frac{s}{2}}\r) + \frac{32 s}{\sqrt{k}},
\end{equation}
where $c(\gamma)>0$ depends only on $\gamma$, is also an upper bound for $U(\delta,s)$ of strictly concave type. 
Moreover, define
\begin{align*}
R_4(\ell,\m F)&:=\sup_{f\in \m F} \mb E^{1/4} \Big( \ell(f(X))- \mb E \ell(f(X)) \Big)^4, 
\\
\nu_4(\delta) &:=\sup_{f\in \m F(\delta)} \mb E^{1/4} \bigg( \ell(f(X)) - \ell(f_\ast(X)) - \mb E\l( \ell(f(X)) - \ell(f_\ast(X))\r) \bigg)^4,
\\
\mathfrak{B}(\ell,\m F) &:= \frac{R_4(\ell,\m F)}{\sigma(\ell,\m F)},
\\
\widetilde B(\delta) &:=\begin{cases}
\frac{\widetilde\nu(\delta)}{\Delta} \frac{1}{M_\Delta}, & R_4(\ell, \m F) = \infty,
\\
\frac{\mathfrak{B}^3(\ell,\m F)}{\sqrt n}\Bigg( \frac{\widetilde\nu(\delta)}{\Delta}\frac{1}{M_\Delta^2} 
+ \frac{\widetilde\nu_4(\delta)}{\Delta} \frac{1}{M_\Delta^3 \sqrt{n}} \Bigg), & R_4(\ell, \m F) < \infty.
\end{cases}
\end{align*}
where $\widetilde \nu_4(\delta)$ upper bounds $\nu_4(\delta)$ and is of concave type. 
Below, we will use a crude bound $\nu_4(\delta)\leq 2 R_4(\ell, \m F)$, but additional improvements are possible if better estimates of $\nu_4(\delta)$ are available.  

\begin{lemma}
\label{lemma:modulus}
With probability at least $ 1 - e^{-2s}$,
\[
\sup_{f\in \m F(\delta)}\l| \hGk(0;f) - \hGk(0;f_\ast)  \r| \leq U(\delta,s) 
+ C(\rho)\sqrt{k} \, \widetilde B(\delta) + 4\frac{\m O}{\sqrt k}.
\]
where $C(\rho)>0$ is constant that depends only on $\rho$. 
\end{lemma}
\begin{proof}
See section \ref{proof:modulus}.
\end{proof}
\noindent Next, we state the ``uniform version'' of Lemma \ref{lemma:modulus}: 
\begin{lemma}
\label{lemma:modulus-unif}
With probability at least $1 - e^{-s}$, for all $\delta\geq \delta_{\min}$ simultaneously,
\[
\sup_{f\in \m F(\delta)}\l| \hGk(0;f) - \hGk(0;f_\ast)  \r| \leq C(\rho)
\delta \l( \frac{\widetilde U(\delta_{\min},s)}{\delta_{\min}} + \sqrt{k} \frac{\widetilde B(\delta_{\min})}{\delta_{\min}} \r) + 4\frac{\m O}{\sqrt k}
\]
where $C(\rho)>0$ is constant that depends only on $\rho$. 
\end{lemma}
\begin{proof}
See section \ref{proof:modulus-unif}.
\end{proof}

It follows from Lemma \ref{lemma:modulus-unif} and inequality \eqref{eq:derivative-lower} that on event $\Theta_2$ of  probability at least $1-e^{-s}$, for all $\delta\geq \delta_{\min}$ simultaneously, 
\begin{equation}
\label{eq:final-2}
\sup_{f\in \m F(\delta)}\l| \frac{\hGk\l(0; f \r) - \hGk\l(0;f_\ast \r)}{\partial_z \Gk\l(0; f \r)} \r|
\leq C(\rho)\delta\l( \frac{\widetilde \Delta}{\sqrt{N}}\frac{\widetilde U(\delta_{\min},s)}{\delta_{\min}} + \frac{\widetilde \Delta}{\sqrt n} \frac{\widetilde B(\delta_{\min})}{\delta_{\min}} \r) + 4\widetilde \Delta\sqrt{n} \frac{\m O}{N}.
\end{equation}
Define 
\begin{align*}
\widebar \delta_2&:=\min \l\{ \delta>0: \ C_2(\rho) \frac{\widetilde \Delta}{\sqrt{N}}\frac{\widetilde U(\delta,s)}{\delta} \leq \frac{1}{7}\r\},
\\
\widebar \delta_3&:=\min \l\{ \delta>0: \ C_3(\rho) \frac{\widetilde \Delta}{\sqrt n} \frac{\widetilde B(\delta)}{\delta} \leq \frac{1}{7}\r\}
\end{align*}
where $C_2(\rho), \ C_3(\rho)$ are sufficiently large constants. Then, on event $\Theta_2\cap \l\{ \wh\delta_N >\max(\widebar \delta_2,\widebar\delta_3)\r\}$, 
\begin{equation}
\label{eq:nail2}
\sup_{f\in \m F(\wh\delta_N)}\l| \frac{\hGk\l(0; f \r) - \hGk\l(0;f_\ast \r)}{\partial_z \Gk\l(0; f \r)} \r|
\leq \frac{2 \,\wh\delta_N}{7} + 4\widetilde \Delta\sqrt{n} \frac{\m O}{N}
\end{equation}
for appropriately chosen $C_2(\rho),C_3(\rho)$.

\noindent Finally, we provide an upper bound for the process $R_N(f)$ defined via 
\[
R_N(f) = \hGk\l(0;f \r) + \partial_z \Gk\l(0 ; f \r) \cdot \hEk(f).
\]
\begin{lemma}
\label{lemma:R_N}
Assume that conditions of Theorem \ref{th:unif} hold, and let $\delta_{\min}>0$ be fixed. Then for all $s>0$, $\delta\geq \delta_{\min}$, positive integers $n$ and $k$ such that 
\begin{equation}
\label{eq:key1}
\delta \, \frac{\widetilde U(\delta_{\min},s)}{\delta_{\min}\sqrt{k}} +
\sup_{f\in\m F} G_f(n,\Delta)
+ \frac{s+\m O}{k} \leq c(\rho),
\end{equation}
the following inequality holds with probability at least $1-7e^{-s}$, uniformly over all $\delta$ satisfying \eqref{eq:key1}:
\begin{multline}
\label{eq:A_2-final}
\sup_{f\in \m F(\delta)}| R_N(f) |  
\leq C(\rho) \sqrt{N} \frac{\widetilde \Delta^2}{\Delta^2} \bigg( n^{1/2}\delta^2 \l(\frac{\widetilde U(\delta_{\min},s)}{\delta_{\min}\sqrt{N}}\r)^2 \bigvee \frac{\sigma^2(\ell,f_\ast)}{\Delta^2} \frac{n^{1/2}\, s}{N}
\\ 
\bigvee n^{1/2}\l(\sup_{f\in \m F}  \frac{G_f\big(n,\Delta\big)}{\sqrt{n}}\r)^2 \bigvee n^{3/2} \frac{s^2}{N^2}
\bigvee n^{3/2} \frac{\m O^2}{N^2}
\bigg).
\end{multline}
Moreover, the bound of Theorem \ref{th:unif} holds on the same event.
\end{lemma}
\begin{proof}
See section \ref{proof:R_N}.
\end{proof}

\noindent Recall that  
\[
\widebar \delta_2 = \min \l\{ \delta>0: \ C_2(\rho) \frac{\widetilde \Delta}{\sqrt{N}}\frac{\widetilde U(\delta,s)}{\delta} \leq \frac{1}{7}\r\}
\] 
where $C_2(\rho)$ is a large enough constant. Let $\Theta_3$ be the event of probability at least $1-7e^{-s}$ on which Lemma \ref{lemma:R_N} holds with $\delta_{\min}=\widebar \delta_2$, and consider the event $\Theta_3\cap \{\wh\delta_N >\widebar\delta_2\}$. 
We will now show that on this event, Lemma \ref{lemma:R_N} applies with $\delta=\wh\delta_N$. 
Indeed, the bound of Theorem \ref{th:unif} is valid on $\Theta_3$,  
hence the inequality \eqref{eq:slow-rates1} implies that on $\Theta_3$, 
$\wh\delta_N\leq C(\rho)\frac{\widetilde \Delta}{\sqrt n}$, and it is straightforward to check that condition \eqref{eq:key1} of Lemma \ref{lemma:R_N} holds with $\delta_{\min} = \widebar \delta_2$ and $\delta = \wh\delta_N$. 
It follows from inequality \eqref{eq:derivative-lower} that on event $\Theta_3\cap \{ \wh\delta_N \geq \widebar \delta_2 \}$, 
\begin{multline*}
\sup_{f\in \m F(\wh\delta_N)} \l|  \frac{R_N(f)}{\partial_z \Gk\l(0; f \r)} \r| 
\leq 
C(\rho)\frac{\widetilde \Delta^2}{\Delta^2}
\bigg( \frac{n^{1/2}}{\widetilde \Delta}\wh\delta_N^2\l(\frac{\widetilde \Delta}{\sqrt{N}}\frac{\widetilde U(\delta_{2},s)}{\delta_{2}}\r)^2 \bigvee \widetilde \Delta\frac{\sigma^2(\ell,f_\ast)}{\Delta^2} \frac{n^{1/2}\, s}{N}
\\ 
\bigvee n^{1/2}\widetilde \Delta \l(\sup_{f\in \m F}  \frac{G_f\big(n,\Delta\big)}{\sqrt{n}}\r)^2 \bigvee n^{3/2}\widetilde \Delta \frac{s^2 + \m O^2}{N^2}
\bigg).
\end{multline*}
Consider the expression 
\[
C(\rho)\frac{\widetilde \Delta^2}{\Delta^2}
\frac{n^{1/2}}{\widetilde \Delta}\wh\delta_N^2\l(\frac{\widetilde \Delta}{\sqrt{N}}\frac{\widetilde U(\delta_{2},s)}{\delta_{2}}\r)^2  = 
C(\rho) \frac{\widetilde \Delta^2}{\Delta^2} \l(\frac{\widetilde \Delta }{\sqrt N}\frac{\widetilde U(\delta_{2},s)}{\delta_{2}}\r)^2 \wh\delta_N \cdot \frac{n^{1/2}\wh\delta_N}{\widetilde \Delta}
\]
and observe that whenever Theorem \ref{th:unif} holds, $\frac{n^{1/2}\wh\delta_N}{\widetilde \Delta}\leq c(\rho)$, hence the latter is bounded from above by
\[
\wh\delta_N\cdot C(\rho) \frac{\widetilde \Delta^2}{\Delta^2} \l(\frac{\widetilde \Delta }{\sqrt N}\frac{\widetilde U(\widebar\delta_{2},s)}{\widebar\delta_{2}}\r)^2 \leq \frac{\wh\delta_N}{7}
\] 
whenever $\Delta\geq \sigma(\ell,\m F)$ (so that $\widetilde \Delta = \Delta$) and $C_2(\rho)$ in the definition of $\widebar\delta_2$ is large enough. 
Moreover, 
\[
C(\rho)\frac{\widetilde \Delta^3}{\Delta^3}\frac{\sigma^2(\ell,f_\ast)}{\Delta} \frac{n^{1/2}\, s}{N} 
\leq C'(\rho)\cdot \sigma(\ell,f_\ast) \sqrt{n}\frac{s}{N} \leq C'(\rho)\widetilde \Delta \sqrt{n}\frac{s}{N}
\]
if $\widetilde \Delta \geq \sigma(\ell,f_\ast)$. 
As $\frac{s+\m O}{k}\leq c$ under the conditions of Theorem \ref{th:unif},  
\[
n^{3/2}\widetilde \Delta \frac{s^2 + \m O^2}{N^2} \leq C\widetilde \Delta \sqrt{n}\frac{s + \m O}{N}.
\]
Combining the inequalities obtained above, we deduce on event $\Theta_3\cap \{ \wh\delta_N \geq \widebar \delta_2 \}$,
\begin{equation}
\label{eq:prep}
2\sup_{f\in \m F(\wh\delta_N)} \l|  \frac{R_N(f)}{\partial_z \Gk\l(0; f \r)} \r| 
\leq \frac{2\wh\delta_N}{7} + C(\rho)\widetilde \Delta \l( \sqrt{n}\frac{s+\m O}{N} \bigvee \frac{\sup_{f\in \m F}  \l( G_f\big(n,\Delta\big)\r)^2}{\sqrt{n}} \r)
\end{equation}
whenever $\widetilde\Delta\geq \sigma(\ell,\m F)$. Finally, define 
\[
\widebar\delta_4:=C_4(\rho)\widetilde \Delta \l( \sqrt{n}\frac{s+\m O}{N} \bigvee \frac{\sup_{f\in \m F}  \l( G_f\big(n,\Delta\big)\r)^2}{\sqrt{n}} \r),
\]
where $C_4(\rho)$ is sufficiently large. Then on event $\Theta_3\cap \l\{ \wh\delta_N \geq \max\l(\widebar \delta_2,7\widebar\delta_4\r) \r\}$, 
\begin{equation}
\label{eq:nail3}
2\sup_{f\in \m F(\wh\delta_N)} \l|  \frac{R_N(f)}{\partial_z \Gk\l(0; f \r)} \r| + 4\widetilde \Delta\sqrt{n} \frac{\m O}{N}
\leq \frac{2\wh\delta_N}{7} + \frac{\wh\delta_N}{7} = \frac{3\wh\delta_N}{7}.
\end{equation}
Note that the expression above takes care of the term $4\widetilde \Delta\sqrt{n} \frac{\m O}{N}$ that appeared in \eqref{eq:nail2}. 
Combining \eqref{eq:nail1},\eqref{eq:nail2},\eqref{eq:nail3}, we deduce that on event $\Theta_1\cap \Theta_2\cap \Theta_3 \cap \l\{ \wh\delta_N\geq \max\l( \widebar \delta_1,\widebar \delta_2,\widebar \delta_3,7\,\widebar \delta_4\r)\r\}$, 
\[
\wh\delta_N \leq \frac{6}{7}\wh\delta_N
\]
leading to a contradiction, hence on event $\Theta_1\cap \Theta_2\cap \Theta_3$ of probability at least $1-10e^{-s}$, 
\begin{equation}
\label{eq:result}
\wh\delta_N \leq \max\l( \widebar \delta_1,\widebar \delta_2,\widebar \delta_3, 7\widebar \delta_4\r). 
\end{equation}
Recall the definition \eqref{eq:delta_1} of $\widebar \delta_1$. If condition \ref{ass:bern} (``Bernstein condition'') holds, then $\widetilde \nu(\delta)\leq D\sqrt{\delta}$ for small enough $\delta$, in which case
\[
\widebar\delta_1 \leq C(\rho)D^2\l( \frac{s+\m O}{N} + \frac{G^2_{f_\ast}(n,\Delta)}{n} \r),
\]
where we used the fact that $\frac{s}{k}\leq c$ by assumption. Together with the bound \eqref{eq:g_f} for $G_{f_\ast}(n,\Delta)$, we deduce that, under the assumption that $R_4(\ell,\m F)<\infty$, 
\begin{equation*}
\widebar\delta_1 \leq C(\rho)D^2\l( \frac{s+\m O}{N} + \frac{\l( \mb E\big| f_\ast(X) - \mb Ef_\ast(X)\big|^{3}\r)^2}{\Delta^{6}\, n^{2}} \r).
\end{equation*}
Since $\Delta=\sigma(\ell,\m F)M_\Delta$,
$\frac{\mb E\big| f_\ast(X) - \mb Ef_\ast(X)\big|^{3}}{\Delta^{3}} \leq \frac{\sup_{f\in \m F}\mb E\big|f(X) - \mb Ef(X)\big|^{3}}{\sigma^3(\ell,\m F) M_\Delta^3} \leq \frac{\mathfrak{B}^3(\ell,\m F)}{M_\Delta^3}$, where 
\[
\mathfrak{B}(\ell,\m F) = \frac{\sup_{f\in \m F}\mb E^{1/4}\l( \ell(f(X)) - \mb E\ell(f(X)) \r)^4}{\sigma(\ell, \m F)},
\]
hence 
\begin{equation}
\label{eq:delta-1}
\widebar\delta_1 \leq C(\rho)D^2\l( \frac{s+\m O}{N} + \frac{\mathfrak{B}^6(\ell,\m F)}{n^2 M_\Delta^6}\r).
\end{equation}
At the same time, if only $\sigma(\ell,\m F)<\infty$, we similarly obtain that 
\begin{equation}
\label{eq:delta-1-weak}
\widebar\delta_1 \leq C(\rho)D^2\l( \frac{s+\m O}{N} + \frac{1}{M_\Delta^4\, n} \r).
\end{equation}

\noindent Next we will estimate $\widebar\delta_3$. Recall that, when $R_4(\ell,\m F)<\infty$, 
\[
\widetilde B(\delta) = 
\frac{\mathfrak{B}^3(\ell,\m F)}{\sqrt n} \Bigg( \frac{\widetilde\nu(\delta)}{\Delta}\frac{1}{M_\Delta^2} 
+ \frac{\widetilde\nu_4(\delta)}{\Delta} \frac{1}{M_\Delta^3 \sqrt{n}} \Bigg).
\]
For sufficiently small $\delta$ (namely, for which condition \ref{ass:bern} holds) and $\Delta\geq \sigma(\ell,\m F)$,
\[
\frac{\widetilde \Delta}{\sqrt n} \widetilde B(\delta) \leq \frac{\mathfrak{B}^3(\ell,\m F)}{n} \l( \frac{\widetilde \nu(\delta)}{M_\Delta^2} + \frac{R_4(\ell, \m F)}{M_\Delta^3 \sqrt n} \r) \leq 
\frac{\mathfrak{B}^3(\ell,\m F)}{n} \l( D\frac{\sqrt{\delta}}{M_\Delta^2} + \sigma(\ell, \m F)\frac{\mathfrak{B}(\ell,\m F)}{M_\Delta^3 \, \sqrt n} \r)  
\]
and 
\begin{equation}
\label{eq:delta-3}
\widebar \delta_3 \leq C(\rho)\l( D^2 \frac{\mathfrak{B}^6(\ell,\m F)}{n^2\, M_\Delta^4 } + \sigma(\ell, \m F)\frac{\mathfrak{B}^4(\ell,\m F) }{n^{3/2} M_\Delta^3 }\r).
\end{equation}
At the same time, if only the second moments are finite, $\widetilde B(\delta) = \frac{\widetilde \nu(\delta)}{\Delta} \frac{1}{M_\Delta}$, and it is easy to deduce that in this case,
\begin{equation}
\label{eq:delta-3-weak}
\widebar \delta_3 \leq C(\rho)\frac{D^2}{M_\Delta^2 \,n}.
\end{equation}
Next, we obtain a simpler bound for $\widebar \delta_4$: as $\Delta\geq \sigma(\ell,\m F)$ by assumption, $\widetilde \Delta = \Delta = \sigma(\ell,\m F)\, M_\Delta$, and the estimate \eqref{eq:g_f} for $G_{f_\ast}(n,\Delta)$ implies (if $R_4(\ell,\m F)<\infty$) that
\begin{equation}
\label{eq:delta-4}
\widebar \delta_4 \leq C(\rho) \, \sigma(\ell,\m F)\l( \sqrt{n}M_\Delta \,\frac{s+\m O}{N} +  \frac{\mathfrak{B}^6(\ell,\m F)}{M_\Delta^5 n^{3/2}}\r).
\end{equation}
If only $\sigma(\ell,\m F)<\infty$, we similarly deduce from \eqref{eq:g_f} that 
\begin{equation}
\label{eq:delta-4-weak}
\widebar \delta_4 \leq C(\rho) \, \sigma(\ell,\m F)\l( \sqrt{n}M_\Delta \cdot\frac{s+\m O}{N} + \frac{1}{M_\Delta^3 \sqrt{n}} \r).
\end{equation}
Finally, recall that 
\[
\widetilde U(\delta,s)=\frac{2}{\Delta}\l( c(\gamma)\,\widetilde\omega(\delta) +  \widetilde\nu(\delta)\sqrt{\frac{s}{2}}\r) + \frac{32 s}{\sqrt{k}}
\]
and 
$\widebar \delta_2 = \min \l\{ \delta>0: \ C_2(\rho) \frac{\widetilde \Delta}{\sqrt{N}}\frac{\widetilde U(\delta,s)}{\delta} \leq \frac{1}{7}\r\}$, 
hence 
\begin{equation}
\label{eq:delta-2}
\widebar\delta_2 \leq \widebar \delta \bigvee C(\rho) D^2 \frac{s}{N} \bigvee C(\rho)\sigma(\ell,\m F)\frac{s\sqrt{n}M_\Delta }{N},
\end{equation}
where $\widebar\delta$ was defined in \eqref{eq:delta-bar}. 
Combining inequalities \eqref{eq:delta-1}, \eqref{eq:delta-2} \eqref{eq:delta-3}, \eqref{eq:delta-4} and \eqref{eq:result}, we obtain the final form of the bound under the stronger assumption $R_4(\ell,\m F)<\infty$. 
Similarly, the combination of  \eqref{eq:delta-1-weak}, \eqref{eq:delta-2} \eqref{eq:delta-3-weak}, \eqref{eq:delta-4-weak} and \eqref{eq:result} yields the bound under the weaker assumption $\sigma(\ell,\m F)<\infty$.


\subsection{Proof of Theorem \ref{cor:fast}.}
\label{proof:fast}

Recall that $\widehat{\m E}_N(f_\ast) := \hL(f_\ast) - \hL(\wh f'_N)$ is the ``empirical excess risk'' of $f_\ast$, and let $\widehat \delta_N:=\m E(\widehat f'_N)$.  
It follows from Remark \ref{remark:emp-risk} that (using the notation used in the proof of Theorems \ref{th:fast-rates-1} and \ref{th:fast-rates-2})
\[
\widehat{\m E}_N(f_\ast) \leq \l| \frac{\hGk\l(0;\wh f'_N \r)}{\partial_z G\l(0; \wh f'_N \r)}  - \frac{\hGk\l( 0 ;f_\ast \r)}{\partial_z G\l(0; f_\ast \r)} \r|
+ 2\sup_{f\in \m F(\wh\delta_N)} \l|  \frac{R_N(f)}{\partial_z \Gk\l(0; f \r)} \r|.
\]
On the event of Theorem \ref{th:fast-rates-2} of probability at least $1-10e^{-s}$, 
\[
\m E(\widehat f'_N) \leq \delta^{\prime} := \widebar \delta + 
C(\rho)\l( D^2 \sigma(\ell,\m F)\sqrt{n}M_\Delta\r) \l(\frac{\mathfrak{B}^6(\ell,\m F)}{M_\Delta^4 n^2} + \frac{s+\m O}{N}\r),
\]
hence on this event 
\begin{equation*}
\widehat{\m E}_N(f_\ast) \leq \sup_{f\in \m F(\delta^{\prime})}\l| \frac{\hGk\l(0; f\r)}{\partial_z G\l(0; f \r)}  - \frac{\hGk\l( 0 ;f_\ast \r)}{\partial_z G\l(0; f_\ast \r)} \r|
+ 2\sup_{f\in \m F(\delta^{\prime})} \l|  \frac{R_N(f)}{\partial_z \Gk\l(0; f \r)} \r| 
\leq \frac{6}{7}\delta^{\prime} 
\end{equation*}
where the last inequality again follows from main steps in the proof of Theorem \ref{th:fast-rates-2}. 
\footnote{Similar result holds if $\delta^{\prime}$ is replaced by its analogue from Theorem \ref{th:fast-rates-2}. }
Consider the set $\widehat{\m F}(\delta^{\prime}) = \l\{ f\in \m F: \ \widehat{\m E}_N(f) \leq \delta^{\prime} \r\}$. 
First, observe that on the event $\m E_1$ of Theorem \ref{th:fast-rates-2}, $f_\ast \in \widehat{\m F}(\delta^{\prime})$ as implied by the previous display. 
We will next show that $\widehat{\m F}(\delta^{\prime}) \subseteq \m F(7\delta^{\prime})$ on the event $\m E_1$ of Theorem \ref{th:fast-rates-2}, meaning that for any $f\in \widehat{\m F}(\delta^{\prime})$, $\m E(f)\leq 7\delta^{\prime}$. 
Indeed, let $f\in \widehat{\m F}(\delta^{\prime})$ be such that $\m E(f)=\sigma$. Then \eqref{eq:risk-lin} implies that 
\begin{multline*}
\m L(f) - \m L(f_\ast) \leq \hL(f) - \hL(f_\ast) + \l| \frac{\hGk\l( 0; f \r)}{\partial_z \Gk\l(0; f \r)} - \frac{\hGk\l( 0; f_\ast \r)}{\partial_z \Gk\l(0; f_\ast \r)} \r|
+ \l| \frac{R_N(f)}{\partial_z \Gk\l(0; f \r)} + \frac{R_N(f_\ast)}{\partial_z \Gk\l(0; f_\ast \r)} \r|
\\
\leq \widehat{\m E}_N(f) + \sup_{f\in \m F(\sigma)}\l| \frac{\hGk\l( 0; f \r)}{\partial_z \Gk\l(0; f \r)} - \frac{\hGk\l( 0; f_\ast \r)}{\partial_z \Gk\l(0; f_\ast \r)} \r| + 2\sup_{f\in \m F(\sigma)}\l| \frac{R_N(f)}{\partial_z \Gk\l(0; f \r)} \r|.
\end{multline*}
Again, it follows from the arguments used in proof of Theorem \ref{th:fast-rates-2} that on event $\m E_1$ of probability at least $1-10e^{-s}$, 
\begin{equation*}
\sup_{f\in \m F(\sigma)}\l| \frac{\hGk\l( 0; f \r)}{\partial_z \Gk\l(0; f \r)} - \frac{\hGk\l( 0; f_\ast \r)}{\partial_z \Gk\l(0; f_\ast \r)} \r| + 2\sup_{f\in \m F(\sigma)}\l| \frac{R_N(f)}{\partial_z \Gk\l(0; f \r)} \r| \leq \frac{6}{7}\max\l( \delta^{\prime}, \sigma \r).
\end{equation*}
Consequently, $\sigma \leq \delta^{\prime} + \frac{6}{7}\max\l( \delta^{\prime}, \sigma \r)$ on this event, implying that $\sigma \leq 7\delta^{\prime}$. 
Next, Assumption \ref{ass:bern} yields that 
\begin{multline*}
\sup_{f\in \widehat{\m F}(\delta^{\prime})} \var\l( \ell(f(X)) - \ell(\wh f'_N)\r) 
\\
\leq 2 \l( \sup_{f\in \widehat{\m F}(\delta^{\prime})} \var\l( \ell(f(X)) - \ell(f_\ast(X))\r) + \var\l( \ell(\wh f'_N(X))- \ell(f_\ast(X))\r) \r) \leq 2D(\sqrt{7}+1) \delta^{\prime}
\end{multline*}
on $\m E_1$.
It remains to apply Theorem \ref{th:fast-rates-2}, conditionally on $\m E_1$, to the class 
\[
\widehat{\m F}(\delta^{\prime}) - \wh f'_N := \l\{ f - \wh f'_N, \ f\in \widehat{\m F}(\delta^{\prime})\r\}.
\]
To this end, we need to verify the assumption of Theorem \ref{th:unif} that translates into the requirement 
\[
c\Delta_2 \geq  \frac{1}{\sqrt{k_2}} \, \mb E\sup_{f\in \m F(7\delta^{\prime})} \frac{1}{\sqrt{|S_2|}}\sum_{j=1}^{|S_2|} \l( \ell(f(X_j)) -\ell(f_\ast(X_j))- P(\ell(f) - \ell(f_\ast)) \r).
\] 
As $\delta^{\prime} > \widebar \delta$ and $|S_2|\geq \lfloor N/2\rfloor$, we have the inequality 
\[
\mb E\sup_{f\in \m F(7\delta^{\prime})} \frac{1}{\sqrt{|S_2|}}\sum_{j=1}^{|S_2|} \l( \ell(f(X_j)) -\ell(f_\ast(X_j))- P(\ell(f) - \ell(f_\ast)) \r) \leq C\delta^{\prime}\sqrt{N},
\]
hence it suffices to check that $\Delta_2 = DM_{\Delta_2}\sqrt{7\delta^{\prime}}\geq C\delta^{\prime}\sqrt{\frac{N}{k_2}}$. 
The latter is equivalent to $\delta^{\prime}\leq C D^2 M_{\Delta_2}^2 \frac{k_2}{N}$ that holds by assumption. 
Result now follows easily as we assumed that the subsamples $S_1$ and $S_2$ used to construct $\wh f'_N$ and $\wh f''_N$ are disjoint.

\section*{Acknowledgements}

Stanislav Minsker gratefully acknowledges support by the National Science Foundation grant DMS-1712956.	

\bibliographystyle{imsart-nameyear}	
\bibliography{RobustERM,biblio2}

\appendix 

\section{Remaining proofs.}

\subsection{Proof of Lemma \ref{lemma:derivative}.}
\label{proof:derivative}
	
As $\rho$ is sufficiently smooth, 
\[
\partial_z \Gk\l(0;f \r) = -\frac{\sqrt{kn}}{\Delta} \mb E \rho''\l(\sqrt{n} \frac{\bL_1(f) - \m L(f) }{\Delta}\r).
\]
Let $W(\ell(f))$ denote a centered normal random variable variance equal to $\var\l( \ell(f(X))\r)$. 
Lemma \ref{lemma:lindeberg} implies that 
\[
\l| \mb E \rho''\l(\sqrt{n} \frac{\bL_1(f) - \m L(f) }{\Delta} \r) - \mb E \rho''\l( \frac{W(\ell(f))}{\Delta}\r) \r| 
\leq C\frac{\|\rho^{(5)}\|_\infty}{\Delta^3 \sqrt{n}} \mb E \big| \ell(f(X)) - P\ell(f)\big|^3. 
\]
Next, as $\rho''(x)\geq I\{ |x|\leq 1\}$ by assumption, 
\[
\mb E \rho''\l( \frac{W(\ell(f))}{\Delta}\r)\geq \pr{\l| W(\ell(f))\r|\leq \Delta}.
\]
Gaussian tail bound implies that 
\[
\pr{\l| W(\ell(f))\r|\leq \Delta} \geq 1 - 2 \exp\l( - \frac{1}{2}\frac{\Delta^2}{\var\l( \ell(f(X))\r)}\r) \geq \frac{1}{2}
\]
whenever $\Delta^2\geq 4\log(2) \var\l( \ell(f(X))\r)$. 
On the other hand, if $\xi\sim N(0,1)$, then clearly $\pr{Z\leq |t|}\geq \frac{2|t|}{\sqrt{2\pi}}e^{-t^2/2}$, hence
\[
\pr{\l| W(\ell(f))\r|\leq \Delta} \geq \frac{2\Delta}{\sqrt{2\pi \var\l( \ell(f(X))\r)}} \exp\l( - \frac{1}{2}\frac{\Delta^2}{\var\l( \ell(f(X))\r)}\r) \geq \frac{\Delta}{\sqrt{8\pi \var\l( \ell(f(X))\r)}}
\]
whenever $\Delta^2 < 4\log(2) \var\l( \ell(f(X))\r)$. Combination of two bounds yields that 
\[
\pr{\l| W(\ell(f))\r|\leq \Delta} \geq \frac{1}{2\sqrt{2\pi}}\min\l(\frac{\Delta}{\sqrt{\var\l( \ell(f(X))\r)}}, 2\sqrt{\log 2}\r).
\]

\subsection{Proof of Lemma \ref{lemma:hGk}.}
\label{proof:hGk}

Observe that 
\begin{multline*}
\frac{1}{\sqrt{k}} \sum_{j=1}^k \rho'\l(\sqrt{n} \frac{\bL_1(f) - \m L(f) }{\Delta} \r) = 
\frac{1}{\sqrt k} \sum_{j\in J} \rho'\l(\sqrt{n} \frac{\bL_1(f) - \m L(f) }{\Delta} \r) + \frac{1}{\sqrt k} \sum_{j\notin J} \rho'\l(\sqrt{n} \frac{\bL_1(f) - \m L(f) }{\Delta} \r) 
\\ 
\leq 
\sqrt{\frac{|J|}{k}} \frac{1}{\sqrt{|J|}} \sum_{j\in J} \rho'\l(\sqrt{n} \frac{\bL_1(f) - \m L(f) }{\Delta} \r) + 2\frac{\m O}{\sqrt k},
\end{multline*}
where we used the fact that $\|\rho'\|_\infty\leq 2$.
Bernstein's inequality implies that 
\begin{multline*}
\l| \frac{1}{\sqrt{|J|}}\l(\sum_{j\in J} \rho'\l(\sqrt{n} \frac{\bL_1(f) - \m L(f) }{\Delta} \r) - \mb E \rho'\l(\sqrt{n} \frac{\bL_1(f) - \m L(f) }{\Delta} \r) \r)\r| 
\\
\leq  2\l( \var^{1/2}\l(\rho'\l( \sqrt{n} \frac{\bL_1(f) - \m L(f) }{\Delta}\r)\r)\sqrt{s} + \frac{2s}{\sqrt{|J|}}\r)
\end{multline*}
with probability at least $1 - 2e^{-s}$, where we again used the fact that $\|\rho'\|_\infty\leq 2$. Moreover, 
$\var\l( \rho'\l( \sqrt{n} \frac{\bL_1(f) - \m L(f) }{\Delta}\r) \r)\leq \frac{\sigma^2(\ell,f)}{\Delta^2}$ by Lemma \ref{lemma:variance}, hence with the same probability 
\[
|\hGk(0;f)| \leq \sqrt{k}\l| \mb E  \rho'\l(\sqrt{n} \frac{\bL_1(f) - \m L(f) }{\Delta} \r) \r| + 2\l(\frac{\sigma(\ell,f)}{\Delta}\sqrt{s} + \frac{2s}{\sqrt k} + \frac{\m O}{\sqrt k}\r).
\] 
Lemma 6.2 in \citep{minsker2018uniform} implies that 
\[
\l| \mb E  \rho'\l(\sqrt{n} \frac{\bL_1(f) - \m L(f) }{\Delta} \r)\r| 
\leq 
\underbrace{\mb E\rho'\l( \frac{W(\ell(f))}{\Delta} \r)}_{=0} + 2 G_f(n,\Delta),
\]
hence the claim follows.

\subsection{Proof of Lemma \ref{lemma:bias}.}
\label{proof:lemma-bias}

Since $F$ is smooth, for any $x,y\in \mb R$, $F(y) - F(x) = \int_{0}^1 F'(x + \alpha(y-x))d\alpha \cdot(y-x)$. 
Let $S_n^\xi = \sum_{j=1}^n \xi_j, \ S_n^\eta = \sum_{j=1}^n \eta_j$. Then
\begin{equation*}
F\l( S_n^\xi \r) - F\l( S_n^\eta \r)
 = \l( S_n^\xi  - S_n^\eta\r) \int_0^1 F'\l( S_n^\eta + \alpha\l( S_n^\xi - S_n^\eta \r) \r)d\alpha,
\end{equation*}
hence 
\[
\mb E\l(  F\l( S_n^\xi \r) - F\l( S_n^\eta \r) \r) = 
\int_0^1  \mb E\l[ \l( S_n^\xi  - S_n^\eta\r)F'\l( S_n^\eta + \alpha\l( S_n^\xi - S_n^\eta \r) \r) \r] d\alpha.
\]
H\"{older}'s inequality yields that 
\begin{multline*}
\Big|  \mb E \l( S_n^\xi  - S_n^\eta\r) F'\l( S_n^\eta + \alpha\l( S_n^\xi - S_n^\eta \r) \r) \Big| \leq 
\l( \mb E \l| S_n^\xi  - S_n^\eta \r|^2 \r)^{1/2} \l(\mb E \l| F'\l( S_n^\eta + \alpha\l( S_n^\xi - S_n^\eta \r) \r) \r|^{2} \r)^{1/2}
\\
\leq \sqrt{n} \, \var^{1/2}\l( \xi_1 - \eta_1\r) \l(\mb E \l| F'\l( S_n^\eta + \alpha\l( S_n^\xi - S_n^\eta \r) \r) \r|^{2} \r)^{1/2},
\end{multline*}
implying the first inequality.
The rest of the proof is devoted to the second inequality of the lemma. Let $(W,Z)$ be a centered Gaussian vector with the same covariance as $(\xi_1,\eta_1)$, and let $(W_1,Z_1),\ldots,(W_n,Z_n)$ be i.i.d. copies of $(W,Z)$. We also set $S_n^W = \sum_{j=1}^n W_j, \ S_n^Z = \sum_{j=1}^n Z_j$. 
As $\mb E F\l( S_n^W\r) = \mb E F\l( S_n^Z \r) = 0$ for bounded odd $F$, it is easy to see that 
\begin{multline*}
\l| \mb E\l(  F\l( S_n^\xi \r) - F\l( S_n^\eta \r) \r) \r| = 
\l| \mb E\l(  F\l( S_n^\xi \r) - F\l( S_n^\eta \r) \r) - \mb E\l( F\l( S_n^W \r) - F\l( S_n^Z \r) \r) \r|
\\
=\Big| \int_0^1  \Big( \mb E \l( S_n^\xi  - S_n^\eta\r)F'\l( S_n^\eta + \alpha\l( S_n^\xi - S_n^\eta \r) \r) 
- \mb E \l( S_n^W  - S_n^Z\r)F'\l( S_n^Z + \alpha\l( S_n^W - S_n^Z \r) \r) \Big) d\alpha \Big|
\\
\leq \int_0^1 \Big|  \mb E \l( S_n^\xi  - S_n^\eta\r)F'\l( S_n^\eta + \alpha\l( S_n^\xi - S_n^\eta \r) \r) 
- \mb E  \l( S_n^W  - S_n^Z\r)F'\l( S_n^Z + \alpha\l( S_n^W - S_n^Z \r) \r) \Big| d\alpha.
\end{multline*}
Next we will estimate, for each $\alpha\in[0,1]$, the expression 
\begin{equation}
\label{eq:diff-alpha}
\Big|  \mb E \l( S_n^\xi  - S_n^\eta\r)F'\l( S_n^\eta + \alpha\l( S_n^\xi - S_n^\eta \r) \r) 
- \mb E  \l( S_n^W  - S_n^Z\r)F'\l( S_n^Z + \alpha\l( S_n^W - S_n^Z \r) \r) \Big|.
\end{equation}
To this end, we will use Lindeberg's replacement method. For $i=0,\ldots,n$, denote 
\[
T_i = (\xi_1-\eta_1,\ldots,\xi_{i}-\eta_i,W_{i+1}-Z_{i+1},\ldots,W_n-Z_n,\eta_1,\ldots,\eta_{i},Z_{i+1},\ldots,Z_n).
\] 
Then the expression in \eqref{eq:diff-alpha} is equal to 
$|\mb E G(T_n) - \mb E G(T_0)|$, where 
\[
G(T) = \l( \sum_{i=1}^n T^{(i)} \r) F'\l( \sum_{j=1}^n \l(T^{(j+n)} + \alpha T^{(j)} \r) \r)
\]  
and $T^{(j)}$ stands for the $j$-th coordinate of $T$. Clearly, 
\begin{equation}
\label{eq:telescopic}
|\mb E G(T_n) - \mb E G(T_0)| \leq \sum_{i=1}^n \l| \mb E G(T_i) - \mb EG(T_{i-1}) \r|.
\end{equation} 
Fix $i$, and consider the Taylor expansions of $G(T_i)$ and $G(T_{i-1})$ at the point 
\[
T_i^0=(\xi_1-\eta_1,\ldots,\xi_{i-1}-\eta_{i-1},0,W_{i+1}-Z_{i+1},\ldots,W_n-Z_n,\eta_1,\ldots,\eta_{i-1},0,Z_{i+1},\ldots,Z_n)
\] 
(note that $T_i^0$ does not depend on $\xi_i$, $\eta_i$, $W_i$ and $Z_i$). For $G(T_i)$ we get, setting $\delta_i = \xi_i - \eta_i$,  
\begin{multline*}
G(T_i) = G(T_i^0) + \partial_i G(T_i^0)\cdot \delta_i + \partial_{n+i} G(T_i^0)\cdot \eta_i 
\\
+ \frac{1}{2}\l( \partial^2_{i,i} G(T_i^0)\cdot \delta_i^2 +  2\partial^2_{i,n+i} G(T_i^0)\cdot \delta_i \eta_i + \partial^2_{n+i,n+i} G(T_i^0)\cdot\eta_i^2\r)
\\
+ \frac{1}{6}\l(\partial^3_{i,i,i} G(\tilde T_i^0)\cdot \delta_i^3 + \partial^3_{n+i,n+i,n+i} G(\tilde T_i^0) \cdot\eta_i^3 + \partial^3_{n+i,n+i,i} G(\tilde T_i^0)\cdot \eta_i^2 \delta_i + \partial^3_{n+i,i,i} G(\tilde T_i^0)\cdot \eta_i \delta_i^2 \r),
\end{multline*}
where $\tilde T_i^0$ is a point on a line segment between $T_i^0$ and $T_i$. Similarly, setting $\Delta_i=W_i - Z_i$,
\begin{multline}
\label{eq:gaussian}
G(T_{i-1}) = G(T_i^0) + G(T_i^0) + \partial_i G(T_i^0)\cdot \Delta_i + \partial_{n+i} G(T_i^0)\cdot Z_i 
\\
+ \frac{1}{2}\l( \partial^2_{i,i} G(T_i^0)\cdot \Delta_i^2 +  \partial^2_{i,n+i} G(T_i^0)\cdot \Delta_i Z_i + \partial^2_{n+i,n+i} G(T_i^0)\cdot Z_i^2\r)
\\
+ \frac{1}{6}\l(\partial^3_{i,i,i} G(\tilde T_i^0)\cdot \Delta_i^3 + \partial^3_{n+i,n+i,n+i} G(\tilde T_i^0) \cdot Z_i^3 + 3\partial^3_{n+i,n+i,i} G(\tilde T_i^0)\cdot Z_i^2 \Delta_i + 3\partial^3_{n+i,i,i} G(\tilde T_i^0)\cdot Z_i \Delta_i^2 \r),
\end{multline}
where $\bar T_i^0$ is a point on a line segment between $T_i^0$ and $T_{i-1}$. 
Using independence of $T_i^0$ and $(\xi_i,\eta_i,W_i,Z_i)$ and the fact that covariance structures of $(\xi_i,\eta_i)$ and $(W,Z)$ are the same, we deduce that
\begin{multline*}
\l| \mb E G(T_i) - \mb EG(T_{i-1}) \r| \leq \frac{1}{6} \mb E \Big| \partial^3_{i,i,i} G(\tilde T_i^0)\cdot \delta_i^3 + \partial^3_{n+i,n+i,n+i} G(\tilde T_i^0) \cdot\eta_i^3 + 3\partial^3_{n+i,n+i,i} G(\tilde T_i^0)\cdot \eta_i^2 \delta_i 
\\
+ 3\partial^3_{n+i,i,i} G(\tilde T_i^0)\cdot \eta_i \delta_i^2 \Big|
\\
+  \frac{1}{6}\mb E \Big| \partial^3_{i,i,i} G(\tilde T_i^0)\cdot \Delta_i^3 + \partial^3_{n+i,n+i,n+i} G(\tilde T_i^0) \cdot Z_i^3 + 3\partial^3_{n+i,n+i,i} G(\tilde T_i^0)\cdot Z_i^2 \Delta_i + 3\partial^3_{n+i,i,i} G(\tilde T_i^0)\cdot Z_i \Delta_i^2 \Big|.
\end{multline*}
It remains estimate each of the terms above. Assume that $\tau\in[0,1]$ is such that 
\[
\tilde T_i^0 =(\xi_1-\eta_1,\ldots,\xi_{i-1}-\eta_{i-1},\tau(\xi_i-\eta_i),W_{i+1}-Z_{i+1},\ldots,W_n-Z_n,\eta_1,\ldots,\eta_{i-1},\tau\eta_i,Z_{i+1},\ldots,Z_n).
\]
\begin{enumerate}

\item Direct computation implies that 
\begin{multline*}
\partial^3_{i,i,i} G(\tilde T_i^0) = 3\alpha^2 F'''\l( \sum_{j\ne i}\Big(\eta_j + \alpha \delta_j  \Big) + \tau (\eta_i + \alpha \delta_i)\r) 
\\
+ \alpha^3 F''''\l( \sum_{j\ne i} \Big( \eta_j + \alpha \delta_j \Big) + \tau (\eta_i + \alpha \delta_i )\r) \Big( \sum_{j\ne i}\delta_j + \tau \delta_i \Big),
\end{multline*}
hence
\begin{multline}
\label{eq:der-1}
\mb E \Big| \partial^3_{i,i,i} G(\tilde T_i^0)\cdot \delta_i^3 \Big| \leq 
3\alpha^2 \| F''' \|_\infty \mb E|\delta_i^3| + 
\alpha^3 \| F'''' \|_\infty \l( \mb E\big|\sum_{j\ne i}\delta_j \big| \, \mb E|\delta_i|^3 + \mb E|\delta_i|^4\r)
\\
\leq 3\alpha^2 \| F''' \|_\infty \l(\mb E\delta_i^2 \r)^{1/2} \l( \mb E \delta_i^4\r)^{1/2} + 
\alpha^3 \| F'''' \|_\infty \l(\sqrt{\sum_{j\ne i} \mb E\delta_j^2} \, \l(\mb E\delta_i^2 \r)^{1/2} \l( \mb E \delta_i^4\r)^{1/2} + \mb E|\delta_i|^4\r),
\end{multline}
where we used H\"{o}lder's inequality in the last step.

\item Next, 
\begin{equation*}
\partial^3 G_{\eta_i,\eta_i,\eta_i}(\tilde T_i^0) = F''''\l( \sum_{j\ne i} \Big( \eta_j + \alpha \delta_j \Big) + \tau (\eta_i + \alpha \delta_i) \r) \Big( \sum_{j\ne i}\delta_j + \tau \delta_i  \Big),
\end{equation*}
hence H\"{o}lder's inequality, together with the identity $\| F'''' \|_\infty = M^{-3}\| H'''' \|_\infty$, imply that
\begin{multline}
\label{eq:der-2}
\mb E\Big| \partial^3_{n+i,n+i,n+i} G(\tilde T_i^0)\cdot \eta_i^3 \Big| \leq 
\| F'''' \|_\infty \l( \mb E|\eta_i|^3 \mb E \l|  \sum_{j\ne i}\delta_j \r| + \mb E |\delta_i \eta_i^3| \r)  
\\
\leq \| F'''' \|_\infty \l( \mb E|\eta_i|^3 \sqrt{\sum_{j\ne i } \mb E\delta_j^2} + \l(\mb E\delta_i^4\r)^{1/4} \l(\mb E \eta_i^4\r)^{3/4}\r).
\end{multline}

\item Proceeding in a similar fashion, we deduce that
\begin{multline*}
\partial^3 G_{n+i,n+i,i}(\tilde T_i^0) = F'''\l( \sum_{j\ne i}\Big(\eta_j + \alpha \delta_j \Big) + \tau (\eta_i + \alpha \delta_i)\r) 
\\
+ \alpha F''''\l( \sum_{j\ne i} \Big( \eta_j + \alpha \delta_j\Big) + \tau (\eta_i + \alpha \delta_i)\r) \Big( \sum_{j\ne i}\delta_j + \tau \delta_i \Big),
\end{multline*}
so that, applying H\"{o}lder's inequality, we obtain 
\begin{multline}
\label{eq:der-3}
\mb E\Big| \partial^3_{n+i,n+i,i} G(\tilde T_i^0)\cdot \eta_i^2 \delta_i \Big| \leq 
\| F''' \|_\infty \l(\mb E \eta_i^4\r)^{1/2} \l( \mb E \delta_i^2 \r)^{1/2}
+ \alpha \| F'''' \|_\infty \mb E\bigg| \eta_i^2\delta_i \bigg( \sum_{j\ne i }\delta_j + \tau\delta_i \bigg) \bigg|  
\\
\leq \| F''' \|_\infty \l(\mb E \eta_i^4\r)^{1/2} \l( \mb E \delta_i^2 \r)^{1/2} + \alpha \| F'''' \|_\infty
\l( \sqrt{\sum_{j\ne i}\mb E \delta_j^2} \l( \mb E \eta_i^4\r)^{1/2} \l( \mb E \delta_i^2 \r)^{1/2} + \sqrt{\mb E\delta_i^4 \, \mb E\eta_i^4} \r).
\end{multline}

\item Finally, 
\begin{multline*}
\partial^3 G_{n+i,i,i}(\tilde T_i^0) = 2\alpha F'''\l( \sum_{j\ne i}\Big(\eta_j + \alpha \delta_j \Big) + \tau (\eta_i + \alpha \delta_i)\r) 
\\
+ \alpha^2 F''''\l( \sum_{j\ne i} \Big( \eta_j + \alpha \delta_j\Big) + \tau (\eta_i + \alpha \delta_i)\r) \Big( \sum_{j\ne i}\delta_j + \tau \delta_i \Big).
\end{multline*}
H\"{o}lder's inequality implies that $\mb E|\eta_i \delta_i^2| = \mb E |\eta_i\delta_i \delta_i|\leq \l(\mb E \delta_i^2 \r)^{1/2} \l( \mb E \delta_i^4\r)^{1/4} \l( \mb E \eta_i^4\r)^{1/4}$, hence
\begin{multline}
\label{eq:der-4}
\l| \mb E \partial^3_{n+i,i,i} G(\tilde T_i^0)\cdot \eta_i \delta_i^2 \r| \leq 
2\alpha \| F''' \|_\infty \l(\mb E \delta_i^2 \r)^{1/2} \l( \mb E \delta_i^4\r)^{1/4} \l( \mb E \eta_i^4\r)^{1/4} 
+ \alpha^2 \| F'''' \|_\infty \mb E\Big|  \eta_i \delta_i^2 \Big( \sum_{j\ne i}\delta_j + \tau \delta_i \Big) \Big|
\\
\leq 2\alpha \| F''' \|_\infty \l(\mb E \delta_i^2 \r)^{1/2} \l( \mb E \delta_i^4\r)^{1/4} \l( \mb E \eta_i^4\r)^{1/4}
\\
 + \alpha^2 \| F'''' \|_\infty \l( \sqrt{\sum_{j\ne i}\mb E \delta_j^2} \l(\mb E \delta_i^2 \r)^{1/2} \l( \mb E \delta_i^4\r)^{1/4} \l( \mb E \eta_i^4\r)^{1/4} 
+ \l( \mb E \delta_i^4\r)^{3/4} \l( \mb E \eta_i^4\r)^{1/4}\r).
\end{multline}
Similar calculations yield an analogous bound for the terms in the expansion \eqref{eq:gaussian} of $G(T_{i-1})$. The equivalence of the moments of Gaussian random variables together with the fact that the covariance structure of $(W,Z)$ matches that of $(\xi_1,\eta_1)$ imply that the upper bounds \eqref{eq:der-1},\eqref{eq:der-2},\eqref{eq:der-3},\eqref{eq:der-4} remain valid for the terms in \eqref{eq:gaussian}, up to an additional absolute multiplicative constant. 
Hence, combination of \eqref{eq:telescopic}, \eqref{eq:der-1},\eqref{eq:der-2},\eqref{eq:der-3}, \eqref{eq:der-4} and straightforward application of H\"{o}lder's inequality yields the result.
\end{enumerate}

\subsection{Proof of Lemma \ref{lemma:modulus}.}
\label{proof:modulus}

Define 
\begin{equation*}
D(\delta):=\sup_{\ell(f)\in \m F(\delta)} \mb E^{1/2}\l( \rho'\l( \sqrt{n}\frac{ \bL_j(f) - \m L(f) }{\Delta} \r) - 
\rho'\l( \sqrt{n}\frac{ \bL_j(f_\ast) - \m L(f_\ast) }{\Delta} \r)\r)^2.
\end{equation*}
Recall that $\rho'$ is Lipschitz continuous and $L(\rho')=1$, hence 
\begin{multline}
\label{eq:var-diff}
\l( \rho'\l( \sqrt{n}\frac{ \bL_1(f) - \m L(f) }{\Delta} \r) - 
\rho'\l( \sqrt{n}\frac{ \bL_1(f_\ast) - \m L(f_\ast) }{\Delta} \r)\r)^2 
\\
\leq \l( \sqrt{n}\frac{\bL_1(f) - \bL_1(f_\ast) - (\m L(f) - \m L(f_\ast))}{\Delta}\r)^2 ,
\end{multline}
which implies that 
\begin{equation}
\label{eq:D-bound}
D(\delta)\leq \frac{\nu(\delta)}{\Delta} .
\end{equation}
Next, observe that $\hGk(0;f) = \frac{1}{\sqrt{k}}\sum_{j\in J} \rho'\l( \sqrt{n}\,\frac{ (\bL_j(f) - \m L(f) ) - z}{\Delta} \r) + \frac{1}{\sqrt{k}}\sum_{j\notin J} \rho'\l( \sqrt{n}\,\frac{ (\bL_j(f) - \m L(f) ) - z}{\Delta} \r)$, hence application of the triangle inequality yields that 
\begin{multline}
\label{eq:base-triangle}
\sup_{f\in \m F(\delta)}\l| \hGk(0;f) - \hGk(0;f_\ast)  \r| \leq \sup_{f\in \m F(\delta)}\l| \Gk(0;f) - \Gk(0;f_\ast) \r|
\\
+ \sqrt{\frac{|J|}{k}}\sup_{f\in \m F(\delta)}\l| \hGj(0;f) - \hGj(0;f_\ast) - \mb E\l( \hGj(0;f) - \hGj(0;f_\ast)\r)  \r| 
 + 4\frac{\m O}{\sqrt k}, 
\end{multline}
where $\hGj(0;f):=\frac{1}{\sqrt{|J|}}\sum_{j\in J} \rho'\l( \sqrt{n}\,\frac{ (\bL_j(f) - \m L(f) ) - z}{\Delta} \r)$. 
Talagrand's concentration inequality (specifically, the bound of Lemma \ref{lemma:klein}) implies, together with the inequalities $\l\| \rho' \r\|_\infty\leq 2$ and $|J|>k/2$, that for any $s>0$
\begin{multline}
\label{eq:talagrand}
\sup_{f\in \m F(\delta)}\l| \hGj(0;f) - \hGj(0;f_\ast) - \mb E\l( \hGj(0;f) - \hGj(0;f_\ast)\r) \r| \leq 
\\
2\bigg[ \mb E\sup_{f\in \m F(\delta)} \l| \hGj(0;f) - \hGj(0;f_\ast) - \mb E\l( \hGj(0;f) - \hGj(0;f_\ast)\r) \r|
+ D(\delta)\sqrt{\frac{s}{2}} + \frac{32\sqrt{2}s}{3\sqrt{k}}\bigg]
\end{multline}
with probability at least $1-2e^{-s}$. 
According to \eqref{eq:D-bound}, $D(\delta) \leq \frac{L(\rho')}{\Delta}\,\nu(\delta)$. Hence, it remains to estimate the expected supremum. 
Sequential application of symmetrization, contraction and desymmetrization inequalities implies that
\begin{multline}
\label{eq:expectedsup}
\mb E\sup_{f\in \m F(\delta)} \l| \hGj(0;f) - \hGj(0;f_\ast) - \mb E\l( \hGj(0;f) - \hGj(0;f_\ast)\r) \r|
\\
\leq 2\mb E \sup_{f\in \m F(\delta)} \l| \frac{1}{\sqrt{|J|}} \sum_{j\in J} \eps_j \l( \rho'\l( \sqrt{n}\,\frac{ \bL_j(f) - \m L(f) }{\Delta} \r) \r) - \rho'\l( \sqrt{n}\,\frac{ \bL_j(f_\ast) - \m L(f_\ast) }{\Delta} \r)\r| 
\\
\leq \frac{4 L(\rho')}{\Delta} \mb E \sup_{f\in \m F(\delta)} \l| \frac{\sqrt{n}}{\sqrt{|J|}} \sum_{j\in |J|} \eps_j \l( (\bL_j(f) - \m L(f) )(X_j) - (\bL_j(f_\ast) - \m L(f_\ast))(X_j) \r) \r| 
\\
\leq \frac{8\sqrt{2} L(\rho')}{\Delta} \mb E \sup_{f\in \m F(\delta)} \l|  \frac{1}{\sqrt{N}}\sum_{j=1}^{N_J} \Big( (\ell(f) - \ell(f_\ast))(X_j) - P( \ell(f) - \ell(f_\ast)) \Big) \r| \leq \frac{8\sqrt{2}}{\Delta}\, \omega(\delta)
\end{multline}
since $L(\rho')=1$. 
To estimate $\sup_{f\in \m F(\delta)}\l| \Gk(0;f) - \Gk(0;f_\ast) \r|$, we consider 2 cases: the first case when only 2 finite moments of $\ell(f(X)), \ f\in \m F$ exist, and the second case when 4 moments are finite. 
To obtain the bound in the first case, we observe that, since $\mb E \l(\sqrt{n}\frac{\bL_1(f) - \m L(f)}{\Delta}\r) = 0$ for any $f\in \m F$, 
\begin{multline*}
\l| \mb E \rho'\l( \sqrt{n}\frac{\bL_1(f) - \m L(f)}{\Delta} \r) - \mb E \rho'\l( \sqrt{n}\frac{\bL_1(f_\ast) - \m L(f_\ast)}{\Delta} \r) \r| \\
= \l| \mb E T\l(  \sqrt{n}\frac{\bL_1(f) - \m L(f)}{\Delta} \r) - \mb E T\l( \sqrt{n}\frac{\bL_1(f_\ast) - \m L(f_\ast)}{\Delta} \r)\r|
\end{multline*}
where $T(x) = x - \rho'(x)$. Next, we apply Lemma \ref{lemma:bias} with $F = T$, $\xi_j = \frac{\ell(f(X_j) - \mb E\ell(f(X_j)) }{\Delta \sqrt{n}}$ and $\eta_j = \frac{\ell(f_\ast(X_j) - \mb E\ell(f_\ast(X_j)) }{\Delta \sqrt{n}}$. 
The first inequality of the lemma implies that  
\begin{multline*}
\l| \mb E \rho'\l( \sqrt{n}\frac{\bL_1(f) - \m L(f)}{\Delta} \r) - \mb E \rho'\l( \sqrt{n}\frac{\bL_1(f_\ast) - \m L(f_\ast)}{\Delta} \r) \r| 
\leq \sqrt{\var\l( \frac{\ell(f(X)) - \ell(f_\ast(X))}{\Delta} \r)} 
\\
\times\max_{\alpha\in[0,1] } \sqrt{ \mb E \l(T'\l( \alpha  \sqrt{n}\frac{\bL_1(f) - \m L(f)}{\Delta} + (1-\alpha) \sqrt{n}\frac{\bL_1(f_\ast) - \m L(f_\ast)}{\Delta} \r)\r)^2}.
\end{multline*}
Observe that $T'(x) = 1 - \rho''(x)\leq I\l\{ |x|\geq 1\r\}$ by Assumption \ref{ass:1}. It implies that for any $\alpha\in[0,1]$, 
\begin{multline*}
\mb E \l(T'\l( \alpha  \sqrt{n}\frac{\bL_1(f) - \m L(f)}{\Delta} + (1-\alpha) \sqrt{n}\frac{\bL_1(f_\ast) - \m L(f_\ast)}{\Delta} \r)\r)^2 \\
\leq \pr{\l| \alpha  \sqrt{n}\frac{\bL_1(f) - \m L(f)}{\Delta} + (1-\alpha) \sqrt{n}\frac{\bL_1(f_\ast) - \m L(f_\ast)}{\Delta} \r|\geq 1}
\\
\leq \sup_{f\in \m F}\var\l( \sqrt{n}\frac{\bL_1(f) - \m L(f)}{\Delta} \r) = \sup_{f\in \m F}\frac{\sigma^2(\ell,f)}{\Delta^2}. 
\end{multline*}
by Chebyshev's inequality. Hence
\begin{equation*}
\l| \mb E \rho'\l( \sqrt{n}\frac{\bL_1(f) - \m L(f)}{\Delta} \r) - \mb E \rho'\l( \sqrt{n}\frac{\bL_1(f_\ast) - \m L(f_\ast)}{\Delta} \r) \r| 
\leq \var^{1/2}\l( \ell(f(X)) - \ell(f_\ast(X)) \r) \frac{\sigma(\ell,\m F)}{\Delta^2}.
\end{equation*}
and, taking supremum over $f\in \m F(\delta)$ and recalling that $\Delta = M_\Delta \cdot \sigma(\ell,\m F)$ for $M_\Delta\geq 1$, we obtain the inequality
\begin{equation}
\label{eq:bias1}
\sup_{f\in \m F(\delta)}\l| \Gk(0;f) - \Gk(0;f_\ast) \r| \leq \sqrt{k} \frac{\nu(\delta)}{\Delta} \frac{1}{M_\Delta}
\leq \sqrt{k}\, \widetilde B(\delta).
\end{equation}

\noindent On the other hand, under the assumption of existence of 4 moments, we get that 
\begin{multline*}
\l| \mb E \rho'\l( \sqrt{n}\frac{\bL_1(f) - \m L(f)}{\Delta} \r) - \mb E \rho'\l( \sqrt{n}\frac{\bL_1(f_\ast) - \m L(f_\ast)}{\Delta} \r) \r| 
\\
\leq \frac{C(\rho)}{\sqrt{n}\Delta}\bigg(  \var^{1/2}(\ell(f(X)) - \ell(f_\ast(X))) \l( \frac{R_4^2(\ell,\m F)}{\Delta^2}
+ \frac{R_4^3(\ell,\m F)}{\Delta^3}\r)
\\ 
+ \frac{\mb E^{1/4} (\ell(f(X)) - \ell(f_\ast(X)))^4}{\sqrt{n}}\frac{R_4^3(\ell,\m F)}{\Delta^3}\bigg),
\end{multline*}
Again, taking supremum over $f\in \m F(\delta)$ and recalling that $\Delta = M_\Delta \cdot \sigma(\ell,\m F)$ for $M_\Delta\geq 1$, we deduce that    
\begin{multline}
\label{eq:bias-delta}
\sup_{f\in \m F(\delta)}\l| \Gk(0;f) - \Gk(0;f_\ast) \r| \leq 
C(\rho)\sqrt{\frac{k}{n}}\Bigg( \frac{\nu(\delta)}{\Delta} \l( \frac{\mathfrak{B}^3(\ell,\m F)}{M^3_\Delta} \vee \frac{\mathfrak{B}^2(\ell,\m F)}{M^2_\Delta}\r)
+ \frac{\nu_4(\delta)}{\Delta} \frac{\mathfrak{B}^3(\ell,\m F)}{M_\Delta^3 \sqrt{n}}
\Bigg)
\\
\leq C(\rho)\sqrt{\frac{k}{n}} \mathfrak{B}^3(\ell,\m F) \Bigg( \frac{\nu(\delta)}{\Delta}\frac{1}{M_\Delta^2} 
+ \frac{\nu_4(\delta)}{\Delta} \frac{1}{M_\Delta^3 \sqrt{n}}
\Bigg) \leq C(\rho)\sqrt{k} \, \widetilde B(\delta),
\end{multline}
implying the result.

\subsection{Proof of Lemma \ref{lemma:modulus-unif}.}
\label{proof:modulus-unif}

Recall that $\hGj(0;f):=\frac{1}{\sqrt{|J|}}\sum_{j\in J} \rho'\l( \sqrt{n}\,\frac{ (\bL_j(f) - \m L(f) ) - z}{\Delta} \r)$. 
Given $\delta\geq \delta_{\min}$, define
\begin{align*}
\widehat Q_{|J|}(\delta)&:=\sup_{f\in \m F(\delta)}  \frac{\delta_{\min}}{\delta} \l| \hGj(0;f) - \hGj(0;f_\ast) \r|,
\\
\widehat T_{|J|}(\delta_{\min})&:=\sup_{\delta\geq \delta_{\min}} \widehat Q_{|J|}(\delta).
\end{align*}
Observe that for any $\delta \geq \delta_{\min}$,
\begin{equation}
\label{eq:main-1}
\sup_{f\in \m F(\delta)}\l|  \hGj(0;f) - \hGj(0;f_\ast) \r| \leq \frac{\delta}{\delta_{\min}} \, \widehat T_{|J|}(\delta_{\min}).
\end{equation}
Hence, our goal will be to find an upper bound for $\widehat T_{|J|}(\delta_{\min})$. 
To this end, note that 
\begin{multline}
\label{eq:decomp}
\widehat T_{|J|}(\delta_{\min}) \leq \sup_{\delta\geq \delta_{\min}} \sup_{f\in \m F(\delta)} \frac{\delta_{\min}}{\delta} \l| \hGj(0;f) - \hGj(0;f_\ast)  - \mb E\l( \hGj(0;f) - \hGj(0;f_\ast)  \r) \r| 
\\
+ \sup_{\delta\geq \delta_{\min}} \sup_{f\in \m F(\delta)} \frac{\delta_{\min}}{\delta} \l| \Gk(0;f) - \Gk(0;f_\ast)\r|.
\end{multline}
It remains to estimate both terms in the inequality above. 
Inequality \eqref{eq:var-diff} implies the bound
\begin{multline*}
\sup_{\delta\geq \delta_{\min}}\sup_{f\in \m F(\delta)}\frac{\delta_{\min}}{\delta} \var^{1/2}\l( \rho'\l( \sqrt{n}\frac{ \bL_1(f) - \m L(f) }{\Delta} \r) - 
\rho'\l( \sqrt{n}\frac{ \bL_1(f_\ast) - \m L(f_\ast) }{\Delta} \r)\r)
\\
\leq \frac{L(\rho')}{\Delta} \sup_{\delta\geq \delta_{\min}} \frac{\delta_{\min}}{\delta} \nu(\delta) 
\leq \frac{L(\rho')}{\Delta} \sup_{\delta\geq \delta_{\min}} \frac{\delta_{\min}}{\delta} \widetilde\nu(\delta)
\leq \frac{1}{\Delta} \widetilde\nu(\delta_{\min})
\end{multline*}
since $\widetilde \nu$ is a function of concave type. Moreover, it is clear that for any $\delta\geq \delta_{\min}$,
\[
\l| \frac{\delta_{\min}}{\delta} \rho'\l( \sqrt{n}\frac{ \bL_1(f) - \m L(f) }{\Delta} \r) - \frac{\delta_{\min}}{\delta} \rho'\l( \sqrt{n}\frac{ \bL_1(f_\ast) - \m L(f_\ast) }{\Delta} \r) \r| \leq 2\l\| \rho'\r\|_{\infty} \leq 4
\]
almost surely. 
Now, Talagrand's concentration inequality implies that for any $s>0$, 
\begin{multline}
\label{eq:talagrand2}
\sup_{\delta\geq \delta_{\min}} \sup_{f\in \m F(\delta)} \frac{\delta_{\min}}{\delta} \l| \hGj(0;f) - \hGj(0;f_\ast)  - \mb E\l( \hGj(0;f) - \hGj(0;f_\ast) \r)  \r| 
\\
\leq 2\Big[\mb E\sup_{\delta\geq \delta_{\min}} \sup_{f\in \m F(\delta)} \frac{\delta_{\min}}{\delta} \l| \hGj(0;f) - \hGj(0;f_\ast)  - \mb E\l( \hGj(0;f) - \hGj(0;f_\ast) \r) \r| 
\\
+ \frac{L(\rho')}{\Delta} \widetilde\nu(\delta_{\min})\sqrt{\frac{s}{2}} + \frac{32\sqrt{2} s}{3\sqrt{k}} \bigg]
\end{multline}
with probability at least $1 - e^{-s}$. To estimate the expectation, we proceed as follows: for $j\in \mb Z$, set $\delta_j:=2^{-j}$, and observe that 
\begin{multline*}
\mb E\sup_{\delta\geq \delta_{\min}} \sup_{f\in \m F(\delta)} \frac{\delta_{\min}}{\delta} \l| \hGj(0;f) - \hGj(0;f_\ast)  - \mb E\l( \hGj(0;f) - \hGj(0;f_\ast) \r) \r|
\\
\leq \mb E\sup_{j:\delta_j \geq \delta_{\min}} \sup_{\delta\in(\delta_{j+1},\delta_j]} \frac{\delta_{\min}}{\delta}
 \sup_{f\in \m F(\delta)} \l| \hGj(0;f) - \hGj(0;f_\ast)  - \mb E\l( \hGj(0;f) - \hGj(0;f_\ast) \r) \r| 
\\
\leq  \sum_{j:\delta_j\geq \delta_{\min}}  \frac{\delta_{\min}}{\delta_{j+1}} \, \mb E \sup_{\delta\in(\delta_{j+1},\delta_j]}
 \sup_{f\in \m F(\delta)} \l| \hGj(0;f) - \hGj(0;f_\ast)  - \mb E\l( \hGj(0;f) - \hGj(0;f_\ast) \r) \r| 
 \\
 \leq 2 \sum_{j:\delta_j\geq \delta_{\min}}  \frac{\delta_{\min}}{\delta_{j}} \, 
 \mb E  \sup_{f\in \m F(\delta_j)} \l| \hGj(0;f) - \hGj(0;f_\ast)  - \mb  E\l( \hGj(0;f) - \hGj(0;f_\ast) \r) \r|, 
\end{multline*}
where the last inequality relied on the fact that $\m F(\delta)\subseteq \m F(\delta')$ for $\delta\leq\delta'$. 
It follows from \eqref{eq:expectedsup} that
\[
\mb E  \sup_{f\in \m F(\delta_j)} \l| \hGj(0;f) - \hGj(0;f_\ast)  - \mb E\l( \hGj(0;f) - \hGj(0;f_\ast) \r) \r| \leq \frac{8\sqrt{2} L(\rho')}{\Delta} \omega(\delta_j) \leq \frac{8\sqrt{2}}{\Delta}\, \widetilde \omega(\delta_j), 
\]
where $\widetilde \omega(\cdot)$ is an upper bound on $\omega(\cdot)$ of strictly concave type (with exponent $\gamma$ for some $\gamma\in(0,1)$). Hence, applying Proposition 4.2 in \cite{Koltchinskii2011Oracle-inequali00}, we deduce that  
\begin{multline*}
\mb E\sup_{\delta\geq \delta_{\min}} \sup_{f\in \m F(\delta)} \frac{\delta_{\min}}{\delta} \l| \hGj(0;f) - \hGj(0;f_\ast) - \mb E\l( \hGj(0;f) - \hGj(0;f_\ast) \r) \r| 
\\
\leq \frac{16}{\Delta} \delta_{\min} \sum_{j:\delta_j\geq \delta_{\min}}  \frac{ \widetilde \omega(\delta_j)}{\delta_{j}} 
\leq \frac{c(\gamma)}{\Delta} \delta_{\min} \frac{\widetilde\omega(\delta_{\min})}{\delta_{\min}} 
=  \frac{c(\gamma)}{\Delta} \, \widetilde\omega(\delta_{\min}),
\end{multline*}
and \eqref{eq:talagrand2} yields the inequality
\begin{equation}
\label{eq:uniform-1}
\sup_{\delta\geq \delta_{\min}} \sup_{f\in \m F(\delta)} \frac{\delta_{\min}}{\delta} \l| \hGj(0;f) - \hGj(0;f_\ast) - \mb E\l( \hGj(0;f) - \hGj(0;f_\ast) \r)  \r| 
\leq \widetilde U(\delta_{\min},s),
\end{equation}
where $\widetilde U(\delta,s)$ was defined in \eqref{eq:U-tilde}. 
For the second term in \eqref{eq:decomp}, inequality \eqref{eq:bias-delta} implies that
\begin{multline*}
\sup_{\delta\geq \delta_{\min}} \sup_{f\in \m F(\delta)} \frac{\delta_{\min}}{\delta} \l| \Gk(0;f) - \Gk(0;f_\ast)\r| 
\\
\leq C(\rho)\delta_{\min}\sqrt{\frac{k}{n}}\widebar R^3(\ell,\m F,\Delta)
\sup_{\delta\geq \delta_{\min}}
\Bigg( \frac{\nu(\delta)}{\delta \Delta}\frac{1}{M_\Delta^2}
+ \frac{\nu_4(\delta)}{\delta\,\Delta} \frac{1}{M_\Delta^3 \sqrt{n}}\Bigg)
\\
\leq C(\rho)\sqrt{k} \,\mathfrak{B}^3(\ell,\m F)
\l(  \frac{\widetilde\nu(\delta_{\min})}{\Delta} \frac{1}{M_\Delta^2}
+ \frac{\widetilde\nu_4(\delta_{\min})}{\Delta} \frac{1}{M_\Delta^3 \sqrt{n}}\r)
\end{multline*}
since $\nu(\delta)\leq \widetilde \nu(\delta), \ \nu_4(\delta)\leq \widetilde \nu_4(\delta)$ and $\widetilde \nu(\delta), \  \widetilde \nu_4(\delta)$ are functions of concave type. 
Combining the bound above with \eqref{eq:uniform-1}, we deduce that 
\[
\widehat T_{|J|}(\delta_{\min}) \leq \widetilde U(\delta_{\min},s) + C(\rho)\sqrt{k} \widetilde B(\delta_{\min}),
\]
hence \eqref{eq:base-triangle} and \eqref{eq:main-1} imply that for all $\delta\geq\delta_{\min}$ simultaneously,
\[
\sup_{f\in \m F(\delta)}\l|  \hGk(0;f) - \hGk(0;f_\ast) \r| \leq C(\rho) \delta \l( \frac{\widetilde U(\delta_{\min},s)}{\delta_{\min}} + \sqrt{k} \frac{\widetilde B(\delta_{\min})}{\delta_{\min}} \r) + 4\frac{\m O}{\sqrt k}
\]
with probability at least $1-e^{-s}$. 

\subsection{Proof of Lemma \ref{lemma:R_N}.}
\label{proof:R_N}

The following identity is immediate: 
\[
R_N(f) = \underbrace{\hGk\l(\hEk(f);f \r)}_{=0} +\, \partial_z \Gk\l(0; f \r) \cdot  \hEk(f) -
\l(  \hGk\l(\hEk(f);f \r) - \hGk\l(0;f \r) \r).
\]
Assumptions on $\rho$ imply that for any $f\in \m F$ and $j=1,\ldots,k$, there exists $\tau_j\in[0,1]$ such that 
\begin{multline}
\label{eq:R_N-1}
\rho'\l( \sqrt{n}\frac{ \bL_j(f) - \m L(f) - \hEk(f)}{\Delta} \r) = 
\rho'\l( \sqrt{n}\frac{ \bL_j(f) - \m L(f)}{\Delta} \r) 
 - \frac{\sqrt{n}}{\Delta} \rho''\l( \sqrt{n}\frac{ \bL_j(f) - \m L(f)}{\Delta} \r) \cdot \hEk(f) 
 \\
 + \frac{n}{\Delta^2} \rho''' \l( \sqrt{n}\frac{ \bL_j(f) - \m L(f) - \tau_j \hEk(f) }{\Delta} \r)\cdot \l(\hEk(f) \r)^2,
\end{multline}
hence 
\begin{multline*}
 \hGk\l(\hEk(f);f \r) - \hGk\l(0;f \r) = 
- \frac{\sqrt{n}}{\Delta}\frac{\hEk(f) }{\sqrt{k}}\sum_{j=1}^k  \rho''\l( \sqrt{n}\frac{ \bL_j(f) - \m L(f)}{\Delta} \r) 
\\
+\frac{n}{\Delta^2} \frac{\l(\hEk(f) \r)^2}{\sqrt{k}}\sum_{j=1}^k  \rho''' \l( \sqrt{n}\frac{ \bL_j(f) - \m L(f) - \tau_j \hEk(f)}{\Delta} \r) ,
\end{multline*}
and 
\begin{multline}
\label{eq:R_N-2}
R_N(f) =  \frac{\sqrt{n}}{\Delta}\frac{\hEk(f) }{\sqrt{k}}\sum_{j=1}^k  \l( \rho''\l( \sqrt{n}\frac{ \bL_j(f) - \m L(f)}{\Delta} \r) - \mb E \rho''\l( \sqrt{n}\frac{ \bL_j(f) - \m L(f)}{\Delta} \r) \r)
 \\
- \frac{n}{\Delta^2} \frac{\l(\hEk(f) \r)^2}{\sqrt{k}}\sum_{j=1}^k \rho''' \l( \sqrt{n}\frac{ \bL_j(f) - \m L(f) - \tau_j \hEk(f)}{\Delta} \r). 
\end{multline}

\noindent We will need the following modification of Theorem \ref{th:unif} that is stated below and proved in Section \ref{proof:unif-delta-bak}. 
\begin{lemma}
\label{th:unif-delta-bak}
Then there exist positive constants $c(\rho),\, C(\rho)$ with the following properties. Fix $\delta_{\min}>0$. Then for all $s>0$, $\delta\geq \delta_{\min}$, positive integers $n$ and $k$ such that 
\begin{equation}
\label{eq:key}
\delta \, \frac{\widetilde U(\delta_{\min},s)}{\delta_{\min}\sqrt{k}} +
\sup_{f\in\m F} G_f(n,\Delta)
+ \frac{s+\m O}{k} \leq c(\rho),
\end{equation}
the following inequality holds with probability at least $1 - 2e^{-s}$:
\begin{multline}
\label{eq:slow-rates2}
\sup_{f\in \m F(\delta)}\l| \hEk(f) \r| \leq 
C(\rho) \widetilde \Delta  
\Bigg[\frac{\delta}{\sqrt{N}} \, \frac{\widetilde U(\delta_{\min},s)}{\delta_{\min}} 
+ \frac{\sigma(\ell,f_\ast)}{\Delta}\sqrt{\frac{s}{N}} 
+ \frac{\sup_{f\in\m F} G_f(n,\Delta)}{\sqrt{n}} +  \frac{(s+\m O)\sqrt{n}}{N} \Bigg].
\end{multline}
\end{lemma}
In the rest of the proof, we will assume that conditions of Lemma \ref{th:unif-delta-bak} and Theorem \ref{th:unif} hold, and let $\Theta'$ be an event of probability at least $1 - 4e^{-s}$ on which inequalities \eqref{eq:slow-rates2} and \eqref{eq:slow-rates1} are valid. 
On event $\Theta'$, the last term in \eqref{eq:R_N-2} can thus be estimated as 
\begin{multline}
\label{eq:R_N-2.1}
\sup_{f\in \m F(\delta)} \l| \frac{n}{\Delta^2} \frac{\l(\hEk(f) \r)^2}{\sqrt{k}}\sum_{j=1}^k  \rho''' \l( \sqrt{n}\frac{ \bL_j(f) - \m L(f) - \tau_j \hEk(f)}{\Delta} \r) \r| \leq 
C_1(\rho) \frac{\sqrt{nN}}{\Delta^2} \sup_{f\in \m F(\delta)}\l| \hEk(f) \r|^2 
\\
\leq C_2(\rho) \sqrt{N} \frac{\widetilde \Delta^2}{\Delta^2} \bigg( \frac{n^{1/2}\delta^2}{N}\l(\frac{\widetilde U(\delta_{\min},s)}{\delta_{\min}}\r)^2 \bigvee \frac{\sigma^2(\ell,f_\ast)}{\Delta^2} \frac{n^{1/2}\, s}{N}
\\ 
\bigvee n^{1/2}\l(\sup_{f\in \m F}  \frac{G_f\big(n,\Delta\big)}{\sqrt{n}}\r)^2 \bigvee n^{3/2} \frac{s^2+\m O^2}{N^2}
\bigg).
\end{multline}
where we used the fact that $\|\rho'''\|_\infty<\infty$. It remains to estimate the first term in \eqref{eq:R_N-2}. The required bound will follow from the combination of Theorem \ref{th:unif-delta-bak} and the following lemma that is proved in Section \ref{proof:simple}.
\begin{lemma}
\label{lemma:simple}
Fix $\delta_{\min}>0$. With probability at least $ 1 - 3e^{-s}$, for all $\delta\geq \delta_{\min}$ simultaneously,
\begin{multline*}
\sup_{f\in \m F(\delta)}\l|\frac{1}{\sqrt{k}}\sum_{j=1}^k  \l( \rho''\l( \sqrt{n}\frac{ \bL_j(f) - \m L(f)}{\Delta} \r) - \mb E \rho''\l( \sqrt{n}\frac{ \bL_j(f) - \m L(f)}{\Delta} \r) \r) \r| 
\\
\leq C(\rho) \l( \delta\,\frac{\widetilde U(\delta_{\min},s)}{\delta_{\min}} + \frac{\sigma(\ell,f_\ast)}{\Delta }\sqrt{s} + \frac{s+\m O}{\sqrt k}\r).
\end{multline*}
\end{lemma}
\noindent Let $\Theta''$ be the event of probability at least $1-3e^{-2s}$ on which the inequality of Lemma \ref{lemma:simple} holds. Then simple algebra yields that on event $\Theta'\cap \Theta''$ of probability at least $1-7e^{-s}$,
\begin{multline}
\label{eq:R_N-2.3}
\sup_{f\in \m F(\delta)} \l|\frac{\sqrt{n}}{\Delta}\frac{\hEk(f) }{\sqrt{k}}\sum_{j=1}^k  \l( \rho''\l( \sqrt{n}\frac{ \bL_j(f) - \m L(f)}{\Delta} \r) - \mb E \rho''\l( \sqrt{n}\frac{ \bL_j(f) - \m L(f)}{\Delta} \r) \r) \r|
\\
\leq C_3(\rho)\sqrt{N} \frac{\widetilde \Delta}{\Delta} \bigg( \frac{n^{1/2}\delta^2}{N}\l(\frac{\widetilde U(\delta_{\min},s)}{\delta_{\min}}\r)^2 \bigvee \frac{\sigma^2(\ell,f_\ast)}{\Delta^2} \frac{n^{1/2}\, s}{N} 
\\
\bigvee n^{1/2}\l(\sup_{f\in \m F}  \frac{G_f\big(n,\Delta\big)}{\sqrt{n}}\r)^2 \bigvee n^{3/2} \frac{s^2 + \m O^2}{N^2} \bigg).
\end{multline}
Combination of inequalities \eqref{eq:R_N-2.1} and \eqref{eq:R_N-2.3} that hold with probability at leat $1 - 7e^{-s}$ yields the result.


\subsection{Proof of Lemma \ref{th:unif-delta-bak}.}
\label{proof:unif-delta-bak}

In the situation when $\delta$ is fixed, the argument mimics the proof of Theorem 4.1 in \citep{minsker2018uniform}, with minor modifications outlined below. 
Recall that
\[
\hGk(z;f) = \frac{1}{\sqrt{k}}\sum_{j=1}^k \rho'\l( \sqrt{n}\,\frac{ (\bL_j(f) - \m L(f) ) - z}{\Delta} \r).
\]
Let $z_1,z_2$ be such that on an event of probability close to $1$, $\hGk(z_1;f) > 0$ and $\hGk(z_2;f) < 0$ for all $f\in \m F(\delta)$ simultaneously. 
Since $\hGk$ is decreasing in $z$, it is easy to see that $\hEk(f)\in (z_1,z_2)$ for all $f\in \m F(\delta)$ on this event. 
Hence, our goal is to find $z_1,z_2$ satisfying conditions above and such that $|z_1|,\, |z_2|$ are as small as possible. 
Observe that 
\begin{equation*}
\hGk(z;f) = \frac{1}{\sqrt{k}}\sum_{j\in J} \rho'\l( \sqrt{n}\,\frac{ (\bL_j(f) - \m L(f) ) - z}{\Delta} \r) + \frac{1}{\sqrt{k}}\sum_{j\notin J} \rho'\l( \sqrt{n}\,\frac{ (\bL_j(f) - \m L(f) ) - z}{\Delta} \r)
\end{equation*}
and $\l|\frac{1}{\sqrt{k}}\sum_{j\notin J} \rho'\l( \sqrt{n}\,\frac{ (\bL_j(f) - \m L(f) ) - z}{\Delta} \r)\r| \leq 2\frac{\m O}{\sqrt k}$. 
Moreover,
\begin{multline*}
\frac{1}{\sqrt{k}}\sum_{j\in J} \rho'\l( \sqrt{n}\,\frac{ (\bL_j(f) - \m L(f) ) - z}{\Delta} \r) 
\\
=  \frac{1}{\sqrt{k}}\sum_{j\in J} \Big( \rho'\l( \sqrt{n}\,\frac{ (\bL_j(f) - \m L(f) ) - z}{\Delta} \r) 
- \rho'\l( \sqrt{n}\,\frac{ (\bL_j(f_\ast) - \m L(f_\ast) ) - z}{\Delta} \r) 
\\
- \mb E\l[\rho'\l( \sqrt{n}\,\frac{ (\bL_j(f) - \m L(f) ) - z}{\Delta} \r) -\rho'\l( \sqrt{n}\,\frac{ (\bL_j(f_\ast) - \m L(f_\ast) ) - z}{\Delta} \r)\r] \Bigg)  
\\
+  \frac{1}{\sqrt{k}}\sum_{j\in J} \Big( \rho'\l( \sqrt{n}\,\frac{ (\bL_j(f_\ast) - \m L(f_\ast) ) - z}{\Delta} \r) - \mb E  \rho'\l( \sqrt{n}\,\frac{ (\bL_j(f_\ast) - \m L(f_\ast) ) - z}{\Delta} \r) \Bigg)
\\
+ \frac{1}{\sqrt{k}}\sum_{j\in J} \l( \mb E \rho'\l( \sqrt{n}\,\frac{ (\bL_j(f) - \m L(f) ) - z}{\Delta} \r) - 
\mb E \rho'\l( \frac{W(\ell(f)) - \sqrt{n} z }{\Delta} \r) \r) 
\\
+  \frac{1}{\sqrt{k}}\sum_{j\in J} \mb E \rho'\l( \frac{W(\ell(f)) - \sqrt{n} z }{\Delta} \r).
\end{multline*}
We will proceed in 4 steps: first, we will find $\eps_1>0$ such that for any $z\in \mb R$ and all $f\in \m F(\delta)$,
\begin{multline*}
\frac{1}{\sqrt{k}}\sum_{j\in J} \Big( \rho'\l( \sqrt{n}\,\frac{ (\bL_j(f) - \m L(f) ) - z}{\Delta} \r) 
- \rho'\l( \sqrt{n}\,\frac{ (\bL_j(f_\ast) - \m L(f_\ast) ) - z}{\Delta} \r) 
\\
- \mb E\l[\rho'\l( \sqrt{n}\,\frac{ (\bL_j(f) - \m L(f) ) - z}{\Delta} \r) -\rho'\l( \sqrt{n}\,\frac{ (\bL_j(f_\ast) - \m L(f_\ast) ) - z}{\Delta} \r)\r] \Bigg)  \leq \eps_1
\end{multline*}
with high probability, then $\eps_2>0$ such that 
\begin{equation*}
 \frac{1}{\sqrt{k}}\sum_{j\in J} \Big( \rho'\l( \sqrt{n}\,\frac{ (\bL_j(f_\ast) - \m L(f_\ast) ) - z}{\Delta} \r) - \mb E  \rho'\l( \sqrt{n}\,\frac{ (\bL_j(f_\ast) - \m L(f_\ast) ) - z}{\Delta} \r) \Bigg) \leq \eps_2,
\end{equation*}
$\eps_3$ satisfying 
\begin{equation*}
\sup_{f\in \m F(\delta)}\l| \frac{1}{\sqrt{k}}\sum_{j \in J} \l( \mb E \rho'\l( \sqrt{n}\,\frac{ (\bL_j(f) - \m L(f) ) - z}{\Delta} \r) - 
\mb E \rho'\l( \frac{W(\ell(f)) - \sqrt{n} z }{\Delta} \r) \r)\r| \leq \eps_3,
\end{equation*}
and finally we will choose $z_1<0$ such that for all $f\in \m F(\delta)$,
\begin{equation}
\label{eq:z1}
 \frac{1}{\sqrt{k}}\sum_{j\in J} \mb E \rho'\l( \frac{W(\ell(f)) - \sqrt{n} z }{\Delta} \r) > \eps_1 + \eps_2+ \eps_3 + 2\frac{\m O}{\sqrt k}.
\end{equation}
Talagrand's concentration inequality \cite[][Corollary 16.1]{van2016estimation}, together with the bound $\|\rho'\|_\infty\leq 2$, implies that for any $s>0$, 
\begin{multline*}
\sqrt{\frac{|J|}{k}}\sup_{f\in \m F(\delta)}\l| \hGj(z;f) - \hGj(z;f_\ast) - \mb E\l( \hGj(z;f) - \hGj(z;f_\ast) \r) \r| \leq 
\\
2\bigg[ \mb E\sup_{f\in \m F(\delta)} \l| \hGj(z;f) - \hGj(z;f_\ast) - \mb E\l( \hGj(z;f) - \hGj(z;f_\ast) \r) \r|
+ D(\delta)\sqrt{\frac{s}{2}} + \frac{32}{3}\frac{s}{\sqrt{k}}\bigg]
\end{multline*}
with probability at least $1-2e^{-s}$. It has been observed in \eqref{eq:D-bound} that $D(\delta) \leq \frac{\nu(\delta)}{\Delta}$. It remains to estimate the expected supremum. 
Sequential application of symmetrization, contraction and desymmetrization inequalities, together with the fact that $L(\rho')=1$, implies that
\begin{multline*}
\mb E\sup_{f\in \m F(\delta)} \l| \hGj(z;f) - \hGj(z;f_\ast) - \mb E\l( \hGj(z;f) - \hGj(z;f_\ast) \r) \r|
\\
\leq 2\mb E \sup_{f\in \m F(\delta)} \l| \frac{1}{\sqrt{|J|}} \sum_{j\in J} \eps_j \l( \rho'\l( \sqrt{n}\,\frac{ \bL_j(f) - \m L(f) - z }{\Delta} \r) \r) - \rho'\l( \sqrt{n}\,\frac{ \bL_j(f_\ast) - \m L(f_\ast) - z }{\Delta} \r)\r| 
\\
\leq \frac{4}{\Delta} \mb E \sup_{f\in \m F(\delta)} \l| \frac{\sqrt{n}}{\sqrt{|J|}} \sum_{j\in J}  \eps_j \l( (\bL_j(f) - \m L(f) ) - (\bL_j(f_\ast) - \m L(f_\ast)) \r) \r| 
\\
\leq \frac{8\sqrt{2}}{\Delta} \mb E \sup_{f\in \m F(\delta)} \l|  \frac{1}{\sqrt{N}}\sum_{j=1}^{N_J} \Big( (\ell(f) - \ell(f_\ast))(X_j) - P( \ell(f) - \ell(f_\ast)) \Big) \r| \leq \frac{8\sqrt{2}}{\Delta}\, \omega(\delta).
\end{multline*}
Hence, it suffices to choose 
\[
\eps_1 =  \frac{8\sqrt{2}}{\Delta}\, \omega(\delta) + \frac{\nu(\delta)}{\Delta}\,\sqrt{s} + \frac{32}{3}\frac{s}{\sqrt{k}}.
\]
Next, Bernstein's inequality and Lemma \ref{lemma:variance} together yield that with probability at least $1 - 2e^{-s}$, 
\begin{multline*}
\frac{1}{\sqrt{k}}\sum_{j \in J} \Big( \rho'\l( \sqrt{n}\,\frac{ (\bL_j(f_\ast) - \m L(f_\ast) ) - z}{\Delta} \r) - \mb E  \rho'\l( \sqrt{n}\,\frac{ (\bL_j(f_\ast) - \m L(f_\ast) ) - z}{\Delta} \r) \Bigg) 
\\
\leq 2 \l( \frac{\sigma(\ell,f_\ast)}{\Delta}\sqrt{s} + \frac{3s}{\sqrt{k}}\r),
\end{multline*}
thus we can set $\eps_2 = 2 \l( \frac{\sigma(\ell,f_\ast)}{\Delta}\sqrt{s} + 3\frac{s}{\sqrt{k}} \r)$. 
Lemma \ref{lemma:step2} impliies that $\eps_3$ can be chosen as
\[
\eps_3 = \sqrt{k}\,\sup_{f\in\m F(\delta)} G_f(n,\Delta).
\]
Finally, we apply Lemma 6.3 of \citep{minsker2018uniform} with 
\[
\eps := \eps_1 + \eps_2 + \eps_3 + 2\frac{\m O}{\sqrt k}
\]
to deduce that 
\[
z_1 = - C \frac{\widetilde \Delta}{\sqrt{N}}
\cdot \l( \eps_1 + \eps_2 + \eps_3 + 2\frac{\m O}{\sqrt k}\r),
\]
satisfies \eqref{eq:z1} under assumption that 
$\frac{\eps_1 + \eps_2 + \eps_3}{\sqrt k} + \frac{\m O}{k}\leq c$ for some absolute constants $c,C>0$.
Proceeding in a similar way, it is easy to see that setting $z_2=-z_1$ guarantees that $\hGk(z_2;f)<0$ for all $f\in \m F(\delta)$ with probability at least $1 - e^{-s}$, hence the claim follows.

It remains to make the bound uniform in $\delta\geq \delta_{\min}$. To this end, we need to repeat the ``slicing argument'' of Lemma \ref{lemma:modulus-unif} below (specifically, see equation \eqref{eq:uniform-1}) to deduce that with probability at least $1 - 2e^{-s}$,
\begin{equation*}
\sup_{f\in \m F(\delta)}\l| \hGj(z;f) - \hGj(z;f_\ast) - \mb E\l( \hGj(z;f) - \hGj(z;f_\ast) \r) \r| \leq
\delta \, \frac{\widetilde U(\delta_{\min},s)}{\delta_{\min}}
\end{equation*}
uniformly for all $\delta\geq \delta_{\min}$, hence the value of $\eps_1$ should be replaced by $\eps_1 = \delta \, \frac{\widetilde U(\delta_{\min},s)}{\delta_{\min}}$.

\subsection{Proof of Lemma \ref{lemma:simple}.}
\label{proof:simple}

Observe that 
\begin{multline*}
\frac{1}{\sqrt{k}}\sum_{j=1}^k  \l( \rho''\l( \sqrt{n}\frac{ \bL_j(f) - \m L(f)}{\Delta} \r) - \mb E \rho''\l( \sqrt{n}\frac{ \bL_j(f) - \m L(f)}{\Delta} \r) \r) 
\\ 
= \frac{1}{\sqrt{k}}\sum_{j\notin J}  \l( \rho''\l( \sqrt{n}\frac{ \bL_j(f) - \m L(f)}{\Delta} \r) - \mb E \rho''\l( \sqrt{n}\frac{ \bL_j(f) - \m L(f)}{\Delta} \r) \r) 
\\
+ \frac{1}{\sqrt{k}}\sum_{j\in J}  \Bigg( \rho''\l( \sqrt{n}\frac{ \bL_j(f) - \m L(f)}{\Delta} \r) - \rho''\l( \sqrt{n}\frac{ \bL_j(f_\ast) - \m L(f_\ast)}{\Delta} \r)
\\
- \mb E \l( \rho''\l( \sqrt{n}\frac{ \bL_j(f) - \m L(f)}{\Delta} \r) - \rho''\l( \sqrt{n}\frac{ \bL_j(f_\ast) - \m L(f_\ast)}{\Delta} \r)\r) \Bigg) \\
+\frac{1}{\sqrt{k}}\sum_{j\in J} \l( \rho''\l( \sqrt{n}\frac{ \bL_j(f_\ast) - \m L(f_\ast)}{\Delta} \r) - \mb E \rho''\l( \sqrt{n}\frac{ \bL_j(f_\ast) - \m L(f_\ast)}{\Delta} \r) \r).
\end{multline*}
Clearly, as $\|\rho''\|_\infty \leq 1$, 
$\l|\frac{1}{\sqrt{k}}\sum_{j\notin J}  \l( \rho''\l( \sqrt{n}\frac{ \bL_j(f) - \m L(f)}{\Delta} \r) - \mb E \rho''\l( \sqrt{n}\frac{ \bL_j(f) - \m L(f)}{\Delta} \r) \r) \r| \leq 2\frac{\m O}{\sqrt k}$. 
Next, repeating the ``slicing argument'' of Lemma \ref{lemma:modulus-unif}, it is not difficult to deduce that with probability at least $1-2e^{-2s}$,
\begin{multline*}
\sup_{f\in \m F(\delta)}\Bigg| \frac{1}{\sqrt{k}}\sum_{j\in J}  \Big( \rho''\l( \sqrt{n}\frac{ \bL_j(f) - \m L(f)}{\Delta} \r) - \rho''\l( \sqrt{n}\frac{ \bL_j(f_\ast) - \m L(f_\ast)}{\Delta} \r)
\\
- \mb E \l( \rho''\l( \sqrt{n}\frac{ \bL_j(f) - \m L(f)}{\Delta} \r) - \rho''\l( \sqrt{n}\frac{ \bL_j(f_\ast) - \m L(f_\ast)}{\Delta} \r)\r) \Bigg) \Bigg|
\leq C(\rho)\,\delta \, \frac{\widetilde U(\delta_{\min},s)}{\delta_{\min}}
\end{multline*}
uniformly for all $\delta\geq \delta_{\min}$. 
Next, we will apply Bernstein's inequality to estimate the remaining term. 
Since $\rho$ is convex, $\rho''$ is nonnegative, moreover, it follows from Assumption \ref{ass:1} that $\rho''(x)\ne 0$ for $|x|\leq 2$, $\rho''(x)=1$ for $|x|\leq 1$, and $\|\rho''\|_\infty = 1$, hence $\l(\mb E\rho''\l( \sqrt{n}\frac{ \bL_j(f) - \m L(f)}{\Delta} \r) \r)^2\geq \l(\pr{\l| \sqrt{n}\frac{ \bL_j(f) - \m L(f)}{\Delta} \r|\leq 1}\r)^2$, 
\begin{equation*}
\mb E\l( \rho''\l( \sqrt{n}\frac{ \bL_j(f) - \m L(f)}{\Delta} \r) \r)^2\leq \pr{\l| \sqrt{n}\frac{ \bL_j(f) - \m L(f)}{\Delta} \r|\leq 1} 
+ \pr{\l| \sqrt{n}\frac{ \bL_j(f) - \m L(f)}{\Delta} \r| \in [1,2]},
\end{equation*}
and 
\begin{multline*}
\var\l(  \rho''\l( \sqrt{n}\frac{ \bL_j(f) - \m L(f)}{\Delta} \r) \r) \leq \pr{\l| \sqrt{n}\frac{ \bL_j(f) - \m L(f)}{\Delta} \r|\leq 1}  - \l(\pr{\l| \sqrt{n}\frac{ \bL_j(f) - \m L(f)}{\Delta} \r|\leq 1}\r)^2 
\\
+\pr{\l| \sqrt{n}\frac{ \bL_j(f) - \m L(f)}{\Delta} \r|\geq 1}
\\
\leq 2 \pr{\l| \sqrt{n}\frac{ \bL_j(f) - \m L(f)}{\Delta} \r|\geq 1}
\leq 2 \frac{\var\l( \ell(f(X))\r)}{\Delta^2}.
\end{multline*}
Bernstein's inequality implies that with probability at least $1 - e^{-s}$,
\begin{equation*}
\frac{1}{\sqrt{k}}\sum_{j\in J}  \l( \rho''\l( \sqrt{n}\frac{ \bL_j(f_\ast) - \m L(f_\ast)}{\Delta} \r) - \mb E \rho''\l( \sqrt{n}\frac{ \bL_j(f_\ast) - \m L(f_\ast)}{\Delta} \r) \r) \leq 2\l( \frac{\sigma(\ell,f_\ast)}{\Delta}\sqrt{s} +\frac{s}{\sqrt k}\r),
\end{equation*}
hence the desired conclusion follows.

\subsection{Proof of Lemma \ref{lemma:quadratic}.}
\label{proof:quadratic}

In the context of regression with quadratic loss, $\omega(\delta)$ takes the form 
\[
\omega(\delta) = \mb E \sup_{\ell(f)\in \m F(\delta)}
\l|\frac{1}{\sqrt{N}}\sum_{j=1}^N \Big( (Y_j - f(Z_j))^2 - (Y_j - f_\ast(Z_j))^2 - \mb E\l( (Y_j - f(Z_j))^2 - (Y_j - f_\ast(Z_j))^2\r) \Big)\r|.
\]
In view of Bernstein's assumption verified above, $\omega(\delta)$ is bounded by 
\[
\mb E \sup_{\| f - f_\ast\|^2_{L_2(\Pi)}\leq 2\delta}
\l|\frac{1}{\sqrt{N}}\sum_{j=1}^N \Big( (Y_j - f(Z_j))^2 - (Y_j - f_\ast(Z_j))^2 - \mb E\l( (Y_j - f(Z_j))^2 - (Y_j - f_\ast(Z_j))^2\r) \Big)\r|.
\]
To estimate the latter quantity, we will use the approach based on the $L_\infty(\Pi_n)$-covering numbers of the class $\m F$ (e.g., see \cite{bartlett2012}). We will also set 
\[
B(\m F;\tau) := \{ f\in \m F: \| f - f_\ast\|^2_{L_2(\Pi)}\leq \tau \}.
\] 
It is easy to see that 
\[
(Y - f(X))^2 - (Y - f_\ast(X))^2 = (f(X) - f_\ast(X))^2 + 2(f(X) - f_\ast(X))(f_\ast(X) - Y),
\]
hence 
\begin{multline}
\label{eq:square-process}
w(\delta) \leq 
\mb E \sup_{ B(\m F;2\delta) }\l|\frac{1}{\sqrt{N}}\sum_{j=1}^N (f(Z_j) - f_\ast(Z_j))^2 - \mb E(f(Z_j) - f_\ast(Z_j))^2 \r| \\
+2 \, \mb E \sup_{ B(\m F;2\delta) }\l|\frac{1}{\sqrt{N}}\sum_{j=1}^N (f(Z_j) - f_\ast(Z_j))(Y_j - f_\ast(Z_j) \r|.
\end{multline}
We will estimate the two terms separately. 
By assumption, the covering numbers of the class $\m F$ satisfy the bound 
\begin{equation}
\label{eq:cov-number}
N\l( \m F, L_2(\Pi_N),\eps \r)\leq \l( \frac{A \|F\|_{L_2(\Pi_N)}}{\eps}\r)^V \vee 1
\end{equation}
for some constants $A\geq 1$, $V\geq 1$ and all $\eps>0$. We apply bound of Lemma \ref{lemma:chaining1} to the first term in \eqref{eq:square-process} to get that 
\begin{multline*}
\mb E \sup_{ B(\m F;2\delta) }\l|\frac{1}{\sqrt{N}}\sum_{j=1}^N (f(Z_j) - f_\ast(Z_j))^2 - \mb E(f(Z_j) - f_\ast(Z_j))^2 \r| 
\\
\leq C \l( \sqrt{2\delta}\sqrt{\Gamma_{N,\infty}(B(\m F;2\delta)) } \bigvee \frac{\Gamma_{N,\infty}(B(\m F;2\delta))}{\sqrt{N}}\r).
\end{multline*}
To estimate $\Gamma_{n,\infty}(B(\m F;2\delta)):=\mb E \gamma_2^2(B(\m F;2\delta); L_\infty(\Pi_N))$, we will use Dudley's entropy integral bound.
Observe that 
\[
\mathrm{diam}\l( B(\m F;2\delta); L_\infty(\Pi_N) \r) \leq 2 \| F\|_{L_\infty(\Pi_N)}.
\]
Moreover, for any $f,g\in \m F$, 
\[
\frac{1}{N}\sum_{j=1}^N (f(Z_j) - g(Z_j))^2 \geq \frac{1}{N}\max_{1\leq j \leq N}(f(Z_j) - g(Z_j))^2,
\]
hence $N( B(\m F;2\delta),L_\infty(\Pi_N),\eps)\leq N\l( B(\m F;2\delta), L_2(\Pi_N),\frac{\eps}{\sqrt{N}} \r)$ and,  whenever \eqref{eq:cov-number} holds, 
\[
\log N(B(\m F;2\delta),L_\infty(\Pi_N),\eps) \leq V \log_+\l( \frac{A\sqrt{N}\| F\|_{L_2(\Pi_N)}}{\eps}\r),
\]
where $\log_+(x):=\max(\log x, 0)$. It yields that 
\begin{multline*}
\Gamma_{N,\infty}(B(\m F;2\delta)) \leq \mb E \l( \sqrt{V} \int\limits_0^{2 \| F \|_{L_\infty(\Pi_N)}} \log_+^{1/2} \l( \frac{A \|F\|_{L_2(\Pi_N)}\sqrt{N}}{\eps} \r) d\eps \r)^2 
\\
\leq C\, V \mb E \l( \l\| F \r\|^2_{L_\infty(\Pi_N)} \log\l(\frac{A\sqrt{N}\|F\|_{L_2(\Pi_N)}}{\l\| F \r\|_{L_\infty(\Pi_N)}} \vee e\r) \r)
\leq C\, V \log(A\sqrt{N}) \mb E  \l\| F \r\|^2_{L_\infty(\Pi_N)}
\end{multline*}
for an absolute constant $C>0$. Finally, since $\| F \|_{\psi_2}<\infty$,
\[
\mb E  \l\| F^2 \r\|_{L_\infty(\Pi_N)} \leq C_1\log(N) \| F^2 \|_{\psi_1}
=C_1\log(N) \| F \|^2_{\psi_2},
\] 
hence
\begin{multline}
\label{eq:quadratic-bound}
\mb E \sup_{ B(\m F;2\delta) } \l|\frac{1}{\sqrt{N}}\sum_{j=1}^N (f(Z_j) - f_\ast(Z_j))^2 - \mb E(f(Z_j) - f_\ast(Z_j))^2 \r|
\\ 
\leq C_2 \l( \sqrt{\delta}\sqrt{V}\log(A^2 N)\| F \|_{\psi_2} \bigvee \frac{V \| F \|^2_{\psi_2}\log^2(A^2 N)}{\sqrt{N}} \r).
\end{multline}
Next, the multiplier inequality \cite{wellner2013weak} implies that 
\begin{multline*}
\mb E \sup_{ B(\m F;2\delta) }\l|\frac{1}{\sqrt{N}}\sum_{j=1}^N (f(Z_j) - f_\ast(Z_j))(Y_j - f_\ast(Z_j) \r| 
\\
\leq C \|\eta\|_{2,1}\max_{k=1,\ldots,N}\mb E\sup_{ B(\m F;2\delta) }\l|\frac{1}{\sqrt{k}}\sum_{j=1}^k (f(Z_j) - f_\ast(Z_j)) \r|. 
\end{multline*}
Using symmetrization inequality and applying Dudley's entropy integral bound, we deduce that for any $k$
\begin{multline*}
\mb E\sup_{ B(\m F;2\delta) }\l|\frac{1}{\sqrt{k}}\sum_{j=1}^k (f(Z_j) - f_\ast(Z_j)) \r| 
\leq C\sqrt{V}\mb E\int_0^{\sigma_k}  \log^{1/2} \l( \frac{A \|F_{2\delta}\|_{L_2(\Pi_k)}}{\eps} \r) d\eps 
\\
\leq C_1 \sqrt{V} \mb E\l(\sigma_k \log^{1/2}\l( \frac{eA \|F_{2\delta}\|_{L_2(\Pi_k)} }{\sigma_k} \r)\r),
\end{multline*}
where $F_{2\delta}$ is the envelope of the class $B(\m F;2\delta)$ and $\sigma_k^2:=\sup\limits_{f\in B(\m F;2\delta)} \| f - f_\ast \|^2_{L_2(\Pi_k)}$. Cauchy-Schwarz inequality, together with an elementary observation that $k\sigma_k^2 \geq \|F_{2\delta}\|^2_{L_2(\Pi_k)}$, gives
\[
\mb E\l(\sigma_k \log^{1/2}\l( \frac{eA \|F_{2\delta}\|_{L_2(\Pi_k)} }{\sigma_k} \r)\r) \leq \sqrt{\mb E \sigma_k^2} \,\log^{1/2}(eA\sqrt{k}).
\]
According to \eqref{eq:quadratic-bound},
\[
\mb E \sigma_k^2 \leq 2\delta + C_2 \l( \sqrt{\delta}\sqrt{\frac{V}{N}}\log(A^2 N)\| F \|_{\psi_2} \bigvee \frac{V \| F \|^2_{\psi_2}\log^2(A^2 N)}{N} \r).
\]
Simple algebra now yields that 
\begin{multline}
\label{eq:prod-bound}
\mb E \sup_{ B(\m F;2\delta) }\l|\frac{1}{\sqrt{N}}\sum_{j=1}^N (f(Z_j) - f_\ast(Z_j))(Y_j - f_\ast(Z_j) \r| 
\\
\leq C\|\eta\|_{2,1}\sqrt{V\log(e^2A^2 N)} \l( \sqrt{\delta} + \sqrt{\frac{V}{N}}\log(A^2 N)\| F \|_{\psi_2} \r).
\end{multline}
Finally, combination of inequalities \eqref{eq:quadratic-bound} and \eqref{eq:prod-bound} implies that 
\[
w(\delta)\leq \widetilde \omega(\delta):=C\l( \sqrt{\delta}\sqrt{V} \log(A^2 N)(\| F \|_{\psi_2} + \|\eta\|_{2,1})
\bigvee \frac{V (\| F \|^2_{\psi_2}+ \|\eta\|^2_{2,1})\log^2(A^2 N)}{\sqrt{N}}\r),
\]
where $\widetilde \omega(\delta)$ is of strictly concave type, hence 
\[
\bar \delta\leq C(\rho) \frac{V\log^2(A^2 N)(\| F \|^2_{\psi_2} + \|\eta\|^2_{2,1})}{N}
\]
thus proving the claim.

\section{Further numerical study.}
\label{sec:extra-numeric}

We present additional sresults of numerical experiments omitted in the main text. 

\subsection{Application to the ``Communities and Crime'' data.}

\begin{figure}[h]
\includegraphics[scale=0.35]{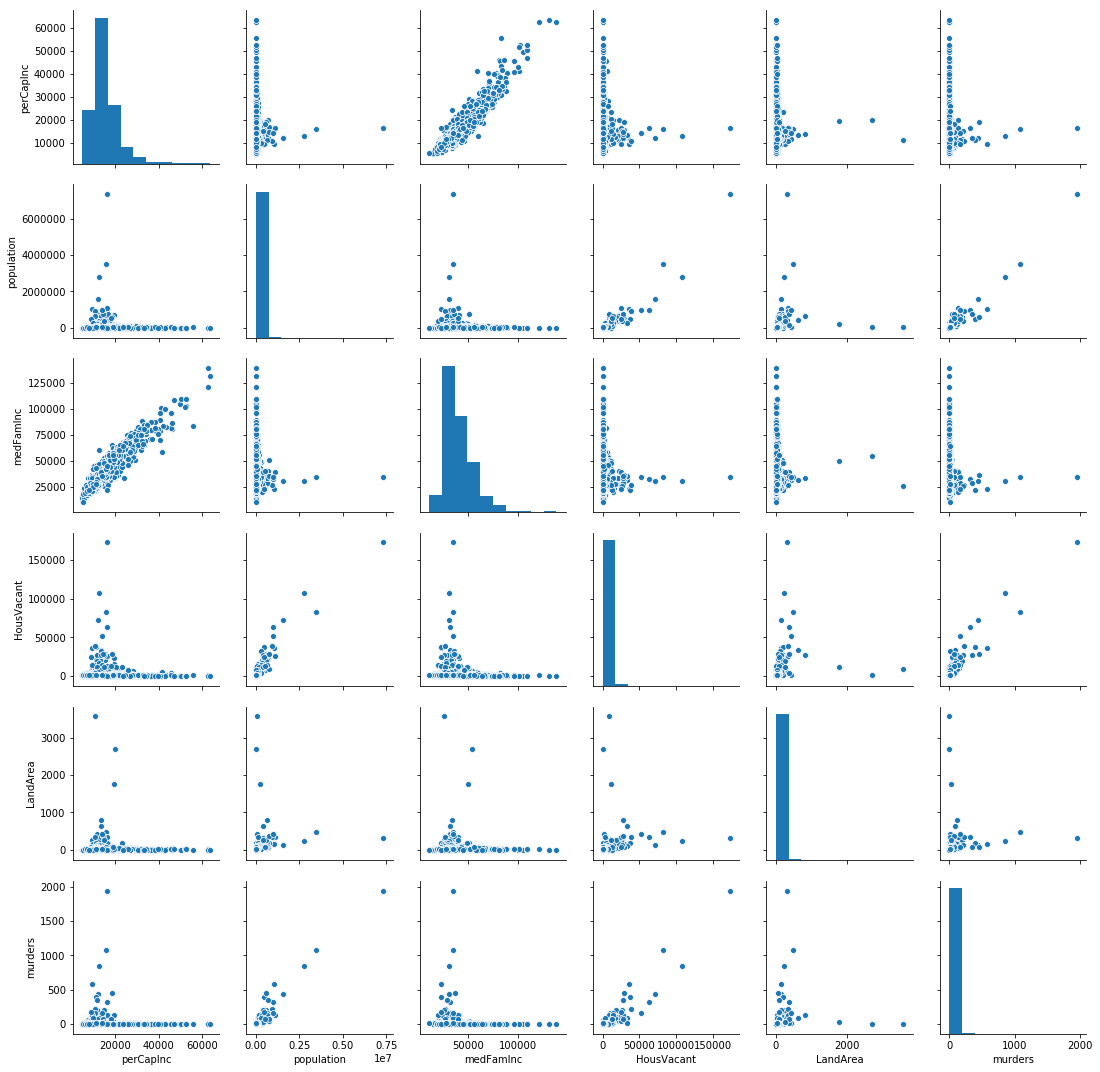}
\caption{Pairplot detailling the 2D marginals of the dataset.\label{fig:pairplot}}
\end{figure}

We compare performance of our methods with the ordinary least squares regression applied to a real dataset.
The dataset we chose is called ``Communities and Crime Unnormalized Data Set'' and is available through the UCI Machine Learning Repository. These data contain $2215$ observations from a census and law enforcement records. 
The task we devised was to predict the crime activity (represented as the count of incidents) using the following features: the population of the area, the per capita income, the median family income, the number of vacant houses, and the land area. The choice of this specific dataset was motivated by the fact that it likely contains a non-negligible number of outliers due to the nature of the features and the fact that the data have not been preprocessed, hence the advantages of proposed approach could be highlighted. 
Figure~\ref{fig:pairplot} presents a pairplot of the dataset; specifically, a pairplot shows all the different scatter plots of one feature versus another (hence, the diagonal consists of the histograms of an individual feature). Such a pairplot offers a visual confirmation of the fact that the data likely contains outliers. 

\begin{figure}[h]
\includegraphics[scale=0.6]{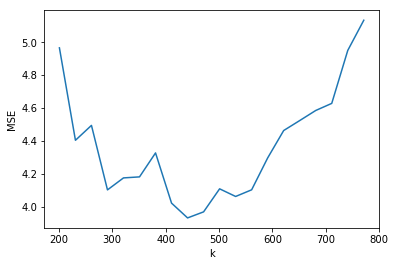}
\caption{Plot of the number of blocks k ($x$-axis) vs the test mean squared error ($y$-axis) obtained with Algorithm 3 on $500$ folds.\label{fig:choice_k_crime}}
\end{figure}

We studied the dependency of the MSE with $k$. Similarly to Figure~\ref{fig:choice_K}, we plotted the MSE as a function of $k$ (figure~\ref{fig:choice_k_crime}). 

\subsubsection{A remark on cross-validation in a corrupted setting.}
\label{sec:cross_val_med}

Cross-validation is a common way to assess the performance of a machine learning algorithm. 
However, cross-validation is not robust when the method itself is not robust (as it is the case here with regression with quadratic loss). For our purposes, we slightly changed the way we approach cross validation. Namely, we still partition the data into $m$ parts used separately for training and testing, however, once we obtain the $m$ scores associated with the $m$ folds, we evaluate the median of these scores instead of the mean.
The rationale behind this approach is that if at least half of the folds do not contain outliers, the results of cross-validation will be robust. To use this approach, we choose $m$, the number of folds, to be large (in the example above, $m=500$). 

\begin{figure}[h]
   \caption{Robust cross-validation with the median.}
   \label{fig:cross_val}
    \begin{algorithmic}
      \STATE \textbf{Input:} the dataset $(X_i,Y_i)_{1\le i\le N}$.
      \STATE Construct the blocks $G_1,\ldots,G_m$, partition of $\{1,\dots,N\}$.
      \FORALL{$j=1,\ldots,m$}
      \STATE Train $\widehat f$ on the dataset $(X_l,Y_l), l \in \bigcup_{i \neq j}G_i$.
      \STATE Compute the test MSE $\mathrm{Score}_j=\frac{1}{|G_j|}\sum_{l \in G_j}(\hat f(X_l)-Y_l)^2$
	   \ENDFOR
      \STATE \textbf{Output:} $\mathrm{Median}\l(\mathrm{Score}_1,\ldots,\mathrm{Score}_m \r)$. 
    \end{algorithmic}
\end{figure}

\begin{figure}[h]
\includegraphics[scale=0.5]{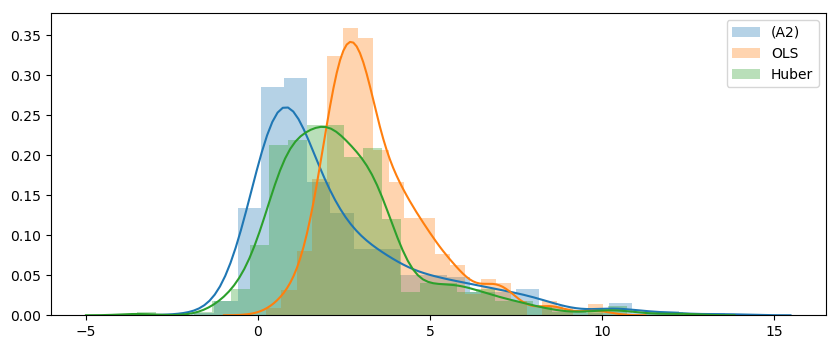}
\caption{Histogram of densities of the logarithm of the MSE for the different methods (light blue corresponds to the approach of this paper (Algorithm 3), orange - to the standard least squares regression, and green - to Huber's regression).
\label{fig:histo}}
\end{figure}

We compared the three algorithms using robust cross-validation with median described above
Our method (based on Algorithm 3) yields MSE of $\simeq e^{4.2}$ while the MSE for the ordinary least squares regression is of order $e^{22.1}$, while the Huber Regression leads to MSE $\simeq e^{8.9}$. 
The empirical density of the logarithm of the MSE over $500$ folds is shown in Figure~\ref{fig:histo}. 

\subsection{Comparison of Algorithm 1 and Algorithm 3.}
\label{sec:num_algo}

We present a numerical evidence that the permutation-invariant estimator $\wh f_N^U$ is superior to the the estimator $\wh f_N$ based on fixed partition of the dataset. 
Evaluation was performed for the regression task where the data contained outliers of type (a), as described in Section \ref{sec:lin-reg}. Average MSE was evaluated over $500$ repetitions of the experiment, and the standard deviation of the MSE was also recored. Results are presented in Figure \ref{fig:comp_shuffle} and confirm the significant improvements achieved by Algorithm 3 over Algorithm 1. We set $k=71$ and $\Delta = 1$ for both algorithms. 


\begin{figure}[h]
\begin{tabular}{|c|c|c|}
  \hline
   \mbox{ } & Algorithm 1 &  Algorithm 3   \\
  \hline
 average MSE &$97.8 $ &  $2$   \\
  \hline
   standard deviation of MSE & $577.3$ & $13$  \\
  \hline
\end{tabular}
\caption{Comparison of Algorithms 1 and 3.
\label{fig:comp_shuffle}}
\end{figure}

\end{document}